\documentclass[manuscript, screen, nonacm]{jair}

\setcopyright{cc}
\acmDOI{10.1613/jair.1.xxxxx} % assigned by JAIR after acceptance

\JAIRAE{Insert JAIR AE Name}
\JAIRTrack{} % Insert JAIR Track Name only if part of a special track
\acmVolume{4}
\acmArticle{6}
\acmMonth{8}
\acmYear{2026}

\RequirePackage[
  datamodel=acmdatamodel,
  style=acmauthoryear,
  backend=biber,
  giveninits=true,
  uniquename=init
  ]{biblatex}

\graphicspath{{figures/}}

\usepackage{tabularx}
\newcolumntype{L}[1]{>{\raggedright\arraybackslash}p{#1}}
\newcolumntype{Y}{>{\raggedright\arraybackslash}X}
\newif\ifprov \provfalse  % provenance tags of the integrated draft, hidden in the JAIR build
\newcommand{\src}[2]{\ifprov{\sffamily\scriptsize[#1 #2]}\ \fi}
\newcommand{\tabnote}[1]{\par\smallskip{\footnotesize #1}}
\xpatchbibmacro{date+extradate}{\setunit*{.\space}}{\setunit*{\addperiod\space}}{}{}
\AtBeginDocument{%
  \fancypagestyle{standardpagestyle}{\fancyhf{}\fancyfoot[C]{\footnotesize\thepage}}%
  \fancypagestyle{firstpagestyle}{\fancyhf{}\fancyfoot[C]{\footnotesize\thepage}}%
  \pagestyle{standardpagestyle}}
\begin{document}

%% Title (short title is used in the running heads).
\title[SMOTE 25 Years of Magic]{SMOTE 25 Years of Magic: What Has Been Done, What the Evidence Shows, and Whom It Reached}

%% =====================================================================
%% Authors. One \author per person, full first names, one e-mail each,
%% ORCID recommended. Placeholders in [brackets] must be filled in.
%% =====================================================================
%% Authors (decision D5, 10 Nov 2026); e-mails, affiliations and ORCID from the team lead, 11 Nov 2026.
\author{Thanapong Intharah}
\authornote{Corresponding Author.}
\orcid{0000-0003-0965-680X}
\email{thanin@kku.ac.th}
\affiliation{%
  \department{Visual Intelligence Laboratory (VILab), Department of Statistics, Faculty of Science}
  \institution{Khon Kaen University}
  \city{Khon Kaen}
  \country{Thailand}
}

\author{Pharamut Makjaroen}
\email{pharamut@kkumail.com}
\affiliation{%
  \department{Visual Intelligence Laboratory (VILab), Department of Statistics, Faculty of Science}
  \institution{Khon Kaen University}
  \city{Khon Kaen}
  \country{Thailand}
}

\author{Kanvara Thawarorit}
\email{kanvara.t@kkumail.com}
\affiliation{%
  \department{Department of Statistics, Faculty of Science}
  \institution{Khon Kaen University}
  \city{Khon Kaen}
  \country{Thailand}
}

\author{Supawit Chaingam}
\email{supawit.c@kkumail.com}
\affiliation{%
  \department{Department of Statistics, Faculty of Science}
  \institution{Khon Kaen University}
  \city{Khon Kaen}
  \country{Thailand}
}

\renewcommand{\shortauthors}{Intharah, Makjaroen, Thawarorit and Chaingam}

%% Structured abstract, as in the JAIR template. Avoid heavy math here:
%% JAIR also publishes the abstract as plain HTML.
\begin{abstract}
{\bf Background:}
SMOTE was published in this journal in 2002 and reviewed here at its fifteenth anniversary in 2018. At twenty-five, it is cited by more than 34,000 indexed works, half of them outside computer science.

{\bf Objectives:}
To establish what the methods built on SMOTE change, how strong the evidence for them is, and whom the technique has reached.
%<<search:f_abstract

{\bf Methods:}
We audited a curated corpus of 76 papers, 37 of which propose a SMOTE-family generator, and tested its coverage with a systematic search of 22,426 records in the PRISMA 2020 format. From the corpus, we collected 1,756 dataset entries, 1,511 published comparisons with SMOTE, and the evaluation design of 71 studies; from the search, we drew 300 applications of SMOTE at random.
%>>search:f_abstract
We used an agentic AI approach to extract, screen, and code under our written rules, and we report how we checked each step.

{\bf Results:}
Every variant re-decides some of the six decisions of the SMOTE rule, and 24 of the 37 are only ever run on two classes. In their own papers, proposed methods beat SMOTE in 79.0\% of the decided comparisons, but independent studies find that the gains shrink with strong learners and tuned thresholds, and that resampling damages calibration.
%<<search:f_abstract_results
The corpus holds 52 of the 3,097 records that pass the screen and covers the early years closely, the recent ones thinly. The reach is very wide, but a benefit at the level of outcomes is undemonstrated, four harms are documented, and an estimated 37\% of applications resample the whole data set before splitting it.
%>>search:f_abstract_results

{\bf Conclusions:}
For tabular classification, a new SMOTE variant should no longer count as a contribution in itself: a new generator should be tested one decision at a time against class weighting and a tuned threshold with a strong learner, with calibration reported. We close with a selection guide, a reporting checklist of fifteen items, and a research agenda.
\end{abstract}

%% JAIR note: do not include ACM CCS Concepts or Keywords.

%% Received/accepted dates are added by JAIR after acceptance:
%\received{DD Month YYYY}
%\received[accepted]{DD Month YYYY}

\maketitle

\section{Introduction}
\label{sec:intro}

\src{M}{1}\textcite{chawla2002smote} introduced the Synthetic Minority Over-sampling Technique (SMOTE) in this journal in 2002. Instead of replicating minority examples, SMOTE creates new ones by interpolating between a minority example and one of its nearest minority neighbors. Sixteen years later, \textcite{fernandez2018smote} reviewed the progress of SMOTE-based learning at its 15-year anniversary. This survey follows the line from 2002 to the present through a corpus of 76 papers.

\src{S}{}Since the 15-year review, the scale of SMOTE's use, the evidence about it, and the software it runs in have all changed. \src{A}{R}On 10 November 2026, OpenAlex indexed 34,416 works that cite the SMOTE paper, more than 5,000 of them published in 2025 alone, and half of the citing works have a primary field other than computer science (Section~\ref{sec:society}). \src{S}{}Works in medicine, engineering, business, and the social sciences now cite it too. \src{A}{R}Since 2018, the field has gained a benchmark of 85 oversamplers on 104 datasets \parencite{kovacs2019empirical}, analyses of the full distribution that SMOTE samples from \parencite{elreedy2019comprehensive,elreedy2024theoretical,sakho2026rebalancing}, and a series of studies, several from clinical prediction, that question whether rebalancing helps at all once a strong learner, a tuned threshold, or a calibrated model is used \parencite{elor2022smote,vandengoorbergh2022harm,carriero2025harms}. The software in which SMOTE is run moved from Weka and KEEL to Python, and over the same period the learners, the measures, and the habits of evaluation changed (Section~\ref{sec:evaluation}).

\src{A}{S}The two earlier accounts in this journal do not answer the questions that these changes raise. The 2002 paper proposed the method. The 2018 review cataloged more than 85 extensions by seven properties of their design and set out open challenges. Neither examined the data on which the methods are tested, the way in which they are evaluated, or the effect of the technique outside the laboratory. This survey therefore asks five questions.

\begin{itemize}
\item \src{A}{S}\textbf{RQ1, methods.} What exactly do the methods that build on SMOTE change, and in which class setting (two classes, several classes, or both) are they shown to work? (Section~\ref{sec:family})
\item \src{A}{S}\textbf{RQ2, data.} On what data is the evidence for these methods built, and can a reader tell exactly which data each comparison used, so that results can be compared across papers and repeated? (Section~\ref{sec:data})
\item \src{A}{S}\textbf{RQ3, evaluation.} How are the methods evaluated, how did that practice change, and why? (Section~\ref{sec:evaluation})
\item \src{A}{S}\textbf{RQ4, effect.} What do independent studies and theory say about when synthetic oversampling helps? (Section~\ref{sec:evidence})
\item \src{A}{S}\textbf{RQ5, reach.} Whom has SMOTE reached, and with what documented benefit and harm? (Section~\ref{sec:society})
\end{itemize}

\src{M}{1}SMOTE creates each synthetic minority example by moving from a minority example $\mathbf{x}_i$ (the seed) toward one of its nearest minority neighbors $\mathbf{x}_j$ (the partner) by a random fraction $\lambda$ of the distance between them, so that the new point lies on the segment between the two (Equation~\ref{eq:smote}). We call a method SMOTE-family when it generates synthetic minority examples with this equation, or with a rule that extends it in one of four ways: (i) a different law for $\lambda$, for example $\lambda \le 0.5$ or a Gaussian step; (ii) a region built around the segment instead of the segment itself; (iii) extrapolation along the line through $\mathbf{x}_i$ and $\mathbf{x}_j$, that is, $\lambda$ outside $[0,1]$; or (iv) a convex combination of more than two instances. A method may change which seeds and partners it uses, how many points it makes, what it removes afterwards, or where in the learning process it runs, and still belong to the family. Methods that sample around a single point without a partner, or draw from a fitted or learned distribution, do not belong to it. \src{S}{}We apply the definition to the generator. A method called ``SMOTE'' that does not interpolate is excluded, and a method with another name that interpolates is included.

\src{A}{S}This survey makes seven contributions.
\begin{enumerate}
\item \src{M}{2, 6}A generation template that writes the SMOTE rule as six decision points (seed selection, sample allocation, neighbor choice, placement, post-filtering, and generation inside a pipeline), and a coding of 37 SMOTE-family methods by the decisions each one changes (Sections~\ref{sec:rule} and~\ref{sec:family}).
\item \src{M}{7}A classification of the 37 methods by the class setting in which their papers actually generate data, which shows that 24 are run on two classes only and that 23 of these test on multiclass sources reduced to two classes (Section~\ref{sec:classsetting}).
\item \src{D}{1}An audit of 1,756 dataset entries: how the data were obtained and modified, how often the same name denotes different data, how difficult the verifiable tasks are when measured directly, and where the printed wins over SMOTE come from (Section~\ref{sec:data}).
\item \src{E}{7}An audit of the evaluation design of 71 empirical studies, link by link from the benchmark data to the statistical test, with an account of why practice changed and a protocol for a new study (Section~\ref{sec:evaluation}).
\item \src{S}{}A synthesis of the independent evidence and theory on whether and when synthetic oversampling helps, most of it published since the 15-year review (Section~\ref{sec:evidence}).
\item \src{S}{}A quantified account of the reach of SMOTE beyond machine learning, set against the evidence of benefit and of harm (Section~\ref{sec:society}), followed by a selection guide and a single reporting checklist that merges the recommendations of the three audits (Section~\ref{sec:guidance}).
\item \src{S}{}A demonstration of a human-in-the-loop agentic framework for surveys: AI agents did the extractions, the screen of 22,426 records, the coding of both samples, and five re-implementations under the authors' written rules; each step was checked against a blind second pass, another AI model, or the authors; and the instructions and the records of the checks are released with the data (Section~\ref{sec:ai} and Table~\ref{tab:aisteps}).
\end{enumerate}

\src{A}{S}The evidence reviewed here leads to one position, which we state at the outset and defend in Section~\ref{sec:position}: for tabular classification, a new SMOTE variant should no longer count as a contribution in itself. Twenty-five years of variants have re-decided the same six choices; in our corpus, they are judged mostly by their own authors against plain SMOTE, and neither the corpus nor the independent studies we review show them to beat the cheaper remedies of class weighting and threshold tuning with a strong learner. A new generator is worth publishing when a controlled comparison, one decision at a time and against those remedies, shows what it adds, and when it reports calibration. The position is limited to tabular classification, the setting of the evidence we review.

\src{A}{S}Our claims are bounded by the evidence we review, which is for tabular data and for the learners that were tested, and they allow that oversampling helps in some settings. We describe the methods by the decisions they change and read their printed results paper by paper, because each paper's results come from its own protocol and cannot be pooled into a ranking. \src{S}{}Our account of deep and generative oversamplers (Section~\ref{sec:beyondtables}) places them against the family, which is what our audits cover.

\src{A}{S}Section~\ref{sec:related} sets this survey against the 15-year review and other reviews of imbalanced learning. \src{S}{}The 15-year review lists the extensions and the challenges that remain. This survey examines what the extensions change, what the evidence for them is worth, and whom the technique reached. \src{S}{}Four of the seven design properties of the earlier review map onto our six decision points (Appendix~\ref{app:accounts}), so the two accounts can be read together.

\section{Background}
\label{sec:background}

\subsection{The Class Imbalance Problem}
\label{sec:imbalance}

\src{A}{R}A classification problem is imbalanced when one class, usually the class of interest, has far fewer examples than the others. A learner that minimizes overall error can then ignore the minority class and still score well. Four families of remedies are in use \parencite{he2009learning,branco2016survey,krawczyk2016learning,haixiang2017learning}: data-level methods, which resample the training set (undersampling the majority, replicating the minority, or creating synthetic minority examples); algorithm-level methods, which change the learner (misclassification costs, class weights, a moved decision threshold); ensembles that combine either of these with bagging or boosting; and hybrids. SMOTE belongs to the first family and has become its reference method.

\src{E}{7.1}The 15-year review recalls why accuracy failed on the data that motivated SMOTE: on the mammography data a decision tree reached about 97\% accuracy while always predicting the majority class gave 97.68\%, and the two kinds of error had different costs, so the trade-off between true and false positives had to be shown with ROC analysis \parencite{fernandez2018smote}.

\src{D}{1.5}The imbalance ratio is not the only source of difficulty. The literature on imbalanced learning attributes much of it to other characteristics of the data: overlap between the classes, noisy and borderline examples, small disjuncts, and a shortage of minority examples \parencite{jo2004class,lopez2013insight,napierala2016types}. \src{S}{}Section~\ref{sec:data} measures these characteristics on the verifiable tasks and relates them to the printed wins over SMOTE.

\subsection{The SMOTE Generation Rule}
\label{sec:rule}

\src{M}{2}Let $S_{\min} \subset \mathbb{R}^d$ be the minority examples of a two-class training set and $S_{\mathrm{maj}}$ the majority examples. For a seed $\mathbf{x}_i \in S_{\min}$, let $N_k(\mathbf{x}_i)$ denote its $k$ nearest neighbors within $S_{\min}$ under the Euclidean distance. SMOTE creates a synthetic example by choosing a neighbor $\mathbf{x}_j \in N_k(\mathbf{x}_i)$ at random and placing a point at a random position on the segment that joins them \parencite[pp.~328--329]{chawla2002smote}:
\begin{equation}
\label{eq:smote}
\mathbf{x}_{\mathrm{new}} = \mathbf{x}_i + \lambda\,(\mathbf{x}_j - \mathbf{x}_i), \qquad \lambda \sim U(0,1).
\end{equation}

\src{M}{2}The amount of oversampling is a percentage $N$: at $N = 200\%$ each minority example produces two synthetic points, toward two of its $k = 5$ neighbors, and amounts below 100\% apply the rule to a random subset of the minority class. In the experiments, SMOTE is combined with random undersampling of the majority class \parencite[pp.~331, 337]{chawla2002smote}. The same paper extends the rule to mixed features (SMOTE-NC) and to nominal features (SMOTE-N), the latter without evaluation \parencite[pp.~348--351]{chawla2002smote}.

\src{M}{2}Read as a procedure, Equation~\ref{eq:smote} fixes six choices in the simplest way (Table~\ref{tab:decisions}): every minority example is a seed with equal weight; every seed receives the same number of points; the partner is drawn uniformly from the five nearest minority neighbors; the point is drawn uniformly on the segment; nothing is removed afterwards; and the generator runs once, before training. We call these the six decision points of the rule. Writing each of them as a symbol turns Equation~\ref{eq:smote} into a generation template in which every SMOTE-family method fills the six slots in its own way:
\begin{equation}
\label{eq:template}
\begin{array}{ll}
\text{(1) seed selection:} & i \sim w(i),\ \ \mathbf{x}_i \in S_{\min} \\
\text{(2) sample allocation:} & \text{repeat } g_i \text{ times for seed } \mathbf{x}_i \\
\text{(3) neighbor choice:} & \mathbf{x}_j \sim q(\cdot \mid \mathbf{x}_i) \\
\text{(4) placement rule:} & \mathbf{x}_{\mathrm{new}} = \mathbf{x}_i + \lambda\,(\mathbf{x}_j - \mathbf{x}_i),\ \ \lambda \sim r(\cdot) \\
\text{(5) post-filtering:} & S' \leftarrow F(S \cup S_{\mathrm{syn}}) \\
\text{(6) inside a pipeline:} & \text{steps (1) to (5) run once, or at every step } t \text{ of a learner}
\end{array}
\end{equation}

\begin{figure}[tp]
\centering
\includegraphics[width=0.92\linewidth]{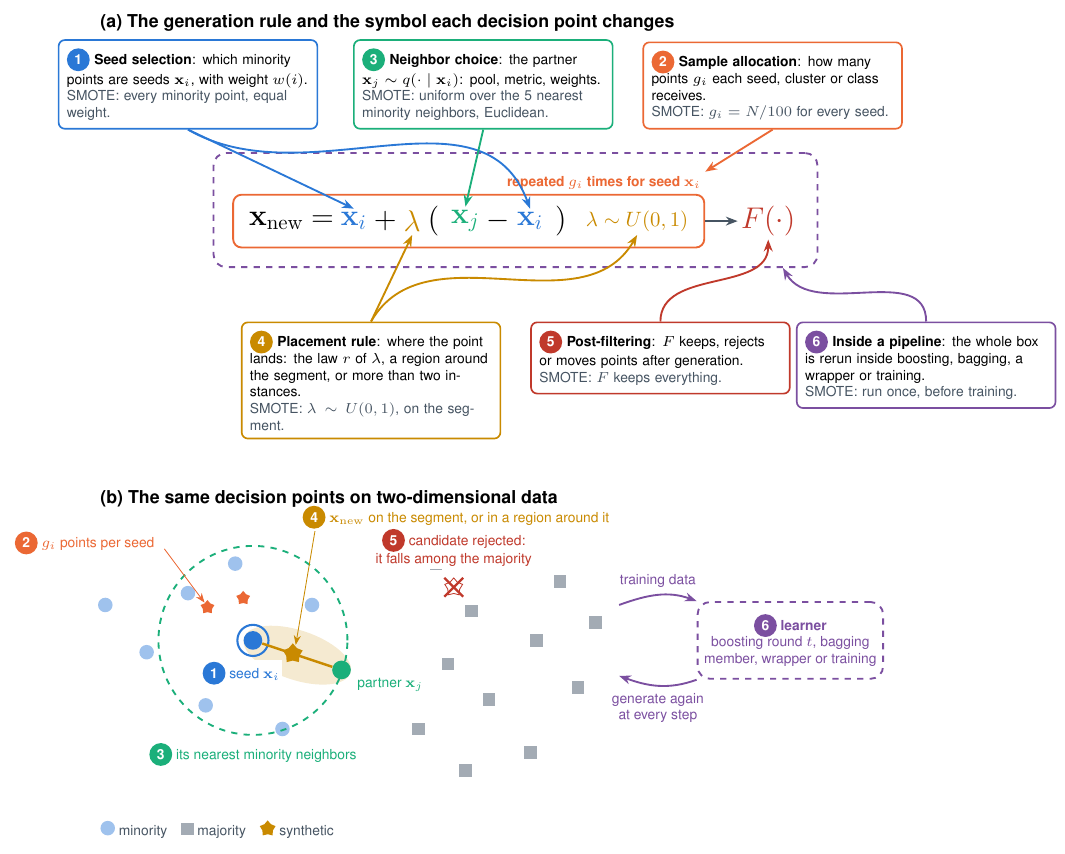}
\caption{\src{M}{2, Fig. 1}The six decision points of the SMOTE generation rule. (a) The symbol of Equation~\ref{eq:smote} that each decision point changes, with SMOTE's own choice under each label. (b) The same decisions on two-dimensional data: which seed, how many points per seed, which partner, where the point lands, which candidates are rejected, and whether generation is repeated inside a learner.}
\Description{\src{A}{S}Two panels. Panel (a) shows the SMOTE equation, x-new equals x-i plus lambda times the difference between x-j and x-i, with six labeled boxes pointing to the part each decision point changes: seed selection, neighbor choice, sample allocation, placement rule, post-filtering, and generation inside a pipeline, each with SMOTE's own choice. Panel (b) shows the same six decisions on a two-dimensional scatter of minority and majority points.}
\label{fig:decisions}
\end{figure}

\src{M}{2}Here $w$ is the seed distribution over $S_{\min}$, $g_i$ the number of points given to seed $\mathbf{x}_i$ (or to its cluster or class), $q$ the partner distribution (its candidate pool, metric, and weights), $r$ the law of $\lambda$ or, more generally, the region in which the new point is placed, and $F$ a filter applied to the enlarged training set. SMOTE is the special case with uniform $w$, constant $g_i = N/100$, $q$ uniform over the five nearest minority neighbors, $r = U(0,1)$ on the segment, $F$ the identity, and a single run before training. Figure~\ref{fig:decisions} shows the same correspondence.

\begin{table}[tp]
\centering\small
\caption{\src{M}{2, Table 1}The six decision points of the SMOTE generation rule. Symbol: the part of the generation template (Equation~\ref{eq:template}) that the decision changes. Papers: SMOTE-family papers in the corpus that change each decision (37 papers; a paper is counted once for every decision it changes).}
\label{tab:decisions}
\begin{tabular}{L{2.5cm}L{1.9cm}L{4.4cm}L{3.9cm}r}
\toprule
Decision point & Symbol & What it decides & SMOTE (2002) & Papers \\
\midrule
Seed selection & $w$: which $\mathbf{x}_i$ & Which minority points become seeds, or how they are weighted & All seeds, equal weight & 24 \\
Sample allocation & $g_i$: how often & How many points each seed, cluster, or class receives & $N/100$ per seed & 21 \\
Neighbor choice & $q$: which $\mathbf{x}_j$ & Which partner is paired with the seed & Uniform over 5 minority neighbors, Euclidean & 22 \\
Placement rule & $r$: $\lambda$ or region & Where the new point lands & $\lambda \sim U(0,1)$, on the segment & 13 \\
Post-filtering & $F$ & What is removed or rejected after generation & Nothing & 12 \\
Inside a pipeline & step (6) & Whether generation runs inside boosting, bagging, a wrapper, or training & Once, before training & 8 \\
\bottomrule
\end{tabular}
\end{table}

\subsection{Earlier Reviews and What This Survey Adds}
\label{sec:related}

\src{E}{7.1}The 15-year review \parencite{fernandez2018smote} cataloged more than 85 extensions by seven properties of their design, listed SMOTE-based ensembles and 22 SMOTE-based methods for 12 other learning paradigms, and set out the open challenges: small disjuncts, noise, class overlap, dataset shift, dimensionality, real-time processing, and big data. It does not examine which measures, protocols, or tests the studies use.

\src{S}{}Table~\ref{tab:related} places this survey among the reviews closest to it: general reviews of imbalanced learning \parencite{he2009learning,branco2016survey,krawczyk2016learning,haixiang2017learning,gao2026comprehensive}, reviews centered on SMOTE and its variants \parencite{fernandez2018smote,kovacs2019empirical,pradipta2021smote,idwan2025comprehensive,udu2025emerging}, reviews of deep generative and other resampling methods \parencite{hayaeianshirvan2025deep,yousefimehr2026balancing}, and a critical review \parencite{hassanat2022stop}. \src{R}{}None of them audits the data and the evaluation behind the published claims, and none goes beyond classifying applications to ask what the technique did outside machine learning.

\begin{table}[tp]
\centering\footnotesize
\caption{\src{R}{}Reviews closest to this survey and what each analyses.}
\label{tab:related}
\begin{tabularx}{\linewidth}{L{3.5cm}L{2.6cm}YL{3.4cm}}
\toprule
Review & Scope & What it analyses & Data, evaluation, or societal analysis \\
\midrule
\textcite{he2009learning}; \textcite{branco2016survey}; \textcite{krawczyk2016learning}; \textcite{haixiang2017learning} & Imbalanced learning in general & Families of remedies, measures, open challenges & Describe the measures, except \textcite{krawczyk2016learning}, which sets out open challenges in seven areas of learning; none audits the data or evaluation of published studies; \textcite{haixiang2017learning} also classify the application domains of the papers they review \\
\textcite{gao2026comprehensive} & Imbalanced learning across data formats & Remedies in four categories for each data format; open-source libraries; challenges & Describes the measures; a small comparison of its own on five tabular and two image data sets; no audit of published claims \\
\textcite{fernandez2018smote} & SMOTE and its extensions to 2017 & More than 85 extensions by seven design properties; other paradigms; challenges & Recalls why accuracy failed; recommends a validation scheme \\
\textcite{kovacs2019empirical} & 85 oversamplers & One experimental protocol on 104 datasets, four learners, and four scores & A common benchmark; no audit of published claims \\
\textcite{pradipta2021smote} & SMOTE variants & Descriptive taxonomy of variants and applications & Not its subject \\
\textcite{idwan2025comprehensive} & Statistical SMOTE variants: 39 studies & Selection-based and adjustment-based variants, with the data characteristics each targets & Not its subject \\
\textcite{udu2025emerging} & SMOTE and GAN variants, mainly 2020 to 2024 & Causes and types of imbalance; variants by application domain; measures & Lists each variant's learners, measures, data sets, and comparators; no audit \\
\textcite{hayaeianshirvan2025deep} & Deep generative oversamplers & GAN and VAE oversamplers against SMOTE, ADASYN, and random oversampling; risks & Compares the results the papers report, across imbalance ratios; no experiments of its own \\
\textcite{yousefimehr2026balancing} & Resampling and augmentation methods, from SMOTE variants to diffusion models & Each method's assumptions, mechanism, and the data it suits; guidelines for practitioners & A search reported with PRISMA; a score of each paper's empirical rigor, among five criteria, to choose the papers it analyses; a small comparison of its own on two fraud data sets \\
\textcite{hassanat2022stop} & More than 70 oversamplers & Whether synthetic points resemble hidden majority examples & One direct check on a few datasets \\
This survey & 76 papers, 37 SMOTE-family methods & Six decision points; class setting; 1,756 dataset entries; evaluation design of 71 studies & All three, with independent evidence and reach \\
\bottomrule
\end{tabularx}
\end{table}

\section{Review Method}
\label{sec:method}

\begin{figure}[tp]
\centering
\includegraphics[width=0.98\linewidth]{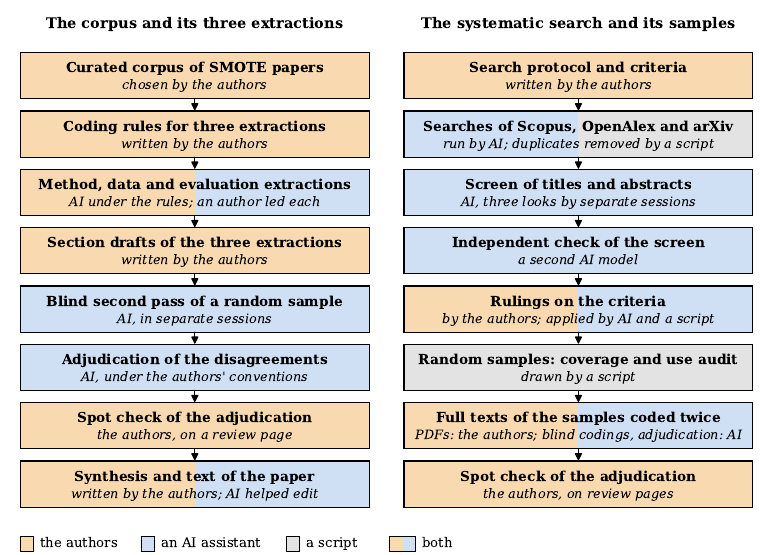}
\caption{\src{S}{}Who did each step of the review. Orange: the authors; blue: an AI assistant (Claude Opus 5.5, and Claude Fable 5.2 for the independent check of the screen and for the editing of the text); gray: a script; both colors: the authors and an AI assistant together.}
\Description{\src{S}{}Two columns of boxes joined by arrows. The left column follows the corpus: the authors chose the papers and wrote the coding rules; an AI assistant did the three extractions under these rules, each led by an author; the authors wrote the section drafts; an AI assistant did a blind second pass on a random sample and adjudicated the disagreements under the authors' conventions, which the authors spot-checked on a review page; an AI assistant helped the authors edit the text, and the authors read, edited, and approved every sentence of it. The right column follows the systematic search: the authors wrote the protocol; an AI assistant ran the searches and a script removed duplicates; an AI assistant screened the records in three looks and a second AI model checked the screen; the authors set the rulings, which an AI assistant and a script applied; a script drew the samples, whose full texts the authors obtained and AI sessions coded twice, blind to each other, with an AI adjudication of the differences, which the authors spot-checked.}
\label{fig:workflow}
\end{figure}

\src{A}{S}Figure~\ref{fig:workflow} gives the steps of the review in order and says, for each, whether the authors, an AI assistant, or a script did it.

\subsection{Corpus}
\label{sec:corpus}

\src{M}{3.1}We call the papers analyzed in this survey the SMOTE25 core corpus. \src{S}{}It is a curated collection of 76 distinct papers, published from 1972 to 2026, that the authors assembled for three main tasks, which they aimed to carry out independently of one another. The method extraction records how each generator builds on SMOTE. The data extraction records the datasets that each paper uses and the scores that it prints for them. The evaluation extraction records how each empirical study is designed and reported, from its measures to its statistical tests. For each task, the authors searched the literature on SMOTE, favoring highly cited papers; all authors then discussed each paper in turn and kept only those that bear on a task and that they judged to have high impact: papers highly cited for their age or published in a leading journal or conference, judged paper by paper. A paper can serve more than one task, so the three sets overlap: the method extraction covers the 37 papers that propose a SMOTE-family generator, the data extraction the 73 that evaluate something on data, and the evaluation extraction the 71 that are empirical studies. Section~\ref{sec:extractions} describes the three extractions, and Table~\ref{tab:corpusmap} in Appendix~\ref{sec:denominators} lists the sets of papers behind every count. On 10 November 2026, the 75 corpus papers with an OpenAlex record had a median of 179 citations, and 47 of them had at least 100; the median was 1,921 for the 17 papers that OpenAlex dates up to 2009 and 27 for the 23 that it dates from 2021 on. A systematic search tests the coverage of the corpus (Section~\ref{sec:search}): 57 of the 76 papers are among its 22,426 records, and 52 of them pass its screen.

\src{M}{3.1}The 76 papers play four roles. Five are foundational works that later methods build on or compare with (ENN, Tomek links, AdaBoost, the multiclass ensemble analysis of \textcite{wang2012multiclass}, and the CTGAN and TVAE tabular generators); 66 propose a new method for imbalanced classification (29 oversampling, 17 hybrid sampling, 9 ensemble, 4 undersampling, 4 deep generative, 2 algorithm-level, and 1 cost-sensitive); three are empirical studies of SMOTE or resampling; and two are surveys.

\src{S}{}The 15-year review was also assembled as a curated selection rather than from a database search: it reports no databases, search terms, time window, inclusion criteria, or screening counts, which this survey reports for its own search (Section~\ref{sec:search}). \src{M}{3.1}It presents the SMOTE-based approaches its authors judged most significant (p.~868), states that more than 85 extensions exist (p.~871), and lists about 90 of them, from 2004 to 2017, in its Table~1 (p.~872).

\src{M}{3.1}We selected the papers on the same principle: the approaches the authors judged most significant for learning from imbalanced data since SMOTE, the foundational works they build on, the empirical studies and critical reviews that evaluate them, and recent work up to 2026. \src{S}{}Unlike the 15-year review, this survey codes and compares as methods only the 37 papers whose generator meets the definition of the SMOTE family; the other 39 are kept as context.

\src{S}{}The corpus overlaps only in part with the two largest lists of the family: ten of its 17 SMOTE-family methods published from 2003 to 2017 appear in the tables of the 15-year review, and eleven of its 37 methods are among the oversamplers compared by \textcite{kovacs2019empirical}. Appendix~\ref{app:accounts} names them and sets the seven design properties of the 15-year review against our six decision points.

\subsection{What Counts as a SMOTE-Family Method}
\label{sec:definition}

\src{S}{}We sorted the 76 papers by how they relate to SMOTE. Those that relate to it received one of four labels: Core (34 papers), where interpolation is the main generator; Partial (3: INOS, EMDO, and GLOS), where it is one of several generators; Study (3), which evaluates or reviews SMOTE without proposing a generator \parencite{fernandez2013analysing,pradipta2021smote,hassanat2022stop}; and Not counted (12), related to SMOTE but generating in another way or not at all. The other 24 papers have no direct link to SMOTE and carry no label. The 34 Core and 3 Partial papers are the 37 SMOTE-family papers of this survey: the SMOTE paper itself and 36 papers that propose variants of it, which we call the 36 variants.

\src{M}{3.3}Ten of the twelve Not counted papers generate synthetic minority examples, but not between two instances: on the Mahalanobis contour of a seed (MDO, AMDO), inside a sphere or ball around one point (MC-CCR, SHSampler, MC-NCLWO, MLOS), along a random direction at a Gaussian distance (GDO), by perturbing points near a cluster centroid (CBHSID), or by hill climbing on a potential field (RBO, MC-RBO). The other two derive from SMOTE but drop its generator: RUSBoost replaces SMOTE in SMOTEBoost by random undersampling, and AdaBoost.NC with random oversampling duplicates examples. Two inclusions are judgment calls: G-SMOTE samples in a hypersphere whose radius is set by the chosen neighbor and which can be shrunk back to the segment, and GAMO replaces two-point interpolation by a learned convex combination of all instances of a class. Both are counted because they generalize the segment and do not replace it.

\subsection{Three Extractions}
\label{sec:extractions}

\src{S}{}In the method extraction, all 37 SMOTE-family papers were read in full, including supplementary material where the main paper left the class setting open. \src{M}{3.4}For each paper, we recorded which of the six decision points the generator changes, a short account of how it builds on SMOTE, and its class category, defined by what the paper shows its generator doing. \emph{Binary only}: every experiment in which the generator creates synthetic data is a two-class problem; a multiclass dataset reduced to two classes counts here, and so does a multiclass extension that is described but never run. \emph{Multiclass only}: the generator creates synthetic data with three or more classes kept, and is never run on a two-class problem \src{A}{S}except to illustrate the choice of a parameter (the one two-class run of GLOS). \src{M}{3.4}\emph{Binary + Multiclass}: the same generator is run on at least one two-class problem and on at least one problem with three or more classes kept. Experiments reported only in supplementary material count. Every statement in the per-method accounts carries a page, section, table, or equation reference. \src{S}{}An earlier pass had recorded the decision points of the 36 variants, and this reading corrected them for 21.

\src{S}{}In the data extraction, for every paper we read the full text and any supplementary file, and recorded each dataset as printed, with its group in the paper, source repository, size, number of features and classes, minority and majority counts, and the imbalance ratio where the paper prints them, together with the table and page they come from. \src{D}{1.1}For each dataset, we also recorded how it reached the experiment: \emph{original} (used as distributed, with its native classes), \emph{repository-defined} (a two-class split published by a repository such as KEEL or imbalanced-learn, used as is), \emph{modified} (the authors chose the minority classes, merged or dropped classes, set a threshold, undersampled, or injected noise), or \emph{synthetic} (generated by the authors). \src{S}{}For the results, we copied, per dataset, the printed score of the proposed method and of SMOTE. A paper may print several results tables, so a fixed rule selects one. The metric is G-mean; if the paper does not report it, AUC; failing that, F-measure; and failing that, accuracy. The classifier is the one that the abstract highlights or, if it highlights none, the first that the paper reports. Where a paper also adds noise or missing values to its data as an experimental condition, the results on the data without them are taken. With the classifier and the condition fixed in this way, we then copied every other metric that the paper reports for both methods. \src{A}{S}When a paper proposes several variants, the proposed method is the variant its abstract names as the contribution, or the first it names. \src{D}{1.1}Plain SMOTE was distinguished from SMOTE embedded in a multiclass or ensemble wrapper. Every extracted value was checked against the PDF a second time.

\src{S}{}In the evaluation extraction, for each empirical study we coded, from the full text and supplementary files, the benchmark data, the base learners, the competing methods, every measure together with its exact definition, the validation protocol, how hyperparameters were chosen, the statistical tests, and the form in which results are reported. \src{E}{7.1}A field the paper does not state was coded as not reported. A second reading recorded why each study made its choices, with a page reference and a short verbatim quote for every code. To show change over time, the studies are grouped into four periods of similar size: 1997 to 2009 (14 studies), 2010 to 2015 (13), 2016 to 2020 (20), and 2021 to 2026 (24). \src{S}{}With periods this small, a single study changes a period's share by 4 to 8 percentage points, so we call a difference between periods a trend only when a share changes by more than 15 percentage points, or when a practice is absent from one period and present in another.

\src{E}{7.1}When we explain why a practice changed, we use three levels of evidence and say which one applies. \emph{Direct} evidence means that the studies themselves give the reason, or that the link can be measured in the corpus. \emph{Circumstantial} evidence means that timing and association point the same way but no study states the link. A \emph{conjecture} is a mechanism we propose that the corpus cannot confirm. \src{A}{S}All three rest on observational evidence.

\src{S}{}Appendix~\ref{sec:denominators} lists the nested sets of papers behind the counts of this survey (Table~\ref{tab:corpusmap}) and says how the quantities that the three extractions define differently were settled.

\subsection{A Systematic Search to Test Coverage}
\label{sec:search}

\begin{figure}[tp]
\centering
\includegraphics[width=0.98\linewidth]{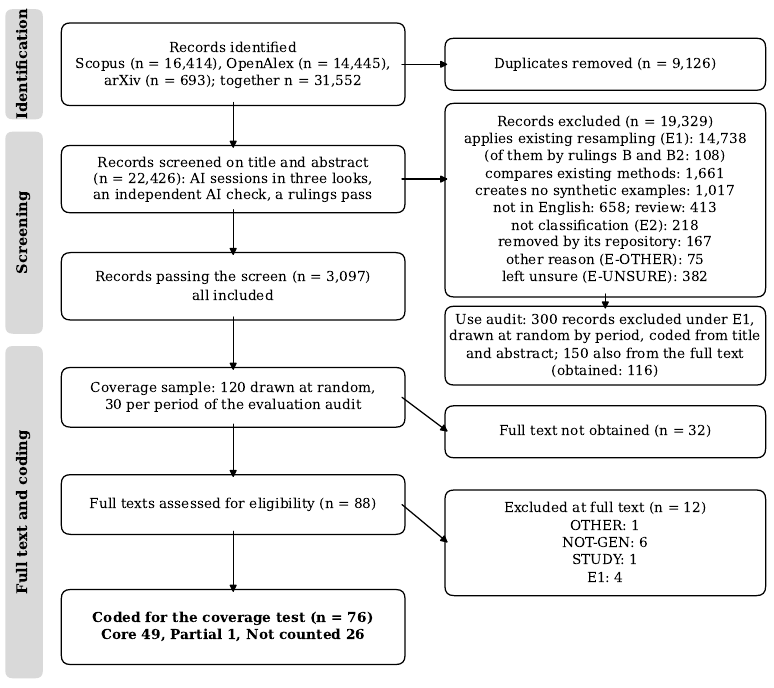}
\caption{\src{A}{S}Flow of the systematic search in the PRISMA 2020 format, with the coverage sample of Section~\ref{sec:beyond} and the use audit of Section~\ref{sec:use}. The screen was done by AI sessions in three looks; an independent AI check by another model then decided the records the looks had left unsure, and a further pass of AI sessions applied rulings B and B2 to the other included records. The records that these steps left unsure are excluded (E-UNSURE).}
\Description{\src{A}{S}A flow diagram. Records identified in Scopus, OpenAlex, and arXiv lose their duplicates and are screened on title and abstract; a box beside the screen gives the excluded records by reason, the largest group being applications of existing resampling (E1). The records that pass the screen lead to a random sample per period assessed in full text, with boxes for papers not retrieved and papers excluded at full text, and ending in the papers coded for the coverage test. A side branch draws the use-audit sample from the exclusions under E1.}
\label{fig:prisma}
\end{figure}

\src{M}{3.2}To test how far the core corpus represents the SMOTE literature, \src{A}{S}we ran a systematic search. Sources: Scopus (title, abstract, and author keywords), OpenAlex, and arXiv for preprints. \src{M}{3.2}Window: January 2002 to the date of the search. The search string combines generation terms (SMOTE, oversampling, over-sampling, ``synthetic minority'', ``adaptive synthetic'', ``synthetic samples'', ``synthetic instances'') with imbalance terms (imbalanced, imbalance, ``minority class'', ``minority classes''). A record is included when it (I1) proposes a method that generates synthetic minority examples, or changes how SMOTE is applied, for classification; (I2) evaluates it on data; and (I3) is a peer-reviewed paper, or a preprint with full text, in English. It is excluded when it (E1) applies existing resampling to a domain without proposing a method; (E2) addresses regression, clustering, or anomaly detection only; (E3) has no accessible full text; or (E4) is an earlier version of an included paper. \src{S}{}An AI assistant screened titles and abstracts under these criteria, with a reason recorded for every decision, and a second AI model checked the screen, as described below; Figure~\ref{fig:prisma} gives the flow in the PRISMA 2020 format \parencite{page2021prisma}, and the online appendix (Section~S12) holds the PRISMA 2020 checklist of the search.

\src{A}{S}The search was run on 10 November 2026. Scopus returned 16,414 documents for the string on title, abstract, and author keywords from 2002 on; OpenAlex returned 14,448 records on title and abstract with the filters of the pilot, of which 14,445 were retrieved; and arXiv returned 693 records on title and abstract. Matching on DOI, then on title and year, leaves 22,426 distinct records from 31,552. The OpenAlex count differs from the pilot's (Table~\ref{tab:pilot}) because OpenAlex changes between releases. Every query, with its date and request log, is given in the online appendix (Section~S4), which also says what is archived and what is deposited.

%<<search:a_screen_body
\src{S}{}The screen was done in three looks by separate sessions of the AI assistant (Claude Opus 5.5), which followed a written codebook (online appendix, Section~S5). The second AI model (Claude Fable 5.2) checked it: blind to the screen's decisions, it decided a random sample of 300 included and 300 excluded records and agrees with the screen on 579 of the 596 that it included or excluded (97.1\%, Cohen's $\kappa = 0.94$), and it then decided the 1,222 records that the screen had left unsure. The authors settled the hard cases with four rulings. Ruling A excludes a record whose only version has been removed by its repository, because it cannot be read in full; these are deposits that Zenodo has taken down, nearly all as spam. Ruling B counts a chain of existing resampling steps as the paper's own method only if the paper presents the chain as its contribution and compares it with at least one other way of resampling. Ruling B2 never counts a standard generator used without a stated change, such as a plain GAN, as the paper's own. Ruling C keeps a record in scope whatever its task: segmentation, survival analysis, or object detection, for example, count like classification. The authors applied rulings A and C themselves. The check applied B and B2 to the records that it decided; for the other included records, a further pass of AI sessions recorded from each title and abstract the facts that the two rulings turn on, and a script applied the rulings to these facts. Appendix~\ref{app:search} gives the pilot of the search string, the three looks, the check, and the rulings in full, and Table~\ref{tab:aisteps} lists each step with its check and its result.

\src{A}{S}Figure~\ref{fig:prisma} shows the flow. The authors excluded the records that the screen left unsure, under a code of their own (E-UNSURE). \src{S}{}Of the 22,426 records, 3,097 pass the screen; 19,329 are excluded, 14,738 of them because they apply existing resampling to a problem of their own (E1), 382 because they remained unsure (E-UNSURE), and 75 for another reason that the check states (E-OTHER), chiefly generators built for a purpose other than imbalanced classification, such as privacy-preserving data sharing, imputation, or causal inference. \src{A}{S}The applications are the frame of the use audit of Section~\ref{sec:use}.
%>>search:a_screen_body

\src{S}{}The corpus and the records of the search can be set against each other paper by paper. Of the 76 corpus papers, 5 were published before 2002, the first year of the search, and 57 are among the 22,426 records: 36 of the 37 SMOTE-family papers and 21 of the other 39. Of these 57, 52 pass the screen, all 36 SMOTE-family papers and 16 others. One of the 36, SMOTEBoost \parencite{chawla2003smoteboost}, had stayed unsure under ruling~B; an unsure record that is a corpus paper, which the authors read in full, was decided by the authors from their reading, and SMOTEBoost was included (Appendix~\ref{app:search}). The 5 that do not pass are a review, three papers coded as comparing existing methods, and one coded as proposing no generator. The search does not return the remaining 14 corpus papers, one of the SMOTE family and 13 of the others. The corpus thus holds 52 of the 3,097 records that pass the screen, and the coverage sample below tests how far its findings hold for the rest.

\src{S}{}As a second test of the search, the authors used an AI session to compare its records with the reference lists of the five reviews of Table~\ref{tab:related} published in 2025 and 2026, and the authors checked the session's coding of the cited works that the records lack. The records that pass the screen hold 136 of the 176 method papers that these lists cite (77\%); the other 40 are mostly named as augmentation, mixup, or generative models, and none has a pair of terms of the search string in its title (Appendix~\ref{app:search}).

%<<search:b_coverage_design
\src{A}{S}Coding thousands of papers in full is beyond a survey, so the coverage of the corpus is tested on a sample. \src{S}{}From the records that pass the screen, we drew 30 at random in each of four periods, 120 in all: up to 2009, 2010 to 2015, 2016 to 2020, and 2021 to 2026, the periods into which the evaluation extraction groups its studies. Each drawn paper was assessed in full text under the criteria above. Each eligible paper was then coded with the fields used for the corpus. The first is its label: Core if interpolation is its main generator, Partial if interpolation is one of several generators, and Not counted if it generates in another way. The others are the decisions of the SMOTE rule that its generator changes, the class setting in which it is run (two classes, more than two, or both), and items of its evaluation, among them whether it repeats its partitions, tests significance, and releases its code. \src{A}{S}The coding was done by AI sessions under a written codebook (online appendix, Section~S5). \src{A}{S}Full texts came from open routes (arXiv, Europe PMC, and open copies that Semantic Scholar and OpenAlex point to) or, failing these, through the authors' university library; 88 of the 120 drawn papers could be obtained, and the counts rest on them. \src{S}{}Separate AI sessions coded these papers a second time, blind to the first coding, an AI session settled the differences, and the authors checked a sample of the settled codes against the papers (Appendix~\ref{app:search}; Table~\ref{tab:aisteps}).
%>>search:b_coverage_design

\subsection{Measured Difficulty, the Independent Benchmark, and Reach}
\label{sec:other-methods}

\src{D}{1.5}Because the papers rarely describe how hard their data are beyond the imbalance ratio, we measured difficulty ourselves on the entries whose data can be identified. \src{A}{S}An entry counts as verifiable when its name matches a public file and the printed number of instances (and of minority examples, when printed) equals that file's, or when the paper prints no size, the name matches a public file and the paper uses that file unmodified, or when the paper's authors publish the modified file themselves. \src{D}{1.5}Every minority example is labeled safe, borderline, rare, or outlier from the classes of its five nearest neighbors \parencite{napierala2016types}, with distances measured by HVDM \parencite{wilson1997improved}. The \emph{noise index} of a task is the share of minority examples that are borderline or outliers. \src{S}{}Three flags, \emph{small disjuncts}, \emph{outliers}, and \emph{small sample}, use published thresholds \parencite{napierala2016types,jain2000statistical}, which Table~\ref{tab:difficulty} gives.

\src{S}{}Sections~\ref{sec:evaluation} and~\ref{sec:evidence} use an independent benchmark that we ran under the protocol of Section~\ref{sec:protocol}. Its core tier compares SMOTE and nine SMOTE-family methods that have public implementations with four baselines that create no synthetic data (no resampling, class weights, a tuned decision threshold, and random oversampling), under six learners, on 110 two-class datasets from 36 source families; Appendix~\ref{app:benchmark} gives the design and the three further tiers.

\src{A}{R}For the reach of SMOTE (Section~\ref{sec:society}), we queried the OpenAlex API on 10 November 2026 for all works that cite the SMOTE paper (OpenAlex work W2148143831) and grouped them by year of publication, primary field and domain, Sustainable Development Goal tag, country of the authors' institutions, and document type. The online appendix (Section~S4) lists the queries. \src{S}{}The counts are bibliometric and show who cites SMOTE. Whether those who cite it benefited is the question of Section~\ref{sec:outcome}. \src{A}{R}For the harms of Section~\ref{sec:harm}, we searched PubMed for studies that compare a model trained on oversampled data with one trained without resampling on a measure of harm; Section~\ref{sec:harm} gives the rule and the result, and the online appendix (Section~S6) the queries and the studies found.

\subsection{Use of AI Assistance}
\label{sec:ai}

\src{S}{}The authors set the criteria and the coding rules, settled the rulings and conventions, decided after each check, and made the checks done by people (Figure~\ref{fig:workflow}). Under these rules, AI sessions (Claude Opus 5.5, Anthropic) worked as agents: each took one task under written instructions that restate the authors' rules for it, read the papers or records itself, and wrote its decisions and codes, with the evidence for them, to files (online appendix, Section~S5). Large tasks were divided into batches, each given to a session of its own, many of which ran in parallel; scripts checked the files for completeness, combined them, and computed the counts of the paper. Table~\ref{tab:aisteps} lists the steps that AI agents did, how each was checked, and the result.

\src{S}{}An AI assistant (Claude Opus 5.5) did the three extractions under the authors' written coding rules, each extraction led by an author (Figure~\ref{fig:workflow}). \src{A}{S}AI sessions read the 37 SMOTE-family papers, extracted the dataset entries and printed scores, and checked them against the PDFs in a verification pass, and in a second reading of the empirical studies recorded each evaluation code with the page and the quote that support it, so that a reader can verify it. \src{S}{}A spot check of 20 of these codes against the PDFs confirmed all 20.

\src{S}{}A second, blind pass followed in November 2026: separate AI sessions recoded a seeded random sample of each extraction, an AI session adjudicated every disagreement from the paper under four conventions set by the authors, and the authors checked a sample of the adjudications and the remaining ones that move a count. For each such check, an AI session built a web page, which we call a review page, that shows every item with its codes, the reason given, and the passage of the paper, and that records the authors' decision. Appendix~\ref{app:aichecks} gives the samples, the conventions, the agreement, and the corrections, and Table~\ref{tab:aisteps} the results.

\src{A}{S}AI sessions also did three further pieces of work, each described where its result is reported: they screened the records of the systematic search (Section~\ref{sec:search}) and of the harm search (Section~\ref{sec:harm}) on title and abstract under written codebooks, and coded the samples drawn from the first (Sections~\ref{sec:beyond} and~\ref{sec:use}); and they implemented from its paper each of the five methods of Table~\ref{tab:recommend} that had no usable public implementation, with tests and a validation against a number the paper prints, for the rerun of Section~\ref{sec:whichmethod} (online appendix, Section~S7). \src{S}{}The authors spot-checked the settled codes of both samples on review pages and read each validation.

\begin{table}[tp]
\centering\small
\caption{\src{S}{}The steps that AI agents did, how each was checked, and the result. Agreement is the share of values or decisions on which two independent passes agree. Appendices~\ref{app:search} and~\ref{app:aichecks} give the detail of the first four rows.}
\label{tab:aisteps}
\begin{tabularx}{\linewidth}{L{3.5cm}YY}
\toprule
Step done by AI agents & How it was checked & Result \\
\midrule
Three extractions of the corpus: 37 method papers, 1,756 dataset entries, 71 empirical studies (Section~\ref{sec:ai}) & A blind second pass of a seeded sample by separate sessions; an AI adjudication under the authors' conventions; the authors' check of 80 adjudications and of the rest that move a count & 84.2\% of 1,706 Data values agree; 13 corrected and 112 blanks filled; all 351 agreements checked hold; the authors agreed with 79 of the 80 and with all of the rest \\
\addlinespace
Screen of the 22,426 records on title and abstract, in three looks (Section~\ref{sec:search}) & A pilot run by other sessions on 400 records; a blind check by another model (Claude Fable 5.2) of 300 included and 300 excluded records & 95.8\% agree in the pilot ($\kappa = 0.859$); 579 of the 596 records the check decided agree (97.1\%, $\kappa = 0.94$); the check's 11 corrections, and its decisions on the 1,222 records left unsure, replace the screen's \\
\addlinespace
Rulings B and B2 on the included records: AI sessions recorded the facts that the rulings turn on, and a script applied the rulings to these facts (Section~\ref{sec:search}) & A comparison with the check by the second model on the 299 included records that both decided & Of the 299, the script fails 44 under B or B2 and the check 10, all of them among the 44, so the script is the stricter; a record that fails is excluded only if its facts were recorded with high confidence, and is otherwise marked unsure \\
\addlinespace
Coding of the two samples: 88 coverage papers and the 300 records of the use audit (Sections~\ref{sec:search} and~\ref{sec:use}) & A blind second coding of all of them by separate sessions; an AI adjudication with the two codes in random order; the authors' check of 55 and 86 settled codes, among them every change that can alter a count & 94.8\% and 93.2\% of the values agree; the adjudication changed 47 and 67 first codes; the authors found every checked code correct \\
\addlinespace
Five methods without a usable public implementation, implemented from their papers for the rerun (Section~\ref{sec:whichmethod}) & A validation of each against a number its paper prints; the authors read each validation & Two reproduce their papers' numbers, one comes close, and two do not; all five are run as their papers describe them (Table~\ref{tab:recommend}) \\
\addlinespace
First pass of the harm search: 429 PubMed records (Section~\ref{sec:harm}) & Reported as an AI first pass & 27 studies included \\
\addlinespace
Comparison of the records with the reference lists of five reviews of 2025 and 2026 (Appendix~\ref{app:search}) & The authors coded the 61 cited works that the session had coded as methods or left unclear & 136 of the 176 method papers that the reviews cite are among the records that pass the screen; 19 cited works left undecided \\
\bottomrule
\end{tabularx}
\end{table}

\src{S}{}The authors also used multiple AI sessions to assist in editing the text of this paper, which they approved passage by passage and then read, edited, and approved sentence by sentence (see the Declarations).

\src{R}{}Agents are now being built for evidence synthesis. otto-SR, an end-to-end workflow of large language models, screens records and extracts data for systematic reviews, and its authors used it to reproduce and update a full issue of Cochrane reviews \parencite{cao2025otto}; AgentSLR evaluates five frontier models, stage by stage, on an automated workflow for epidemiological reviews, and finds structured data extraction the weakest stage and the models not yet reliable enough to work unsupervised \parencite{padarha2026agentslr}. Frameworks that keep people in the loop have also been proposed: TrialMind pairs a language-model pipeline for search, screening, and data extraction with human experts who oversee its output \parencite{wang2025trialmind}, and \textcite{shen2025llmframework} propose a framework that brings language models into all six stages of a systematic review and meta-analysis, with retrieval, agent-based designs, and people in the loop. AutoSurvey, by contrast, writes surveys of a research field with language models from retrieval to the final text \parencite{wang2024autosurvey}. Cochrane, the Campbell Collaboration, JBI, and the Collaboration for Environmental Evidence ask reviewers who use AI to remain responsible for the synthesis, to report the use of AI transparently, and to show that it does not compromise the rigor of the review \parencite{flemyng2025position}. This survey reports its use of AI step by step (Figure~\ref{fig:workflow} and Table~\ref{tab:aisteps}) and releases the instructions and the records of the checks, so that a reader can judge each AI step by its check. \src{S}{}Table~\ref{tab:aisteps} lists the steps and their checks, so that others who use agents in a review can compare their own.

\section{What Has Been Done: The SMOTE Family}
\label{sec:family}

\src{M}{6}Table~\ref{tab:methods} lists the 37 SMOTE-family methods with their class category and the decisions each one changes. A method is marked under every decision it changes, so no method is forced into a single category. Variants differ widely in how much they change: of the 36 variants, 8 change one decision, 8 change two, 8 change three, 9 change four, 2 change five, and one (OREM with its multiclass and boosting versions) changes all six. Appendix~\ref{app:accounts} gives, for every method, how its generator relates to Equation~\ref{eq:smote} and the evidence for its class category.

\begin{table}[p]
\centering\footnotesize
\caption{\src{S}{}SMOTE-family methods in the corpus. Year: that of the version coded (the journal version where one was coded; CISO and QC-SMOTE were coded from their 2026 preprints). Class: B binary only, M multiclass only, B+M binary and multiclass. Link: Core or Partial. Decisions changed: Seed selection (S), sample Allocation (A), Neighbor choice (N), Placement rule (P), post-Filtering (F), inside a pipeLine (L). $\Delta$ vs SMOTE: median difference from plain SMOTE in the paper's own per-dataset table (0 to 1 scale, one metric per paper; datasets where the method is higher in brackets); $\dagger$ SMOTE inside a wrapper; plots, avg.: the paper prints only plots or averages over its datasets, and no value is shown; --: no comparison with SMOTE; n.c.: not comparable, because the paper does not print its method and an identified SMOTE column in the same table, as Section~\ref{sec:wins} requires.}
\label{tab:methods}
\setlength{\tabcolsep}{4pt}\renewcommand{\arraystretch}{0.92}
\begin{tabular}{L{6.3cm}rllcccccc@{\hskip 8pt}l}
\toprule
Method & Year & Class & Link & S & A & N & P & F & L & $\Delta$ vs SMOTE \\
\midrule
SMOTE \parencite{chawla2002smote} & 2002 & B & Core &  &  &  &  &  &  & ref. \\
SMOTEBoost \parencite{chawla2003smoteboost} & 2003 & B+M & Core & $\bullet$ &  &  &  &  & $\bullet$ & +.078 (4/4) \\
SMOTE-Tomek, SMOTE-ENN \parencite{batista2004study} & 2004 & B & Core &  &  &  &  & $\bullet$ &  & +.000 (8/15) \\
Borderline-SMOTE \parencite{han2005borderline} & 2005 & B & Core & $\bullet$ &  & $\bullet$ & $\bullet$ &  &  & plots \\
Cluster-SMOTE \parencite{cieslak2006cluster} & 2006 & B & Core &  &  & $\bullet$ &  &  &  & plots \\
ADASYN \parencite{he2008adasyn} & 2008 & B & Core &  & $\bullet$ &  &  &  &  & +.015 (5/5) \\
LS \parencite{cieslak2008ls} & 2008 & B & Core &  & $\bullet$ &  &  &  & $\bullet$ & -- \\
Safe-Level-SMOTE \parencite{bunkhumpornpat2009safelevel} & 2009 & B & Core & $\bullet$ &  &  & $\bullet$ &  &  & plots \\
RAMOBoost \parencite{chen2010ramoboost} & 2010 & B & Core & $\bullet$ & $\bullet$ &  &  &  & $\bullet$ & +.006 (15/19) \\
SSRBF, DSRBF \parencite{fernandeznavarro2011dsrbf} & 2011 & B+M & Core &  & $\bullet$ &  &  &  & $\bullet$ & +.059 (13/13)$^\dagger$ \\
INOS \parencite{cao2013inos} & 2013 & B & Partial &  & $\bullet$ &  &  & $\bullet$ &  & +.029 (7/7) \\
MWMOTE \parencite{barua2014mwmote} & 2014 & B & Core & $\bullet$ & $\bullet$ & $\bullet$ &  &  &  & +.007 (14/20) \\
RBBoost \parencite{diezpastor2015randombalance} & 2015 & B & Core & $\bullet$ & $\bullet$ &  &  &  & $\bullet$ & n.c. \\
SCUT \parencite{agrawal2015scut} & 2015 & M & Core &  & $\bullet$ &  &  &  &  & +.003 (4/8) \\
SMOTE-IPF \parencite{saez2015smoteipf} & 2015 & B & Core &  &  &  &  & $\bullet$ &  & +.018 (9/9) \\
A-SUWO \parencite{nekooeimehr2016asuwo} & 2016 & B & Core & $\bullet$ & $\bullet$ & $\bullet$ &  &  &  & +.010 (11/16) \\
CGMOS \parencite{zhang2016cgmos} & 2016 & B & Core & $\bullet$ &  &  &  &  &  & +.009 (24/30) \\
SMOM \parencite{zhu2017smom} & 2017 & M & Core &  &  & $\bullet$ &  &  &  & +.004 (19/27) \\
k-means SMOTE \parencite{douzas2018kmeans} & 2018 & B & Core & $\bullet$ & $\bullet$ & $\bullet$ &  &  &  & plots \\
G-SMOTE \parencite{douzas2019gsmote} & 2019 & B & Core &  &  & $\bullet$ & $\bullet$ &  &  & avg. \\
GAMO \parencite{mullick2019gamo} & 2019 & B+M & Core & $\bullet$ & $\bullet$ & $\bullet$ & $\bullet$ &  & $\bullet$ & +.040 (9/9)$^\dagger$ \\
AB-SMOTE \parencite{almajzoub2020absmote} & 2020 & B & Core & $\bullet$ &  & $\bullet$ &  &  &  & $-$.029 (2/8) \\
EMDO \parencite{yao2021emdo} & 2021 & M & Partial & $\bullet$ & $\bullet$ & $\bullet$ & $\bullet$ &  &  & +.043 (9/9)$^\dagger$ \\
HDE \parencite{chen2021hde} & 2021 & B & Core &  &  & $\bullet$ & $\bullet$ &  & $\bullet$ & -- \\
RSMOTE \parencite{chen2021rsmote} & 2021 & B+M & Core & $\bullet$ & $\bullet$ &  &  &  &  & +.021 (21/27) \\
FW-SMOTE \parencite{maldonado2022fwsmote} & 2022 & B & Core &  &  & $\bullet$ &  &  &  & +.018 (30/42) \\
KNNOR \parencite{islam2022knnor} & 2022 & B & Core & $\bullet$ &  & $\bullet$ & $\bullet$ & $\bullet$ &  & avg. \\
SVDD-WSMOTE \parencite{tao2022svddw} & 2022 & B & Core & $\bullet$ & $\bullet$ & $\bullet$ &  & $\bullet$ &  & +.019 (24/24) \\
GLOS \parencite{han2023glos} & 2023 & M & Partial & $\bullet$ & $\bullet$ & $\bullet$ & $\bullet$ &  &  & plots \\
NROMM \parencite{liu2023nromm} & 2023 & B+M & Core & $\bullet$ & $\bullet$ & $\bullet$ & $\bullet$ & $\bullet$ &  & +.190 (20/20)$^\dagger$ \\
OREM, OREM-M \parencite{zhu2023orem} & 2023 & B+M & Core & $\bullet$ & $\bullet$ & $\bullet$ & $\bullet$ & $\bullet$ & $\bullet$ & +.004 (13/21) \\
MC-MBRC \parencite{ma2024mcmbrc} & 2024 & B+M & Core & $\bullet$ & $\bullet$ & $\bullet$ &  & $\bullet$ &  & n.c. \\
MKC-SMOTE \parencite{wang2024mkc} & 2024 & M & Core & $\bullet$ &  & $\bullet$ & $\bullet$ & $\bullet$ &  & +.058 (19/21) \\
GDHS \parencite{yan2025gdhs} & 2025 & B+M & Core & $\bullet$ & $\bullet$ & $\bullet$ &  & $\bullet$ &  & +.024 (15/20)$^\dagger$ \\
CISO \parencite{yadav2025ciso} & 2026 & B & Core & $\bullet$ & $\bullet$ & $\bullet$ & $\bullet$ &  &  & avg. \\
Loyalty-SMOTE \parencite{hu2026loyalty} & 2026 & B & Core & $\bullet$ &  & $\bullet$ &  & $\bullet$ &  & +.028 (20/29) \\
QC-SMOTE \parencite{upman2026qcsmote} & 2026 & B & Core & $\bullet$ & $\bullet$ &  & $\bullet$ & $\bullet$ &  & avg. \\
\midrule
Total (37) & & & & 24 & 21 & 22 & 13 & 12 & 8 & \\
\bottomrule

\end{tabular}
\tabnote{\src{A}{S}The $\Delta$ column is not comparable across rows: each value comes from its own paper's metric, learner, and protocol. \src{S}{}RBBoost and MC-MBRC are marked n.c. for this reason: MC-MBRC prints its comparator as ``MDO'' and no text confirms that it is SMOTE, and RBBoost reports its own method and SMOTE in different supplementary tables that it never compares directly. \src{A}{S}CISO is dated 2026 here, the year of the version coded (its second arXiv version), and 2025 in the reference list, the year of its first version.}
\end{table}

\subsection{The Family by Decision Point}
\label{sec:bydecision}

\src{S}{}The first decision is seed selection, $w$. \src{M}{6.1}SMOTE draws every minority point as a seed with equal probability. A seed rule replaces the uniform draw by a score $s$:
\begin{equation}
\label{eq:seed}
\mathbf{x}_i \sim w(i) = \frac{s(\mathbf{x}_i)}{\sum_{k \in S_{\min}} s(\mathbf{x}_k)}, \qquad s(\mathbf{x}_i) \ge 0 .
\end{equation}
\src{A}{S}Twenty-four methods change the choice of seeds: SMOTEBoost, Borderline-SMOTE, Safe-Level-SMOTE, RAMOBoost, MWMOTE, RBBoost, A-SUWO, CGMOS, k-means SMOTE, GAMO, AB-SMOTE, EMDO, RSMOTE, KNNOR, SVDD-WSMOTE, GLOS, NROMM, OREM, MC-MBRC, MKC-SMOTE, GDHS, CISO, Loyalty-SMOTE, and QC-SMOTE. Appendix~\ref{app:accounts} gives the seed rule of each, with the page that defines it.

\src{S}{}The second decision is sample allocation, $g_i$. \src{M}{6.2}SMOTE gives every seed the same number of points, $g_i = N/100$. An allocation rule shares a total $G$ out by a weight $r_i$:
\begin{equation}
\label{eq:alloc}
g_i = G\,\hat{r}_i, \qquad \hat{r}_i = \frac{r_i}{\sum_k r_k} .
\end{equation}
\src{A}{S}Twenty-one methods change the allocation: ADASYN, LS, RAMOBoost, SSRBF and DSRBF, INOS, MWMOTE, RBBoost, SCUT, A-SUWO, k-means SMOTE, GAMO, EMDO, RSMOTE, SVDD-WSMOTE, GLOS, NROMM, OREM, MC-MBRC, GDHS, CISO, and QC-SMOTE. The rule of each, whether a weight per seed or a quota per cluster, region, or class, is in Appendix~\ref{app:accounts}.

\src{S}{}The third decision is the neighbor choice, $q$. \src{M}{6.3}SMOTE draws the partner uniformly from the five nearest minority neighbors under the Euclidean distance. A partner rule changes the candidate pool $P$, the weights $\omega_{ij}$, or the metric:
\begin{equation}
\label{eq:partner}
\mathbf{x}_j \sim q(j \mid i) \propto \omega_{ij}\,\mathbf{1}\!\left[\mathbf{x}_j \in P(\mathbf{x}_i)\right] .
\end{equation}
\src{A}{S}Twenty-two methods change the choice of partner: Borderline-SMOTE, Cluster-SMOTE, MWMOTE, A-SUWO, SMOM, k-means SMOTE, G-SMOTE, GAMO, AB-SMOTE, EMDO, HDE, FW-SMOTE, KNNOR, SVDD-WSMOTE, GLOS, NROMM, OREM, MC-MBRC, MKC-SMOTE, GDHS, CISO, and Loyalty-SMOTE. Appendix~\ref{app:accounts} gives the candidate pool, weights, or metric of each.

\src{S}{}The fourth decision is the placement rule, $r$. \src{M}{6.4}SMOTE places the point uniformly on the segment. A placement rule changes the law of $\lambda$, replaces the segment by a region $R$, or combines more than two instances:
\begin{equation}
\label{eq:place}
\lambda \sim r(\lambda); \quad \text{or} \quad \mathbf{x}_{\mathrm{new}} \in R(\mathbf{x}_i, \mathbf{x}_j); \quad \text{or} \quad \mathbf{x}_{\mathrm{new}} = \sum_m \alpha_m \mathbf{x}_m,\ \ \alpha_m \ge 0,\ \ \sum_m \alpha_m = 1 .
\end{equation}
\src{A}{S}Thirteen methods change where the new point lands: Borderline-SMOTE, Safe-Level-SMOTE, G-SMOTE, GAMO, EMDO, HDE, KNNOR, GLOS, NROMM, OREM, MKC-SMOTE, CISO, and QC-SMOTE. Appendix~\ref{app:accounts} gives the law of $\lambda$, the region, or the combination that each uses.

\src{S}{}The fifth decision is post-filtering, $F$. \src{M}{6.5}SMOTE keeps every synthetic point. A filter accepts candidates with a test $a$ and removes a set $R$ of examples:
\begin{equation}
\label{eq:filter}
S' = \left( S \cup \{ \mathbf{x} \in S_{\mathrm{syn}} : a(\mathbf{x}) = 1 \} \right) \setminus R .
\end{equation}
\src{A}{S}Twelve methods add a filter after generation: SMOTE-Tomek and SMOTE-ENN, INOS, SMOTE-IPF, KNNOR, SVDD-WSMOTE, NROMM, OREM, MC-MBRC, MKC-SMOTE, GDHS, Loyalty-SMOTE, and QC-SMOTE. Appendix~\ref{app:accounts} says what each removes, rejects, or moves.

\src{S}{}The sixth decision is whether generation runs inside a pipeline. \src{M}{6.6}SMOTE generates once and the learner is trained on the result. In a pipeline, the generator runs at every step $t$ of a learner $\mathcal{A}$, for example with the weights $\mathcal{D}_t$ of a boosting round:
\begin{equation}
\label{eq:pipeline}
D_t = D \cup \mathrm{SMOTE}(D; \mathcal{D}_t), \qquad h_t = \mathcal{A}(D_t), \qquad H(\mathbf{x}) = \arg\max_y \sum_t \alpha_t\, h_t(\mathbf{x}, y) .
\end{equation}
\src{A}{S}Eight methods call the generator repeatedly inside a learner: SMOTEBoost, LS, RAMOBoost, SSRBF and DSRBF, RBBoost, GAMO, HDE, and OREM. Appendix~\ref{app:accounts} says at which step of each learner the generator runs.

\src{A}{S}The lists follow the coding sheet (Table~\ref{tab:methods}). \src{S}{}The coding records which decision points a method changes, and every count in this survey rests on these fields. How a method changes a decision point is told in its per-method account and not in a coded field: sorting the seed rules, for example, into those that favor border seeds and those that favor safe ones would rest on our reading of the accounts, so our counts stay at the level of the decision points.

\subsection{Binary, Multiclass, or Both}
\label{sec:classsetting}

\src{M}{7}Table~\ref{tab:classperiod} shows that most SMOTE-family methods generate data for two classes only. Twenty-four of the 37 are Binary only. Twenty-three of these evaluate on multiclass sources reduced to two classes: 15 reduce the classes themselves (one class against the rest, merged classes, or two classes kept and the rest dropped) and 8 use ready-made two-class splits, mostly from KEEL, without describing the reduction; the remaining one, Cluster-SMOTE, collapses the attack types of its own network data into intrusive versus not. Eleven Binary-only papers mention multiclass use without running it: four describe an extension (SMOTE, ADASYN, k-means SMOTE, QC-SMOTE) and seven name it as future work.

\begin{table}[tp]
\centering\small
\caption{\src{M}{7, Table 4}SMOTE-family methods by class category and period (37 papers).}
\label{tab:classperiod}
\begin{tabular}{lrrrr}
\toprule
Period & Binary only & Multiclass only & Binary + Multiclass & Total \\
\midrule
2002--2009 & 7 & 0 & 1 & 8 \\
2010--2017 & 7 & 2 & 1 & 10 \\
2018--2026 & 10 & 3 & 6 & 19 \\
\midrule
All & 24 & 5 & 8 & 37 \\
\bottomrule
\end{tabular}
\end{table}

\src{M}{7}Methods that handle several classes are more recent. One of the 8 SMOTE-family methods of 2002 to 2009 is Multiclass only or Binary + Multiclass (SMOTEBoost), against 3 of 10 in 2010 to 2017 and 9 of 19 in 2018 to 2026. Five methods are Multiclass only (SCUT, SMOM, EMDO, GLOS, MKC-SMOTE); their rules would cover two classes as a special case, \src{A}{S}but their papers never evaluate it. \src{M}{7}Eight are Binary + Multiclass. In three of these, the two-class evidence is limited: GAMO's two-class runs are in its supplement, MC-MBRC's two-class data are synthetic, and OREM is run on two-class data while its multiclass version OREM-M is run on multiclass data.

\src{M}{7, Table 5}The 13 methods that generate for several classes do so in five ways: to the largest class (SMOM, RSMOTE, MKC-SMOTE); to the mean class size (SCUT, GDHS); class by class against the larger classes (SSRBF and DSRBF, NROMM, OREM-M, MC-MBRC); with a per-class generator or region (GAMO, EMDO, GLOS); or through the ensemble (SMOTEBoost, where AdaBoost.M2 carries the classes). None runs a separate binary generator for each one-vs-one or one-vs-rest subproblem. Decomposition with binary SMOTE nevertheless remains a reference point: in the study of \textcite{fernandez2013analysing}, one-vs-one with SMOTE was among the best combinations.

\src{M}{7}A dataset list is therefore a poor guide to what a method can do: most papers that list multiclass datasets have reduced them to two classes, and a multiclass extension stated in a paper is rarely tested. \src{A}{S}The class category records what the generation step handles. Loyalty-SMOTE runs on five multiclass datasets but generates for the smallest class only and scores with binary metrics, so it is Binary only; the only two-class run of GLOS is a demonstration of parameter tuning, so it is Multiclass only.

\src{M}{7.1}Three of the 37 papers generate data other than tables: INOS for time series, GAMO for images, and KNNOR for tables and images. Classified per data type, INOS is Binary only (48 two-class tasks built from seven UCR datasets), GAMO is Multiclass only on images and Binary + Multiclass on tables (its tabular runs appear only in the supplement), and KNNOR is Binary only on both. Outside tables, the SMOTE family is thin: only GAMO generates for several classes, and only on images does it do so in its main text.

\subsection{What the Method View Shows}
\label{sec:methodview}

\src{S}{}Recent variants change more decisions. \src{M}{9}The mean number of decisions changed per variant rose from 1.7 in 2002 to 2009 (7 variants) to 2.0 in 2010 to 2017 (10) and 3.6 in 2018 to 2026 (19). Twelve of the 36 variants change four or more decisions. When a method changes the seeds, the partner, the placement, and the filter together, a comparison with SMOTE alone cannot tell which change produced the gain.

\src{M}{9}The partner, the placement, and the filter changed most after 2018. Since 2018, 17 of 19 variants change the neighbor choice (5 of 17 before), 11 change the placement rule (2 of 17 before), and 9 add a filter after generation (3 of 17 before). Generation inside a pipeline did not grow (2, 3, and 3 variants in the three periods).

\src{M}{9}The questions of 2002 have mostly been taken up, with one exception. The 2002 paper named four directions for future work \parencite[p.~348]{chawla2002smote}: other ways of creating synthetic points (taken up by the 13 methods that change the placement rule), a focus on misclassified examples, and the use of majority neighbors (both taken up by some of the methods that change the seeds or the partner; Appendix~\ref{app:accounts}). The fourth, an adaptive choice of $k$, is not the subject of any method in the corpus; MKC-SMOTE still lists it as future work. \src{A}{S}A search of the full texts of the 37 papers and their supplements for an adaptive, automatic, or dynamic choice of $k$ or of the neighborhood found none. Outside the corpus, such variants exist, for example a SMOTE built on natural neighbors that needs no fixed $k$ \parencite{li2021natural}; their number is not known.

\src{M}{9}Recent variants cost more than SMOTE. CISO, a preprint, reports a median resampling time of 47 ms per fold against 1.2 ms for SMOTE, about 39 times \parencite{yadav2025ciso}, and QC-SMOTE about 39.47 times \parencite{upman2026qcsmote}. CISO also reports no significant improvement over SMOTE, class weighting, or random oversampling in PR-AUC.

\src{S}{}Some new generators no longer interpolate. \src{M}{9}Since 2018, nine new generators in the corpus have left interpolation for spheres, contours, potential fields, or fitted distributions, and deep generative models (CTGAN, SOS, GOIO, DGOT) learn the minority distribution instead of interpolating between examples.

%<<search:c_beyond
\subsection{The Family Beyond the Corpus}
\label{sec:beyond}

\src{A}{S}The systematic search (Section~\ref{sec:search}) shows how much of the literature the 37 methods stand for. \src{S}{}The screen includes 3,097 records that propose a method of the kind this survey covers. Set against the methods of the corpus by the periods of the evaluation audit, the corpus holds 8 methods against 48 included records up to 2009, 7 against 204 from 2010 to 2015, 7 against 524 from 2016 to 2020, and 15 against 2,321 since 2021 (counting 28 forthcoming records dated 2027). \src{A}{S}The corpus thus covers the early literature closely and the recent literature thinly: for each corpus method of the last period, the screen includes about 155 records.

\src{A}{S}The codes of the screen, given from titles and abstracts, show what the included records propose. \src{S}{}New variants of SMOTE are the largest group in every period, 35\% of them up to 2009 and 42\% since 2021, and hybrids that embed SMOTE in another procedure a falling share (33\% and 20\%). \src{A}{S}Generative oversamplers, which learn the minority distribution with a GAN, an autoencoder, a diffusion or a language model, rose from 4\% and 4\% of the included records in the first two periods to 10\% from 2016 to 2020 and 21\% since 2021 (496 records). The corpus counts one learned generator in the family, GAMO (Section~\ref{sec:definition}), and places the deep generative models beyond its boundary (Section~\ref{sec:methodview}); Section~\ref{sec:beyondtables} describes that branch.

\begin{figure}[tp]
\centering
\includegraphics[width=0.98\linewidth]{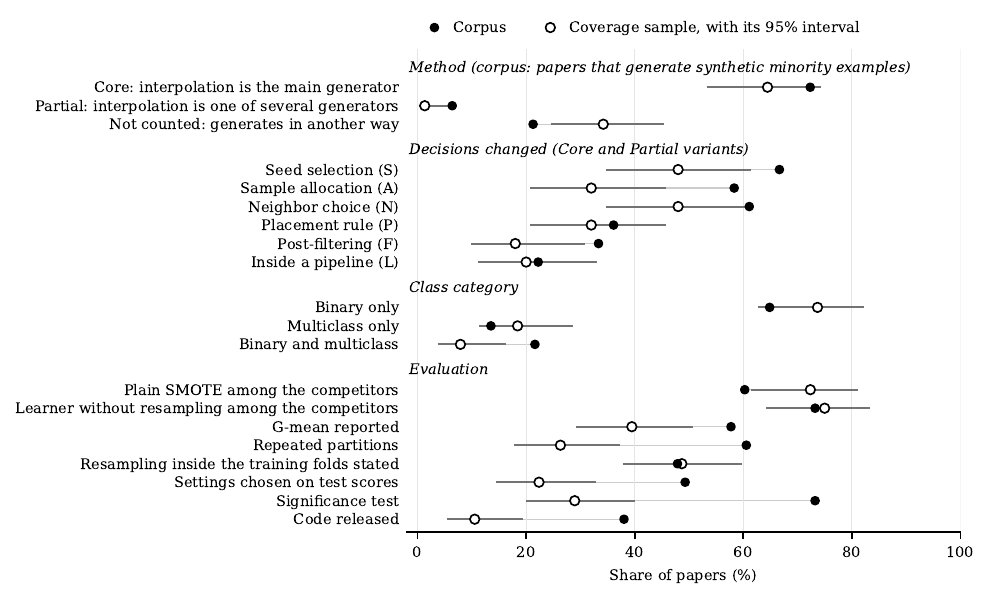}
\caption{\src{S}{}The corpus against the coverage sample of the systematic search (Section~\ref{sec:search}), coded alike: the share of papers with each property. Filled circles: the corpus. Open circles: the eligible papers of the sample, assessed in full text, with a Wilson 95\% interval. The decisions are shares of the Core and Partial methods. The online appendix (Section~S13) gives the counts.}
\Description{\src{S}{}A dot plot with one row for each of twenty properties in four groups (method, decisions changed, class category, evaluation). In each row, a filled circle marks the share of corpus papers with the property and an open circle, with a horizontal line for its 95\% interval, the share in the coverage sample. The two agree on most method and class properties; the corpus is higher on seed selection, sample allocation, repeated partitions, settings chosen on test scores, significance tests, and code release.}
\label{fig:coverage}
\end{figure}

\src{S}{}Whether the findings of Sections~\ref{sec:bydecision} to~\ref{sec:methodview} hold beyond the corpus is tested on the coverage sample (Section~\ref{sec:search}), whose papers are coded in full text as the corpus was. Of its 120 papers, 88 could be assessed and 76 are eligible (49 Core, 1 Partial, and 26 Not counted); Figure~\ref{fig:coverage} sets them against the corpus, and the online appendix (Section~S13) gives the counts. The sample shows a family of the same shape: most methods are Core, most work on two classes only, and about half of the papers state that the resampling stayed inside the training folds. \src{S}{}It differs from the corpus as follows. Its variants change fewer of the six decisions, 2.0 on average against 2.8 in the corpus, with the largest gaps in seed selection and sample allocation. More of its methods generate in another way than by interpolation (34\% against 21\%), most of them published since 2016. Its evaluations are lighter: fewer of its papers repeat their partitions, test significance, or release code. On one point, the sample does better than the corpus: fewer of its papers choose their settings on test scores. The corpus thus describes the methods of the family fairly, while its evaluations are more elaborate than those of the wider literature.

%>>search:c_beyond

\section{What Has Been Done: The Data Behind the Evidence}
\label{sec:data}

\src{D}{1}A reader who wants to compare two SMOTE variants, or to choose a benchmark for a new one, needs answers to questions that the papers rarely answer together. \src{S}{}This section asks four things of the corpus: whether numbers reported on a dataset of the same name can be compared, how broad the evidence behind the common benchmarks is, how difficult the checkable datasets are beyond their imbalance ratio, and on what kind of data a method is reported to beat SMOTE and whether a reader can check the report. It uses the 1,756 dataset entries and the comparisons with SMOTE that we extracted from the corpus. \src{D}{1}Counts use the 73 papers with an evaluation on data.

\src{S}{}In this literature, the data are often described incompletely: the same name can denote different tasks, many modifications cannot be repeated from their description, and 14 of the 57 papers that use SMOTE as a baseline give no per-dataset pair of scores against it. This limits what the reported comparisons show.

\subsection{The Benchmark Landscape}
\label{sec:landscape}

\src{D}{1.2}The 73 papers list 1,756 dataset entries, counting the 93 sets that ten papers generate only for an illustration or a side experiment; 1,755 of them are named. These reduce to 866 distinct names once names are lower-cased, text in parentheses is dropped, and underscores and spaces are treated alike. Treating spellings that differ only in punctuation as one dataset leaves 775 datasets, and grouping each numbered or split version with its main set (for example \texttt{yeast4} or \texttt{glass-0-1-6\_vs\_2} with yeast and glass) leaves 491 source families. Most datasets appear once: 552 of the 775 occur in a single paper. Counted as datasets, ecoli appears in 29 papers, yeast in 22, and glass in 19; counted with their splits, the ecoli family appears in 42 papers, the yeast family in 39, and the glass family in 35, about half of the 73. Half of all uses of a family fall on 41 of the 491 families.

\src{D}{1.2}Splits cut from one source are not copies of each other. Among the verifiable two-class KEEL files, the 14 yeast splits range in imbalance ratio from 2.5 to 41.4 and in noise index from 0.29 to 0.86. They do share the source's attributes, its domain, and often its examples: \texttt{yeast4} and \texttt{yeast-0-5-6-7-9\_vs\_4} use the same 51 minority examples and differ only in which classes form the majority. Outcomes on splits of one source are also related. \src{S}{}In the eleven papers that compare with plain SMOTE on several splits of the same source (282 per-dataset comparisons), whether the proposed method wins does not cluster by family, but the size of its margin does. Metrics differ between papers, so we ranked the margins within each paper and divided each rank by the number of comparisons in the paper, which puts the ranks on a scale from 0 to 1. On this scale, two splits of one source lie 0.29 apart on average, against 0.34 when the family labels are permuted within each paper ($p = 0.002$, 10,000 permutations; an exploratory test). \src{D}{1.2}Papers should therefore report the number of source families alongside the number of datasets, and should not treat each split as independent evidence.

\src{D}{1.2}The frequently used benchmarks are small: among the entries that print a size, half have 800 or fewer instances and half have nine or fewer features. Names printed with 10,000 or more instances in every paper are used by three to nine papers each, so a comparison across method families that wants to cover data size has to look beyond the frequent set. Size, dimensionality, and the number of classes are printed for 75.6\%, 72.4\%, and 78.3\% of the 1,755 named entries, and class proportions in some form for 54.5\%. A measured class overlap exists for 36 entries in two papers, on two incompatible scales. The online appendix (Section~S8) lists every dataset that two written rules select as a candidate benchmark for comparing method families: the 33 datasets used by ten or more papers, and the seven used by three to nine papers whose every printed size is at least 10,000 instances.

\subsection{How Datasets Are Chosen and Modified}
\label{sec:modified}

\src{D}{1.3}Modification by the authors is common (Table~\ref{tab:obtained}): 347 of 1,756 dataset entries were changed by the authors themselves, and for 195 the paper does not let a reader tell how the data were obtained. Six papers give no reason for their choice of datasets, or only call them widely used. \src{S}{}Binarizing a multiclass set or raising the imbalance ratio are legitimate ways to probe a method, provided the paper declares the modification where a reader looks for it. \src{D}{1.3}A KEEL file name such as \texttt{yeast-0-5-6-7-9\_vs\_4} declares its split; a paper that prints only ``Yeast'' after choosing its own minority classes does not.

\begin{table}[tp]
\centering\small
\caption{\src{D}{1.3, Tables 2 and 4}How each dataset entry reached the experiment (1,756 entries), and the share of modified and repository-defined entries by period of publication.}
\label{tab:obtained}
\begin{tabular}{lr@{\hskip 22pt}lrrr}
\toprule
How obtained & Entries & Period & Papers & Modified (\%) & Repository (\%) \\
\midrule
Original & 533 & up to 2004 & 5 & 58 & 0 \\
Repository-defined & 468 & 2005--2009 & 9 & 26 & 0 \\
Modified by the authors & 347 & 2010--2014 & 8 & 36 & 0 \\
Synthetic & 213 & 2015--2019 & 18 & 25 & 35 \\
Cannot be told & 195 & 2020--2024 & 24 & 13 & 27 \\
 & & 2025--2026 & 8 & 5 & 37 \\
\bottomrule
\end{tabular}
\end{table}

\src{D}{1.3}Modifications differ in how completely they are described. Of the 347 modified entries (44 papers), 177 in 30 papers change the class structure, most often by keeping one class against the rest (84 entries), and 153 in 10 papers raise the imbalance ratio on purpose, mostly by removing minority examples at random (131 entries). A purpose for the change is stated in 25 of the 44 papers. For 109 of the 347 entries, information needed to repeat the change is missing: the characteristics of the modified set are not printed for 65, the target imbalance ratio for 18, and the minority class for 21. Twelve entries carry label noise, attribute noise, an outlier, or missing values added by the authors to real datasets (8 papers).

\src{D}{1.3}Whether a modification favors the proposed method cannot be read from a paper, but some features of the design make it checkable. Five of the 44 papers also use an unmodified version of a set they modify, thirteen apply a change at several levels, and nine publish their modified files. Where a paper reports comparisons with SMOTE both on sets it modified and on its other real sets (11 papers), the share of wins is higher on the modified sets in two papers, lower in six, and equal in three. Modified versions are rarely shared: the nine papers that modify vehicle print seven different combinations of size and minority count.

\src{D}{1.3}The way data reach the experiment has shifted over time (Table~\ref{tab:obtained}). Modified entries make up 58\% of the entries before 2005 and 5\% in 2025 to 2026, while repository-defined splits, absent before 2015, make up 27 to 37\% of the entries since. Each period holds only five to 24 papers, so a single paper can move a share considerably.

\subsection{The Same Name Is Not the Same Data}
\label{sec:samename}

\src{D}{1.4}Of the 192 datasets that at least two papers print with a size, 90 appear with more than one size: thyroid with 7,200 instances in five papers and 720 in four, dermatology with 358 (the KEEL version) in eight and 366 (UCI) in two, ecoli with 336, 327, 310, and 301, and pima with 768, 769, 635, and 590. Some of these differences are declared subsamples; the name alone does not tell which. MWMOTE \parencite{barua2014mwmote} illustrates the problem inside a single paper: its Satimage (6,435 instances) and Statlandsat (4,435) are the full Statlog Landsat set and its training file, counted as two of its twenty datasets.

\src{S}{}Imbalance ratios are also defined in several ways. \src{D}{1.4}The corpus contains at least nine definitions: the largest over the smallest class, majority over minority in a binary wording applied to multiclass sets, the percentage of the minority, a per-class imbalance degree, a summed pairwise multiclass ratio, the imbalance degree of \textcite{ortigosa2017measuring}, the minimum class prior, the coefficient of variation of the class sizes, and a proportion pair. Some are not convertible into one another, and some papers print the ratio without a formula. Errors also travel with shared benchmark tables: MC-CCR and MC-RBO both print an imbalance ratio of 1.17 for vehicle, a value no definition produces from its class counts.

\subsection{How Difficult Are the Datasets That Can Be Checked?}
\label{sec:difficulty}

\src{D}{1.5}A numerical overlap measure is printed for 36 of the 1,755 named entries, all in two papers, and noise or borderline examples are described for 5. We therefore measured difficulty ourselves on the verifiable entries (Section~\ref{sec:other-methods}). Verification holds for 990 entries from 65 papers, which use 261 files. \src{S}{}For 703 of these entries, in 46 papers, the two-class task that the paper ran can be rebuilt: the file has two classes, or the paper describes its binarization precisely enough to reproduce, or the task is one that imbalanced-learn defines. These entries share 238 tasks. The other 287 entries use, with one exception, files with three or more classes, for which no two-class task of the paper can be rebuilt. \src{D}{1.5}One-vs-rest tasks that we built from multiclass files (347) are not tasks any paper evaluated and are reported apart.

\begin{table}[tp]
\centering\small
\caption{\src{D}{1.5, Table 6}Difficulty profile of the verifiable tasks. A task can carry several flags; entries count how often the corpus uses a task.}
\label{tab:difficulty}
\begin{tabular}{lrrr}
\toprule
 & Binary tasks & Entries using them & One-vs-rest tasks \\
\midrule
All & 238 & 703 & 347 \\
Small disjuncts (rare $\ge$ 20\%) & 33 (14\%) & 65 (9\%) & 29 (8\%) \\
Outliers ($>$ 20\%) & 83 (35\%) & 198 (28\%) & 49 (14\%) \\
Small sample ($N_{\min}/d < 10$) & 160 (67\%) & 454 (65\%) & 162 (47\%) \\
None of the three & 65 (27\%) & 224 (32\%) & 174 (50\%) \\
Imbalance ratio $>$ 9 & 140 & 376 & 164 \\
\bottomrule
\end{tabular}
\tabnote{Noise index, median [P25, P75]: 0.57 [0.29, 0.77] for the binary tasks and 0.29 [0.07, 0.61] for the one-vs-rest tasks.}
\end{table}

\begin{figure}[tp]
\centering
\includegraphics[width=0.9\linewidth]{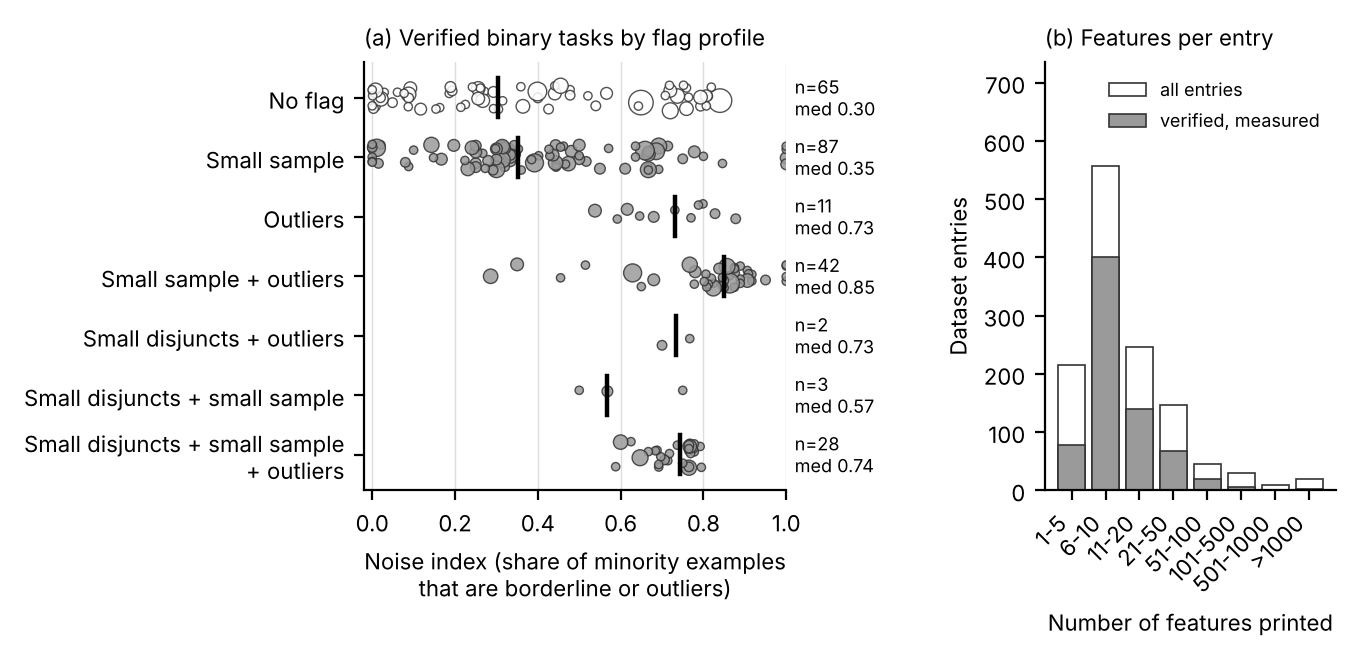}
\caption{\src{D}{1.5, Fig. 1}(a) Noise index of the 238 verifiable binary and binarized tasks, one row per flag profile; marker area grows with the number of entries that use the task, and the vertical bar marks the median. (b) Number of features printed for the 1,271 entries that report it, with the entries that could be verified and measured.}
\Description{\src{A}{S}Panel (a) is a dot plot: one row for each combination of difficulty flags, the noise index of each verifiable task on the horizontal axis from 0 to 1, marker size growing with the number of entries that use the task, and a vertical bar at the median of each row. Panel (b) compares the distribution of the printed number of features for all entries that report it with that of the entries that could be verified and measured; most entries have fewer than 100 features.}
\label{fig:noise}
\end{figure}

\src{D}{1.5}Table~\ref{tab:difficulty} shows that the most common property of the binary tasks is a small sample: 160 of 238 have fewer than ten minority examples per feature. The noise index separates tasks that the flags do not (Figure~\ref{fig:noise}a): among the 65 tasks without a flag it ranges from 0 to 0.84, with 21 tasks at 0.5 or above (haberman 0.84, the numeric German credit data 0.81, contraceptive 0.79, and bupa 0.79 among them). Its median is 0.31, 0.44, 0.50, and 0.65 in the imbalance-ratio bands up to 1.5, 1.5 to 3, 3 to 9, and above 9, and its rank correlation with the imbalance ratio is 0.25. The measurable part of the corpus is also low-dimensional (Figure~\ref{fig:noise}b): the printed number of features has a median of 9, and the 59 entries with more than 100 features come from 54 names in 18 papers, no name used by more than two papers.

\src{D}{1.5}The imbalance ratio, the one difficulty-related number most papers print, says little about where the minority examples lie: tasks with a low ratio can be as mixed as tasks with a high one, and a task without any flag is not necessarily easy. Conclusions drawn from the common benchmarks speak mainly for small, low-dimensional tables with few minority examples per feature. \src{A}{S}For readers who need a small benchmark that covers the measured range, Appendix~\ref{app:reference} describes a reference collection of 42 of the 238 tasks, selected by a fixed rule on the difficulty flags and imbalance-ratio bands.

\subsection{Where Reported Wins over SMOTE Come From}
\label{sec:wins}

\src{D}{1.6}At the paper level, 44 of the 73 papers use plain SMOTE as a baseline, 13 use SMOTE inside a multiclass or ensemble scheme, and 16 do not compare with SMOTE at all. Fewer papers give scores a reader can pair: 29 papers print per-dataset scores for the proposed method and plain SMOTE in the same table. For these papers, we compared the printed values on every metric that the table reports for both methods; a win means the proposed method's printed value is better in the metric's direction, and equal printed values count as a tie. \src{S}{}This gives 1,511 comparisons on 524 datasets; 16 cannot be compared because a value is not printed as a number. Of the other 1,495, the proposed method wins 1,102, ties 100, and loses 293. We call the 1,395 comparisons with a winner decided; the proposed method wins 79.0\% of them (95\% interval 73.3 to 84.5, from a bootstrap that resamples the 29 papers). \src{D}{1.6}Twenty-four of the 29 papers report at least one comparison that SMOTE wins. \src{A}{S}When a paper proposes several variants, the proposed method is the variant its abstract names, or the first it names (Section~\ref{sec:extractions}); this rule changed the variant in four papers. Taking the variants the extraction had first chosen (NB-Comm in NBU, wRACOG, SMOTE with ENN, and SOS with the Sub-VP model) gives 1,107 wins in 1,393 decided comparisons (79.5\%).

\begin{figure}[tp]
\centering
\includegraphics[width=0.98\linewidth]{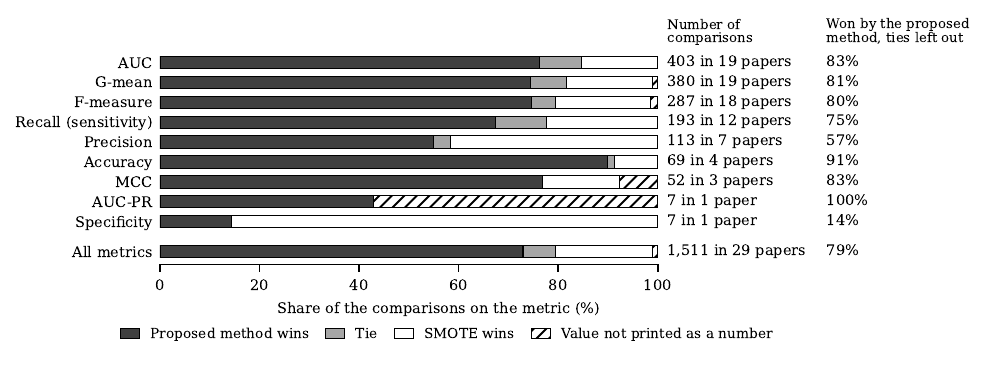}
\caption{\src{S}{}Comparisons of the proposed method with plain SMOTE by metric (29 papers). A comparison is one dataset and one metric in one paper. Each bar divides the comparisons on one metric into those that the proposed method wins, the ties, those that SMOTE wins, and those whose value is not printed as a number. To the right of each bar are the number of comparisons, with the number of papers they come from, and the share that the proposed method wins when only the comparisons with a winner are counted: on AUC, 307 wins against 62 for SMOTE, which is 83\%. Scores are not comparable across papers. The online appendix (Section~S13) gives the counts.}
\Description{\src{S}{}Ten horizontal bars, one for each of nine metrics and one for all metrics, each divided into the share of comparisons that the proposed method wins, ties, that SMOTE wins, and whose value is not printed as a number. The proposed method wins most comparisons on AUC, G-mean, F-measure, recall, accuracy, and MCC, a smaller majority on precision, and one of seven on specificity.}
\label{fig:winsmetric}
\end{figure}

\src{S}{}Figure~\ref{fig:winsmetric} shows that the margin is not the same on every metric. \src{D}{1.6}On AUC, G-mean, F-measure, and recall, the proposed method wins 75 to 83\% of the decided comparisons; on precision, the split is only 62 to 47, and on specificity, SMOTE wins six of seven. Counting one result per dataset (the majority over its metrics) gives 403 wins, 42 ties, and 75 losses on 520 datasets.

\src{D}{1.6}We examined the question that motivated this analysis: whether proposed methods win more often on data the authors modified or generated than on original data. The data do not suggest so. Modified and synthetic data give a win in 75.3\% of decided comparisons (368 of 489) against 80.1\% on original data. With the paper as the unit of clustering, a logistic model (generalized estimating equations) gives an odds ratio of 0.61 (95\% CI 0.37 to 1.00, $p = 0.051$) for modified or synthetic against original data, and a conditional logistic model inside the 12 papers that use both gives 0.53 (0.33 to 0.84, $p = 0.007$). \src{S}{}The direction is the opposite of the hypothesis, and it stays so under the two other ways of counting, but there the estimates cannot be told from no difference: when each dataset counts once (the majority over its metrics), the two models give odds ratios of 0.62 and 0.41, and with one metric per paper, chosen by the rule of Section~\ref{sec:extractions}, 0.79 and 0.57; all four intervals include one ($p$ from 0.08 to 0.51).

\src{A}{S}The comparisons were tallied before any hypothesis was fixed, so this and every other test of the section is exploratory. A Bayesian mixed model gives an odds ratio of 0.64 (95\% interval 0.49 to 0.83); the online appendix (Section~S9) lists every model variant.

\src{D}{1.6}Measured difficulty says more about where the wins fall than the origin of the data. For the 107 verifiable two-class files that appear in these comparisons (23 papers, 496 comparisons), the chance that the proposed method beats SMOTE rises with the noise index: the odds ratio per 0.1 of noise index is 1.18 (95\% CI 1.07 to 1.30, $p = 0.001$) with paper-level clustering and 1.21 (1.07 to 1.37, $p = 0.002$) within papers. The imbalance ratio shows no such relation (odds ratio 0.90 per doubling, $p = 0.42$), nor do the small-sample, outlier, and small-disjunct flags taken one at a time. This analysis is exploratory: it covers the files that could be verified, which are mostly small and low-dimensional. \src{A}{S}Across the five difficulty measures tested one at a time, only the noise index keeps its association after a Holm correction (adjusted $p = 0.006$ with paper-level clustering and $0.009$ within papers; online appendix, Section~S9). Comparisons with SMOTE inside another scheme (558 comparisons in twelve papers) are kept apart.

\src{S}{}Readers can take three points from this. First, read a reported win over SMOTE together with the metric: precision and specificity often favor SMOTE even when the other metrics do not. Second, expect wins over SMOTE to be more frequent on data whose minority examples are borderline or isolated, and less informative on clean data. Third, on this evidence modified data do not inflate the win rate. It matters more whether the data are described well enough to be identified.

\subsection{Can the Reported Results Be Checked?}
\label{sec:checkable}

\src{D}{1.7}Of the 57 papers that use SMOTE as a baseline, 13 give a reader no per-dataset pair of scores in the paper itself: the results are plots, averages, or ranks; a fourteenth, KNNOR, prints its pair of scores only as a mean over its datasets. Even where per-dataset tables exist, they do not always cover the datasets a paper lists. \src{S}{}In the 43 papers with per-dataset comparisons with SMOTE, 201 of the 1,034 listed dataset entries (19.4\%, 95\% interval 8.3 to 31.1, from a bootstrap that resamples the 43 papers) have no row in the table that compares the proposed method with SMOTE; they come from 18 papers, and three papers account for 123 of them.

\src{D}{1.7}Synthetic data are the hardest to check, because a reader can only regenerate them. In 29 papers, 213 dataset entries are generated data. Seven papers give a location for those data (63 entries), three take them from cited work (13 entries), one states that its 60 generated sets are available on request, and thirteen say nothing about where the data can be obtained (64 entries).

\src{D}{1.7}Baseline results are also reused across papers. QC-SMOTE copies its baseline numbers from the VS-SMOTE paper \parencite{qiu2025vssmote}, and every SMOTE-all value in MC-CCR's Table~6 is identical, to two decimals and on all 20 datasets, to the corresponding column of MC-RBO's Table~VI, from the same group and with the same setup. The same pattern appears across research groups: six columns of EMDO's Table~5 \parencite{yao2021emdo} repeat, value for value on all nine shared datasets, the results of AMDO's Table~6 \parencite{yang2018amdo}; EMDO states that it uses the same datasets as AMDO but not that the baseline results were copied. Reusing a baseline is efficient, but unless it is stated, a reader cannot tell whether the comparison was run under identical conditions. For the papers without per-dataset pairs, the headline claim cannot be falsified from what is published. \src{S}{}The claims may be right, but a reader cannot tell whether an average improvement comes from consistent gains or from large gains on a few datasets.

\section{What Has Been Done: How the Family Is Evaluated}
\label{sec:evaluation}

\src{E}{7}Every method in the SMOTE family is judged in the same way. Its authors resample a set of benchmark datasets, train one or more classifiers on the result, score the classifiers on held-out data with a few measures, compare the scores with those of competing methods, and, in most recent papers, test whether the differences are larger than chance. \src{S}{}The method is measured only through its effect on a downstream classifier. \src{E}{7}Each choice in this chain can change which method comes out ahead, and the choices have changed a great deal since 2002. This section follows the chain through the 71 empirical studies of the corpus: 39 evaluate binary problems, 19 multi-class problems, and 13 both.

\subsection{From 2002 to Today}
\label{sec:then-now}

\src{E}{7.2}The original paper evaluated SMOTE on nine datasets with three learners (the C4.5 decision tree, the Ripper rule learner, and naive Bayes) under a single 10-fold cross-validation. For every amount of oversampling, it recorded the true positive rate and the false positive rate, drew the resulting points as ROC curves, and summarized each curve by the area under it (AUC). It argued explicitly that accuracy cannot be used under imbalance, and it did not use G-mean, F1, or a significance test \parencite{chawla2002smote}.

\begin{table}[tp]
\centering\small
\caption{\src{E}{7.2, Table 2}What SMOTE used in 2002 and what the 24 studies published from 2021 to 2026 use.}
\label{tab:thennow}
\begin{tabularx}{\linewidth}{L{2.6cm}L{4.6cm}Y}
\toprule
 & SMOTE (2002) & Studies published 2021 to 2026 ($n = 24$) \\
\midrule
Learners & C4.5, Ripper, naive Bayes & 12 families; median 3 per study; 9 of 24 use a single family \\
Scalar measures & AUC, TPR, TNR & 21 distinct measures; median 3 per study; 19 different combinations among 24 studies \\
Most used measures & AUC (from ROC curves) & G-mean (17 of 24), AUC (13), F1 or F-measure (10), MAUC (6) \\
Other evidence & ROC curves and convex hull, tree size, a 2-D plot & 2-D plots of synthetic samples (15 of 24), runtime (7), direct checks of the resampled data (2) \\
Data and protocol & 9 datasets, single 10-fold CV, no significance test & KEEL data in 67\%, repeated CV in 62\%, a significance test in 83\% \\
Code & not released & released by 12 of 24 \\
\bottomrule
\end{tabularx}
\end{table}

\src{E}{7.2}Table~\ref{tab:thennow} sets this against the recent studies. \src{S}{}The median study still uses three learner families and reports three scalar measures, but the studies now differ more from each other. \src{E}{7.2}Together, the 24 recent studies use 12 learner families and 21 different measures, and among them there are 19 different combinations of measures. A reader who wants to compare the numbers of two recent papers will usually find that they do not share a measure, a learner, or both. \src{S}{}Because of this variety, the evidence for a method cannot be pooled across papers.

\subsection{Learners}
\label{sec:learners}

\src{S}{}Most of the change in learners follows the field's change of software. \src{E}{7.2}Before 2016, studies that name their software name Weka or MATLAB; no study names Python. From 2016, the order reverses: 6 studies name Python or a Python library in 2016 to 2020 and 12 since 2021, while Weka is named by only one study since 2021. C4.5 is the default tree of Weka (J48) and of KEEL, while the decision tree of scikit-learn is CART, and scikit-learn has no C4.5 and no rule learner. Accordingly, C4.5 fell from 64\% of the studies before 2010 to 21\% since 2021, CART rose from 7\% to 50\%, and rule learners fell from 29\% to none. A consequence for a reader is that C4.5 results in older papers and CART results in newer ones are not interchangeable, because the two trees split and prune differently. Random forests and gradient boosting rose from 7\% to 33\%; here the reason the studies give is realism: these are the learners practitioners use, and a resampler that only helps weak learners is of limited value.

\src{E}{7.4}Almost half of the studies use one or two learner families: across the 71 studies, 25 use one family, 9 use two, 12 three, 16 four, and 9 five or more. A single learner limits what a study can conclude, because how much resampling helps depends on the learner: $k$-NN uses the same neighborhoods that SMOTE interpolates in, decision trees are sensitive to pruning that removes the rules for small classes, and strong ensembles already compensate for part of the imbalance. \src{S}{}Forty-nine of the 71 studies say why they chose their learners; the most frequent reason is that the learners are widely used or well known (22 studies), which is a reason of familiarity and says nothing about what the learner tests. \src{E}{7.4}Only 4 studies choose a learner because it is sensitive to imbalance.

\subsection{Measures}
\label{sec:measures}

\src{E}{7.6}Across the 71 studies, we found 27 distinct scalar measures computed from the predictions of a downstream classifier. \src{S}{}They fall into five families and one further measure: overall accuracy or error (13 studies), single-class rates (32), threshold composites such as G-mean, F1, balanced accuracy, and MCC (56), threshold-free ranking measures such as ROC-AUC, its multi-class counterpart MAUC, and PR-AUC (44), multi-class aggregates (21), and misclassification cost (2). \src{E}{7.6}A study reports a median of three; 23 studies rely on one or two. The families answer different questions. A composite asks whether the classifier, at the threshold it actually uses, treats both classes well. A ranking measure asks whether the classifier orders minority examples above majority ones, whatever threshold is later chosen. Resampling acts mainly on the first question, so a study that reports only composites cannot tell whether a resampler improved the classifier or merely moved its threshold. \src{E}{Table 8}Seventy of the 71 studies say why they chose their measures, and a study can give several reasons: that the measures are standard in the literature (40 studies), that accuracy misleads under imbalance (39), a focus on the minority class (25), independence of the threshold (18), suitability for several classes (18), and the costs of an application (7).

\src{E}{7.2}Four causes account for nearly all of the change in measures. \src{S}{}First, single-class rates were folded into composites that contain them. Among the studies that report G-mean or F1, the share that also reports recall on its own fell from 8 of 9 before 2010 to 5 of 21 since 2021, which is why recall fell from 71\% to 21\% while G-mean rose from 29\% to 71\%. Second, the median benchmark grew from 6.5 datasets per study before 2010 to 23 since 2021, and ROC curves, readable for a handful of datasets, gave way to the area under them. Third, multi-class studies grew from 3 in the first period to 15 in the last and brought measures for several classes: MAUC, which averages the AUC over all pairs of classes; MG, the geometric mean of the recalls of all classes; and macro-F1, the mean of the F1 scores of the classes. Fourth, class balance accuracy (CBA) and confusion entropy (CEN) entered this literature in 2020 with two methods of our corpus, MC-CCR and MC-RBO, and every later study that reports them is from the same line of work.

\src{E}{7.6.1}G-mean, the geometric mean of the true positive and true negative rates, is reported by 41 of the 71 studies (58\%), more than any other measure, and it is the only main measure whose share rose in every period. The reason the studies give is mostly popularity. The cause is the way resampling acts on a classifier. Resampling changes the class proportions of the training data, usually to 1:1. \src{S}{}A classifier trained on balanced data predicts the minority class more often, so the true positive rate goes up and the true negative rate goes down. How well the classifier separates the two classes, which is what ROC-AUC measures, changes much less. G-mean multiplies the two rates, so it registers exactly this exchange: it rises when the true positive rate gains proportionally more than the true negative rate loses. A simple case shows how closely the two fit. Suppose that the classifier's score is normally distributed in each class, with the same variance in both. The threshold with the highest G-mean then lies midway between the two class means, where the two rates are equal, and a classifier trained on balanced data uses that same threshold. In this case, balancing the training data moves the classifier to the threshold at which G-mean is highest. \src{E}{7.6.1}A proposed resampler therefore shows its clearest gain under G-mean, and later studies report G-mean to be comparable with it.

\begin{figure}[tp]
\centering
\includegraphics[width=0.88\linewidth]{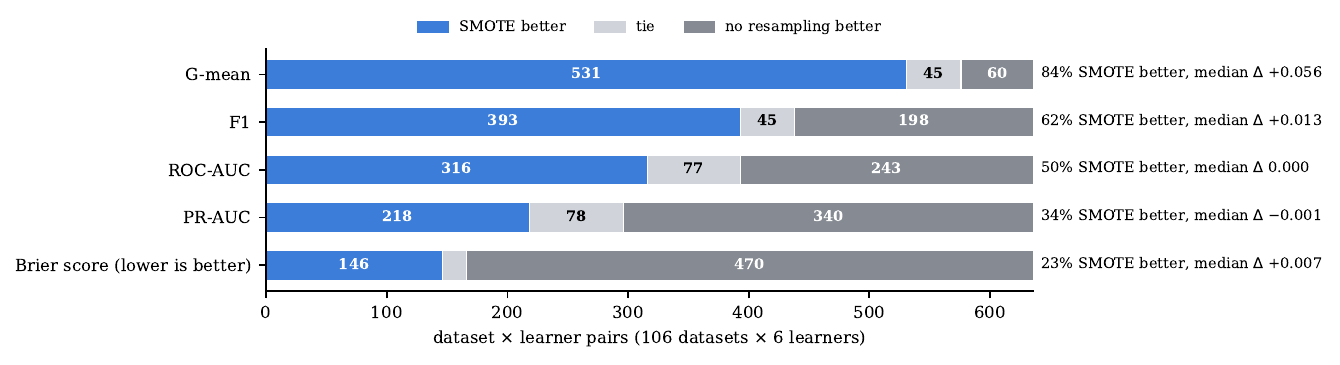}
\caption{\src{A}{S}SMOTE against no resampling, by measure, in the core tier of our benchmark (106 datasets and six learners, 636 dataset-learner pairs; $\Delta$: SMOTE minus no resampling; for the Brier score lower is better).}
\Description{\src{A}{S}Stacked horizontal bars, one for each measure, splitting the 636 dataset-learner pairs into SMOTE better, tie, and no resampling better: SMOTE is better in 84 percent of the pairs for G-mean, 62 percent for F1, 50 percent for ROC-AUC, 34 percent for PR-AUC, and 23 percent for the Brier score; the median difference in score is printed beside each bar.}
\label{fig:smotevsnone}
\end{figure}

\src{A}{S}Our benchmark shows the size of the effect (Figure~\ref{fig:smotevsnone}; Appendix~\ref{app:benchmark}). SMOTE beat no resampling in 531 of the 636 dataset-learner pairs (83.5\%) when measured by G-mean, with a median gain of 0.056. Measured by F1, it won 61.8\% of the pairs, by ROC-AUC 49.7\% (median gain 0.000), and by PR-AUC only 34.3\%, and it raised the Brier score, which penalizes miscalibrated probabilities, in 470 of the pairs. \src{S}{}The gains are in the measures that depend on the threshold (G-mean and F1); in ROC-AUC and PR-AUC, they are small or absent. \src{E}{7.6.1}The same mechanism is a warning: because G-mean rewards any movement of the operating point toward balance, it rewards methods that do nothing but move the threshold, and it ignores precision.

\src{S}{}G-mean and F1 both contain recall. \src{E}{7.6.2}They differ in the second component: G-mean uses the true negative rate, computed within the majority class, while F1 uses precision, which mixes the two classes and so depends on the share $\pi$ of positives:
\begin{equation}
\label{eq:precision}
\mathrm{Precision} = \frac{\mathrm{TPR}\cdot\pi}{\mathrm{TPR}\cdot\pi + \mathrm{FPR}\cdot(1-\pi)}, \qquad \pi = \frac{1}{1 + \mathrm{IR}} .
\end{equation}
\src{E}{7.6.2}When the imbalance ratio is large, a false positive rate that is small as a percentage of the majority class becomes a large number compared with the minority class, so precision falls while G-mean does not change. Averaging F1 over benchmarks whose imbalance ratios range from 1.8 to 130 therefore mixes the quality of a method with the difficulty implied by each dataset's ratio. \src{S}{}A paper's winner can therefore change with the choice between G-mean and F1.

\src{E}{7.6}Definitions are often wrong or missing. In 14 of the 71 studies (20\%), at least one measure is defined in a way that changes its meaning, or its definition cannot be recovered. Two studies call the single-threshold quantity $(1 + \mathrm{TPR} - \mathrm{FPR})/2$, which is balanced accuracy, ``AUC''; one defines G-mean as the geometric mean of recall and precision; two compute MAUC from hard labels instead of scores; and two define average accuracy as the mean one-vs-rest accuracy, which is inflated by true negatives.

\subsection{Competitors and Baselines}
\label{sec:baselines}

\src{E}{7.5}A study compares its proposal with a median of 7 other methods. \src{A}{S}SMOTE is one of them in 57 of the 73 papers with an evaluation on data, 44 times as plain SMOTE and 13 times inside a multiclass or ensemble wrapper (Section~\ref{sec:wins}). \src{A}{S}It is a competitor in 7 of the 15 papers published before 2010 and in 23 of the 25 published since 2021. \src{E}{7.5}A baseline without any resampling appears in 52 studies, but here the trend runs the other way: it was included by 86\% of the studies before 2010 and by only 54\% of those published since 2021.

\src{E}{7.5}The most informative baselines are those that address imbalance without creating data. A cost-sensitive learner or instance weighting is a competitor in only 5 studies, plain class weighting of the learner in one \parencite{yadav2025ciso}, and threshold moving in three. These baselines matter because of the way resampling acts: class weighting, threshold moving, and random oversampling move the operating point too, at almost no cost. A gain of a new resampler is evidence of something more than a shifted threshold only if it exceeds what those simple baselines achieve. The one study that includes class weighting, a preprint run under a protocol it registered in advance, found its new oversampler no better than class weighting, random oversampling, or no resampling across four learners \parencite{yadav2025ciso}.

\subsection{Evidence Beyond Classifier Scores}
\label{sec:directchecks}

\src{E}{7.7}A resampler produces data, yet almost every study judges those data only through a classifier. Only 6 of the 71 studies (8\%) examine the resampled data directly, and none uses such a check instead of a classifier. Three measure how far the synthetic data are from the real distribution, two count what kind of examples the method produces or keeps, and one counts how many synthetic examples a noise filter removes after SMOTE. Outside the empirical studies, \textcite{hassanat2022stop} propose a check that needs no classifier at all: hide part of the majority class, oversample the rest, and count the synthetic points whose nearest original example is a hidden majority example. Direct checks catch failures that scores hide: a resampler that places synthetic points inside majority regions can still raise G-mean, because it moves the threshold, while making the minority class less faithful.

\src{E}{7.3.1}Every one of the 71 studies evaluates resampling through a classification task. \src{S}{}This is a property of the corpus: the 15-year review lists 22 SMOTE-based methods for other learning paradigms, such as regression \parencite{torgo2013smoter}, data streams, and multi-label learning. \src{E}{7.3.1}Every statement about evaluation in this section holds for classification and should not be carried over to other paradigms without separate evidence.

\subsection{Validation Protocol and the Choice of Settings}
\label{sec:protocol}

\begin{figure}[tp]
\centering
\includegraphics[width=0.95\linewidth]{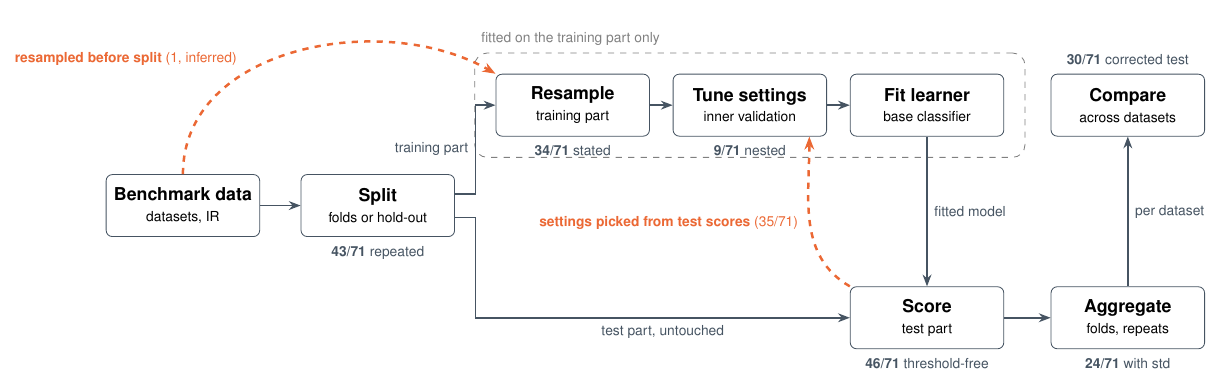}
\caption{\src{E}{7.8, Fig. 5}Where resampling, tuning, and testing belong in one run of an experiment, with the number of the 71 studies that follow each practice.}
\Description{\src{A}{S}A flow diagram of one run of an experiment: the data are split into training and test parts; resampling and the choice of settings happen inside the training part only; the model is then scored once on the test part. Each step carries the number of the 71 studies that follow it, and two dashed arrows mark the faulty practices: resampling before the split (one study, inferred) and settings picked from test scores (35 of 71).}
\label{fig:pipeline}
\end{figure}

\src{E}{7.8}Stratified $k$-fold cross-validation is the norm: 33 studies repeat it, 16 run it once, and 4 use $5\times2$ cross-validation. In total, 43 of the 71 studies (61\%) average over repeated partitions. Resampling must be fitted on the training part of each split only (Figure~\ref{fig:pipeline}); only 34 of the 71 studies (48\%) state or describe unambiguously that it was, and 36 do not say. The 15-year review recommended a validation scheme that prevents partition-induced dataset shift, DOB-SCV \parencite{lopez2014importance}; none of the 71 studies uses it.

\src{E}{7.8}Choosing settings is the weakest step in this literature. Only 9 studies (13\%) tune inside the training data, by nested cross-validation or on a disjoint set of tuning datasets. Twenty (28\%) use fixed or default settings for every method. In 35 studies (49\%), the reported configuration of the proposed method, or the amount of resampling, was chosen by looking at results on the evaluation data. The remaining 7 do not say how settings were chosen. This practice has not declined: its share rose from 36\% of the studies before 2010 to 54\% since 2021. Choosing a model on the data that are then used to estimate its performance biases the estimate upwards \parencite{cawley2010overfitting}, and the bias grows with the number of configurations tried, so it favors whichever method was searched most. Search budgets are often unequal: G-SMOTE is selected from 294 configurations against 2 for SMOTE \parencite{douzas2019gsmote}.

\subsection{Statistical Testing and Reporting}
\label{sec:testing}

\src{E}{7.9}Fifty-two studies (73\%) use a significance test, and the share without any test fell from 64\% before 2010 to 17\% since 2021. The Wilcoxon signed-rank test (26 studies) and the Friedman test (25) dominate. Grouped by what they control, 27 studies apply a corrected multiple-comparison procedure, 19 use only pairwise tests without correction although they compare the proposal with a median of six competitors, 3 stop at an omnibus test, 3 use Bayesian signed-rank tests, and 19 report no test. A significant test does not remove the selection bias described above, because the tests are run on the same folds on which settings were chosen.

\src{S}{}The form of the results tables limits what a reader can check. \src{E}{7.10}Fifty-eight studies give some table of results per dataset \src{A}{S}(43 of the 57 papers that use SMOTE as a baseline pair it with the proposed method dataset by dataset; Section~\ref{sec:checkable}), \src{E}{7.10} 42 mark the best value in bold, 38 give average ranks, and 32 a table of $p$-values. Standard deviations are given by only 24 studies, so in most tables a reader cannot tell whether a bold value differs from the next one by more than the variation between runs. Twenty-seven studies release code.

\subsection{Why Evaluation Practice Changed}
\label{sec:whychanged}

\src{S}{}The practices that spread were those that a tool made cheap and that reviewers came to expect. \src{E}{7.11}Significance testing is the clearest case: the guide of \textcite{demsar2006statistical} gave a complete recipe, KEEL implemented it, and reviewers began to expect it; 19 studies cite it for their tests. Multi-dataset benchmarks spread in the same way after KEEL made dozens of pre-partitioned imbalanced datasets available: KEEL data first appear in 2013 and are used by 67\% of the studies since 2021, and studies with fewer than ten real datasets fell from 64\% to 4\%. Code release and the modern learners followed scikit-learn and imbalanced-learn \parencite{lemaitre2017imbalanced}.

\src{S}{}A well-known recommendation did not spread without such support. \src{E}{7.11}PR-AUC is recommended for imbalanced data by a paper with more than 5,000 citations \parencite{saito2015precision} and is still used by fewer than one study in ten. The equivalence of preprocessing and cost-sensitive learning \parencite{lopez2012analysis} and DOB-SCV were both restated in the 15-year review, and neither changed the comparisons that followed. And the weakest step, choosing settings on evaluation data, became more common. \src{S}{}None of these practices is the default of a tool or the requirement of a benchmark.

\src{E}{7.11}Nor are careful studies visibly rewarded with citations (OpenAlex counts of 18 October 2026; the same ten studies are the most cited in the snapshot of 10 November 2026). Read together with the age of the papers, most differences in citation counts between studies that follow a practice and studies that do not are differences of age. The ten most cited studies use a median of 8 real datasets, only 4 of them use any significance test, and 7 were published before 2010. \src{S}{}Practice improved in the areas that tools and reviewers supported, and not in the choice of settings or of baselines.

\subsection{What the Comparisons Conclude}
\label{sec:conclude}

\src{E}{7.12}Of the 69 studies that propose a method, 68 report it as the best overall: 27 with no stated exception and 41 with an exception they state themselves. Only one reports no advantage over simple baselines \parencite{yadav2025ciso}, and that study, a preprint, is also the only one that registered its protocol in advance. \src{S}{}Because 68 of the 69 studies report their own method as the best, the claim does not distinguish between methods. \src{E}{7.12}These wins do not add up to a ranking: a method that leads in one study is placed below the proposal in every later study that includes it.

\begin{figure}[tp]
\centering
\includegraphics[width=0.98\linewidth]{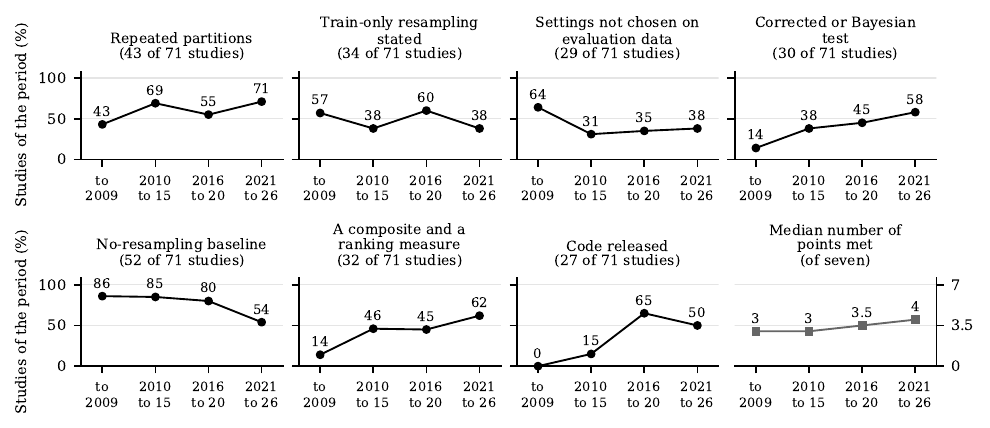}
\caption{\src{S}{}The seven-point minimum standard by period: the share of the studies of each period that meet each point, and the median number of points met. The periods hold 14, 13, 20, and 24 studies (Section~\ref{sec:extractions}). The online appendix (Section~S13) gives the counts.}
\Description{\src{S}{}Eight small line charts over the four periods. Seven show the share of studies that meet one point of the standard: corrected or Bayesian tests, a composite with a ranking measure, and code release rise; a no-resampling baseline and settings not chosen on evaluation data fall; repeated partitions and train-only resampling move up and down. The eighth shows the median number of points met, from 3 to 4.}
\label{fig:standard}
\end{figure}

\src{S}{}To summarize how far the evaluations are from the recommendations of this section, we scored every study against seven points that are cheap to meet and that a reviewer can check from the paper (Figure~\ref{fig:standard}). \src{E}{7.12}The median study meets 3 of the seven points, and the median rose only from 3 before 2010 to 4 since 2021. One study meets all seven \parencite{koziarski2019rbo}, and 39 meet three or fewer. \src{S}{}Figure~\ref{fig:standard} shows the share of studies that meet each point in each period. The score summarizes practice across the corpus; we do not use it to rank single studies.

\section{Does It Work? Independent Evidence and Theory}
\label{sec:evidence}

\src{A}{S}Sections~\ref{sec:data} and~\ref{sec:evaluation} describe what the papers that propose methods report about themselves: 79.0\% of the decided printed comparisons favor the proposal over SMOTE, and 68 of 69 method papers conclude that their own method is the best. This section asks what studies with no method to defend, and theory, say about the more basic question: whether creating synthetic minority examples helps, and when. Most of the material is from 2018 or later and was not available to the 15-year review. \src{S}{}We found the independent studies reviewed here by searching the web.

\subsection{Independent Benchmarks}
\label{sec:benchmarks}

\src{A}{R}The largest experimental comparison of the family is that of \textcite{kovacs2019empirical}, who implemented 85 minority oversamplers and ran them on 104 imbalanced datasets with four learners under one protocol, and released the implementations as an open-source package \parencite{kovacs2019smotevariants}. Its design answers a weakness documented in Section~\ref{sec:evaluation}: every method is run by a party with no stake in the ranking.

\src{A}{R}The benchmark ranks the oversamplers by their mean rank over four scores (AUC, G-mean, F1, and P20) obtained with four learners ($k$-NN, a support vector machine, a decision tree, and a multilayer perceptron). Two of its results matter for this survey. \src{S}{}First, the best methods are close to one another: the mean AUC of the ten best-ranked methods lies between 0.901 and 0.904, and their mean G-mean between 0.862 and 0.871 \parencite[Table~4]{kovacs2019empirical}, as in the documentation of the package \parencite{kovacs2019ranking}. Second, most of them are outside our corpus: the ten are polynom-fit-SMOTE, ProWSyn, SMOTE-IPF, the method of Lee and colleagues, SMOBD, the G-SMOTE of \textcite{sandhan2014gsmote} (a different method from the G-SMOTE of our corpus), CCR, LVQ-SMOTE, Assembled-SMOTE, and SMOTE followed by Tomek links. \src{S}{}Two of these ten are in our corpus, SMOTE-IPF and SMOTE-Tomek, and both remove examples after generation. Our guide (Section~\ref{sec:whichmethod}) arrives at ten methods of its own: SMOTE-family methods that are simple (at most two decisions of the SMOTE rule changed), keep every original training example, and do no worse than SMOTE in our rerun. The two lists of ten have no method in common. \src{A}{R}The benchmark also reports classification without oversampling, as a baseline. Averaged over the 104 datasets, it is lower than SMOTE for every learner and every score: for the area under the ROC curve, for example, 0.857 against 0.900 with the support vector machine and 0.812 against 0.881 with the decision tree, and the author concludes that oversampling is a reasonable way to improve classification of imbalanced data \parencite[Table~3]{kovacs2019empirical}. Plain SMOTE itself is in none of the top lists: for each learner, the mean AUC of the best oversampler exceeds that of SMOTE by 0.003 to 0.009, a gain that the author puts at about one percent and finds smaller than expected \parencite[Table~3 and p.~7]{kovacs2019empirical}. The benchmark compares averages rather than counting wins per dataset, and it includes neither a strong ensemble learner nor class weighting or a tuned threshold, which are the comparisons at issue in the studies below.

\src{A}{R}Three later studies test the premise instead of the variants. \textcite{elor2022smote}, in a preprint, compare balancing schemes with strong tabular learners next to the weaker learners of earlier studies and conclude that balancing helps the weak learners but does not improve prediction performance for the strong ones; they also identify the settings in which balancing does help, and observe that earlier studies demonstrated its utility by focusing on those settings. \textcite{sakho2026rebalancing} compare ten rebalancing strategies, including deep generative and diffusion models, on 13 tabular datasets and find that, for most datasets, applying no rebalancing strategy is competitive with LightGBM, tuned random forests, or logistic regression; only when the imbalance ratio is artificially increased does one of their SMOTE modifications lead to promising results. \textcite{tantithamthavorn2020impact} show for software defect prediction that the effect of rebalancing depends on the evaluation measure and the classifier, and that it changes which features a model treats as important.

\src{A}{S}Our own benchmark points the same way (Appendix~\ref{app:benchmark}). SMOTE beats no resampling in 83.5\% of dataset-learner pairs by G-mean but in 49.7\% by ROC-AUC and 34.3\% by PR-AUC (Section~\ref{sec:measures}). With a random forest or gradient boosting, compared over 34 source families, SMOTE does not differ from no resampling in ROC-AUC after Holm correction, and with the random forest no resampling has the higher PR-AUC (median difference 0.006). No resampling also gives the lower Brier score, and a decision threshold tuned on the training folds, without any resampling, beats SMOTE in G-mean (median differences of 0.039 and 0.040; Table~\ref{tab:benchcore}). None of the ten other oversamplers beats SMOTE in ROC-AUC or PR-AUC with either learner, and class weights do not differ from SMOTE in any of the four measures reported in Table~\ref{tab:benchcore}. When one setting of each learner is tuned by nested cross-validation, the same holds for the tuned random forest and gradient boosting, except that Geometric SMOTE then beats SMOTE in PR-AUC with the random forest. The one corpus study that includes class weighting as a baseline, and the only one with a registered protocol (a preprint), finds its new oversampler no better than class weighting, random oversampling, or no resampling \parencite{yadav2025ciso}. \src{S}{}And the printed wins over SMOTE are much less consistent on precision and specificity than on AUC, G-mean, F-measure, and recall (Figure~\ref{fig:winsmetric}).

\src{A}{S}A benchmark posted as a preprint in 2026 compares more than 40 methods for class imbalance on 57 tabular datasets under one protocol and finds that no method dominates across settings: the best strategy depends on the regime of the data, and resampling, SMOTE included, is more competitive on multiclass tasks than on binary ones \parencite{liu2026tilbench}.

\subsection{Theory}
\label{sec:theory}

\src{A}{R}In 2018, there was little theory of SMOTE beyond results such as those on high dimension discussed below. There is more now, and some of it bears on the design choices of the family. \textcite{elreedy2019comprehensive} and \textcite{elreedy2024theoretical} derive the probability distribution of the points that SMOTE generates and show that it does not in general conform to the minority class distribution, with a discrepancy that depends on the dimension, the number of minority examples, and the number of neighbors $k$. \textcite{sakho2026rebalancing} prove two properties of the SMOTE density: with its default parameter SMOTE tends to copy the original minority samples asymptotically, and its density vanishes near the boundary of the support of the minority distribution. The second property is consistent with, though it does not establish the value of, seed rules that concentrate generation near the class border (Section~\ref{sec:bydecision}), and the first bears on the adaptive choice of $k$ that the 2002 paper left open and that no method in our corpus takes up (Section~\ref{sec:methodview}). \textcite{ahmad2025concentration} give a uniform concentration bound on the gap between the empirical risk over synthetic minority samples and the population risk on the true minority distribution, and an excess risk guarantee for kernel-based classifiers trained on such data. \textcite{xia2026imbalance} treat imbalanced classification as transfer learning under label shift and show that the cost of transfer for SMOTE dominates that of random oversampling in moderately high dimensions.

\src{A}{R}High dimension is the setting in which theory and experiment agree most clearly. \textcite{blagus2013smote} show that on high-dimensional data SMOTE does not change the class-specific mean values while it decreases the variability and introduces correlation between samples; it does not attenuate the bias toward the majority class for most classifiers and is less effective than random undersampling, except for $k$-NN after variable selection. \src{S}{}Our data audit shows that the published claims rarely concern such data: half of the dataset entries that print a size have nine or fewer features (Section~\ref{sec:landscape}). The claims should not be extended to high-dimensional data. \src{A}{S}In the five high-dimensional datasets of our benchmark, SMOTE beat no resampling in 24 of the 30 dataset-learner pairs by G-mean but in 9 by PR-AUC, the pattern of the core tier (Appendix~\ref{app:benchmark}).

\subsection{Probability Calibration}
\label{sec:calibration}

\src{S}{}The strongest independent evidence is on probability calibration, which no method paper in our corpus measures. \src{A}{R}A classifier trained on balanced data learns a class prior that is not the prior of the population, so its predicted probabilities for the minority class are too high. \textcite{vandengoorbergh2022harm} show this for logistic regression: random undersampling, random oversampling, and SMOTE yielded poorly calibrated models that strongly overestimated the probability of the minority class, did not give higher areas under the ROC curve than no correction, and improved the balance between sensitivity and specificity only as much as shifting the probability threshold did. \textcite{carriero2025harms} extend the result to machine-learning algorithms: in all of their simulation scenarios, models developed without a correction for class imbalance had equal or better calibration, and the miscalibration could not always be repaired by recalibration. \textcite{piccininni2024understanding} derive the relation between the probabilities targeted with and without random resampling and give a plug-in correction. Studies on real clinical data agree: \textcite{yang2024impact} developed and externally validated 1,566 prediction models on four observational health databases and found that random oversampling and undersampling generally did not raise the AUROC and led to overestimated risks, and \textcite{andersen2026tipping} report, across ten clinical datasets with more than 600,000 patients, that SMOTE and random resampling gave no systematic improvement in discrimination and consistently worse calibration. Two 2026 preprints reach the same conclusion on a national surgical dataset of more than 1.8 million patients \parencite{roesler2026class} and in a resampling study on a large clinical trial dataset \parencite{sirikul2026class}.

\src{A}{S}This evidence is consistent with the mechanism of Section~\ref{sec:measures}. \src{S}{}Resampling moves the operating point of a classifier, and G-mean and recall reward that movement. A tuned threshold achieves the same movement and leaves the probabilities as they are. \src{A}{S}None of the 71 empirical studies of our corpus reports a calibration measure. A search of the full texts of all 76 corpus papers and their supplements for the Brier score, expected calibration error, calibration plots, and reliability diagrams found none used to evaluate a method; one study uses the Brier score only inside its algorithm, as a measure of how hard an example is to classify \parencite[p.~844]{liu2020spe}, and a blind second coding of 15 of the 71 studies found no calibration measure either.

\subsection{Leakage}
\label{sec:leakage}

\src{P}{overview}A second body of evidence concerns how oversampling is applied. \textcite{santos2018cross} show that oversampling before cross-validation inflates results, because synthetic points generated from test examples enter the training folds. \textcite{vandewiele2021overly} found more than 20 studies of preterm birth prediction whose near-perfect scores collapsed once the data were split before oversampling.

\src{A}{S}Half of the method papers show how careful they are on this point: 34 of the 71 studies state that resampling was confined to the training data, 36 do not say, and in one the reported numbers of synthetic samples imply that the whole dataset was oversampled before cross-validation (Section~\ref{sec:protocol}). The risk may be larger in the applied literature, where SMOTE is a preprocessing step that is easy to place before the split, as the preterm-birth studies above suggest. \src{S}{}Section~\ref{sec:harm} returns to leakage and to calibration as documented harms.

\subsection{Beyond Tables: Deep and Generative Oversampling}
\label{sec:beyondtables}

\src{A}{R}The independent evidence above is for tabular data and for learners that take a fixed vector of features. Two lines of work carry the question further: imbalance in deep networks, where it is studied as long-tailed recognition, and generative models that learn the minority distribution instead of interpolating between examples. Neither belongs to the SMOTE family as defined in Section~\ref{sec:intro}; we describe them only to place the family among its successors.
%<<search:d_deep
\src{A}{S}The account starts from the screen of our search (Section~\ref{sec:search}): of the records it includes, 561 propose such a generative oversampler (as coded from title and abstract), 13 published before 2018, 130 from 2018 to 2022, and 418 since 2023 (Section~\ref{sec:beyond}). We add the reviews and benchmarks on which those records rely.
%>>search:d_deep

\src{A}{R}In deep learning, the same problem appears as long-tailed recognition, with many classes of which most have few examples, and the remedies that dominate there act on the loss or on the classifier rather than on the data \parencite{johnson2019survey,zhang2023deeplt}. Losses reweight the classes by their effective number of examples \parencite{cui2019classbalanced} or weight down the easy examples \parencite{lin2020focal}; margins grow for rare classes \parencite{cao2019ldam}; the logits are adjusted by the class priors \parencite{menon2021logit}; or the representation is learned on the natural distribution and only the classifier is retrained on class-balanced samples \parencite{kang2020decoupling}. The last result has the same shape as the tabular evidence of Section~\ref{sec:calibration}: rebalancing acts mainly on the decision rule, and the features are learned at least as well without it. Resampling survives in other forms: replication of minority images for convolutional networks \parencite{buda2018systematic}; SMOTE-style interpolation in a learned feature space \parencite{dablain2023deepsmote}; translation of majority examples into minority ones \parencite{kim2020m2m}; and convex combinations of pairs of examples and of their labels \parencite{zhang2018mixup}, an interpolation between two examples, as in our definition, put to another use, in a form rebalanced toward the rarer class \parencite{chou2020remix}.

\src{A}{R}For tables, the successors of the family learn the distribution of the minority class. Generative adversarial networks are represented in our corpus by the conditional tabular GAN and its autoencoder counterpart TVAE \parencite{xu2019ctgan}, and by GAMO \parencite{mullick2019gamo}, which we count in the family because its learned convex combination generalizes the segment; diffusion models \parencite{kotelnikov2023tabddpm} and language models fine-tuned to write table rows as text \parencite{borisov2023great} followed. The counts above show how fast mentions of such models grew among the records of our search. Independent evidence on them is thin, and the comparison we found does not favor them: in the comparison of ten rebalancing strategies by \textcite{sakho2026rebalancing}, which included deep generative and diffusion-based ones, applying no rebalancing with a strong learner was competitive on most of the 13 datasets. Whether a learned generator exposes the records it was trained on is a separate question from the one \textcite{ganev2026smote} answer for SMOTE, and it has to be measured for each generator.

\src{A}{R}A tabular foundation model, trained in advance on millions of synthetic datasets, predicts the labels of a new table from its training rows in a single forward pass; on datasets of up to 10,000 samples, its authors report that it outperforms all earlier methods, including an ensemble of strong baselines tuned for four hours \parencite{hollmann2025tabpfn}. Whether rebalancing helps such a model is only beginning to be studied. Among the records of our search, the first comparisons date from 2025 and 2026 and are mostly applications; a benchmark of binary classifiers without any rebalancing reports that the foundation model and boosted ensembles keep more of their performance than traditional classifiers as the minority class shrinks \parencite{nawaz2025beyond}. \src{S}{}If that holds, the question this survey asks of SMOTE (whether oversampling adds anything to a strong learner with a tuned threshold) becomes more pressing.

\src{A}{S}Several of the six decisions of Section~\ref{sec:bydecision} reappear outside tables. Allocation becomes the sampling distribution over classes in a training batch, which ranges from instance-balanced through square-root to class-balanced sampling \parencite{kang2020decoupling}. Neighbor choice and placement become the choice of an interpolation partner and of a mixing weight: mixup takes the partner from any class and the weight from a beta distribution \parencite{zhang2018mixup}; Manifold Mixup moves the interpolation into a hidden layer of the network \parencite{verma2019manifold}; DeepSMOTE moves it into the latent space of an encoder-decoder network and keeps the neighbors in the minority class, as SMOTE does \parencite{dablain2023deepsmote}. The step at which rebalancing enters a pipeline becomes a training schedule: the papers that proposed deferring re-weighting to a late stage of training \parencite{cao2019ldam} and retraining only the classifier on balanced samples \parencite{kang2020decoupling} report that both work better than rebalancing from the start. \src{S}{}The label of a new point, which SMOTE always sets to the minority class, becomes a decision of its own: mixup interpolates the labels too, and Remix shifts them toward the rarer class \parencite{chou2020remix}. The generator, too, becomes a model learned from the data.

\src{A}{S}Evidence from comparisons, rather than from the papers that propose a method, is thinner here than for tables. In deep long-tailed recognition, a systematic comparison of the common training tricks (re-weighting, re-sampling, mixup, and two-stage training), which also adds an augmentation of its own, found that the tricks can conflict, so that in the wrong combination they lower accuracy \parencite{zhang2021bagoftricks}; and networks trained on long-tailed data are more miscalibrated and overconfident than networks trained on balanced data, two-stage rebalanced ones included \parencite{zhong2021mislas}, a deep counterpart of Section~\ref{sec:calibration}. For tables, besides the comparison of rebalancing strategies cited above \parencite{sakho2026rebalancing}, a preprint that applied the classical corrections for imbalance to a tabular foundation model (TabPFN-2.5) found that, for binary tasks, moving the decision threshold worked best among thresholding, downsampling, oversampling, and synthetic upsampling, because of the calibration of the model's probabilities \parencite{mcdowell2026correcting}.

%<<search:j_tabpfn
\src{S}{}Section~\ref{sec:whichmethod} reports a rerun of our own, in which ten SMOTE-family methods and four baselines run under one protocol on the 42 tasks of our reference collection (Appendices~\ref{app:reference} and~\ref{app:benchmark}). We added a tabular foundation model to this rerun: TabPFN-3.5, the default model of version 10.0.0 of the \texttt{tabpfn} package, with its default settings, without resampling and after SMOTE, on the same folds. \src{A}{S}It ran on the graphics processor of the Apple M5 Pro of the core tier and gave a result in every fold of 38 of the 42 tasks, in 28 families. On musk and smoke detection, it ran out of the processor's memory, and on ijcnn1 and one ecoli task, SMOTE could not run in every fold; under the rule fixed before the results, every comparison uses the other 38 tasks. SMOTE made the model worse on all three measures: over families, it lowered ROC-AUC by a median of 0.017 and PR-AUC by 0.032, and raised the Brier score by 0.012 (Holm-adjusted Wilcoxon $p < 0.02$; Bayesian probabilities that the model is better without it 0.98, 1.00, and 0.82). Without resampling, the model ranked better than the random forest and gradient boosting of the rerun, with or without SMOTE, by medians of 0.014 to 0.028 in ROC-AUC and 0.024 to 0.046 in PR-AUC (Holm-adjusted $p < 0.03$; Bayesian probability at least 0.95 that the model is better). Its Brier score was lower than that of gradient boosting, with or without SMOTE, and of the random forest after SMOTE, by 0.013 to 0.027, and on par with that of the random forest without resampling (online appendix, Section~S7). Each learner ran with one configuration, and the cells of the random forest and gradient boosting are those of the rerun, 36 of the 38 tasks from the cloud workspace, so these comparisons mix two machines on the same folds.
%>>search:j_tabpfn

\src{A}{S}We found no comparison that sets learned generators against SMOTE and against no resampling, with strong learners and with probability-based measures and calibration, on many datasets; each study above tests part of that design. Whether resampling helps a tabular foundation model beyond what a moved threshold gives rests so far on one preprint \parencite{mcdowell2026correcting}. And no study changes the deep counterparts of the six decisions one at a time, which is what the attribution problem of Section~\ref{sec:agenda} asks of the family as well.

\src{A}{S}The evidence outside tables is of a different kind, and the position of Section~\ref{sec:position} does not extend to it.

\subsection{Reading the Evidence Together}
\label{sec:together}

\begin{table}[tp]
\centering\small
\caption{\src{A}{S}What the evidence supports. ``Corpus'' refers to Sections~\ref{sec:family} to~\ref{sec:evaluation}; ``independent'' to the studies of this section.}
\label{tab:claims}
\begin{tabularx}{\linewidth}{L{4.4cm}YL{2.4cm}}
\toprule
Claim & Evidence & Status \\
\midrule
Oversampling raises recall and G-mean of weak learners at the default threshold & Corpus (printed comparisons); our benchmark (SMOTE better than no resampling in 83.5\% of pairs by G-mean); independent benchmarks & Supported \\
It improves ranking (ROC-AUC, PR-AUC) & Our benchmark: SMOTE better than no resampling in 49.7\% and 34.3\% of pairs, no difference in ROC-AUC with strong learners; independent studies with strong learners find no gain, while \textcite{kovacs2019empirical} report a higher mean AUC with SMOTE than without resampling for each of their four classifiers & Not supported in general \\
It beats class weighting or a tuned threshold & Almost never tested in the corpus (class weighting in 1 of 71 studies); in our benchmark, class weights do not differ from SMOTE with strong learners and a tuned threshold beats it in G-mean; independent studies find no advantage & Not supported \\
New variants beat SMOTE & 79.0\% of decided printed comparisons, under each paper's own protocol; larger on noisy data; 49\% of studies choose settings on evaluation data; small margins in one independent benchmark \parencite{kovacs2019empirical}; in our benchmark, no variant beats SMOTE in ROC-AUC or PR-AUC with strong learners, except Geometric SMOTE in PR-AUC with a tuned random forest & Supported mainly by the proposers \\
It yields usable probabilities & Not measured in the corpus; in our benchmark, SMOTE raised the Brier score in 470 of 636 pairs; independent clinical studies find systematic overestimation of risk & Contradicted \\
It works for several classes & 13 of 37 methods are run with three or more classes kept & Thin \\
It works in high dimension & Half of the entries have nine or fewer features; theory and experiment are unfavorable; in our benchmark's high-dimensional tier, SMOTE gains G-mean (24 of 30 pairs) but rarely PR-AUC (9 of 30) & Not supported \\
\bottomrule
\end{tabularx}
\end{table}

\src{A}{S}Table~\ref{tab:claims} sets the claims made for synthetic oversampling against the evidence. \src{S}{}For a higher recall or G-mean with a weak learner at its default threshold, the evidence is solid and has been since 2002. For the stronger claims (a better ranking, an advantage over cheaper remedies, usable probabilities, a better method than SMOTE), the evidence comes mostly from the proposers of the methods, and the little independent support is for weaker learners or by small margins (Section~\ref{sec:benchmarks}). \src{A}{S}Our data audit adds one qualification in favor of the variants: their printed wins over SMOTE are more frequent on data whose minority examples are borderline or isolated (Section~\ref{sec:wins}), a setting that some variants were designed for, although the association was measured over all proposals and is exploratory. A common benchmark that varies this property under one protocol would test whether the effect is real.

\section{Whom Did It Reach, and with What Documented Benefit and Harm?}
\label{sec:society}

\src{A}{R}Surveys of methods rarely ask whom a technique reached and what it did for them, and the question needs a definition before it can be answered. We distinguish four levels at which a technique can be said to have affected society, in increasing order of the evidence they require. \emph{Reach}: who cites it, in which fields and countries. \emph{Use}: in which kinds of decisions it is applied. \emph{Outcome}: whether those decisions became better because of it. \emph{Harm}: what it made worse. Bibliometric data can establish the first level, the literature can describe the second, and only studies designed for the purpose can establish the third and fourth. \src{A}{S}It is the least mature part of the survey: the three audits of Sections~\ref{sec:family} to~\ref{sec:evaluation} were designed for other questions, and the answer given here rests on one bibliometric query, a random sample of applications, a search of PubMed for harms, and studies from outside the corpus.

\subsection{Reach}
\label{sec:reach}

\begin{figure}[tp]
\centering
\includegraphics[width=\linewidth]{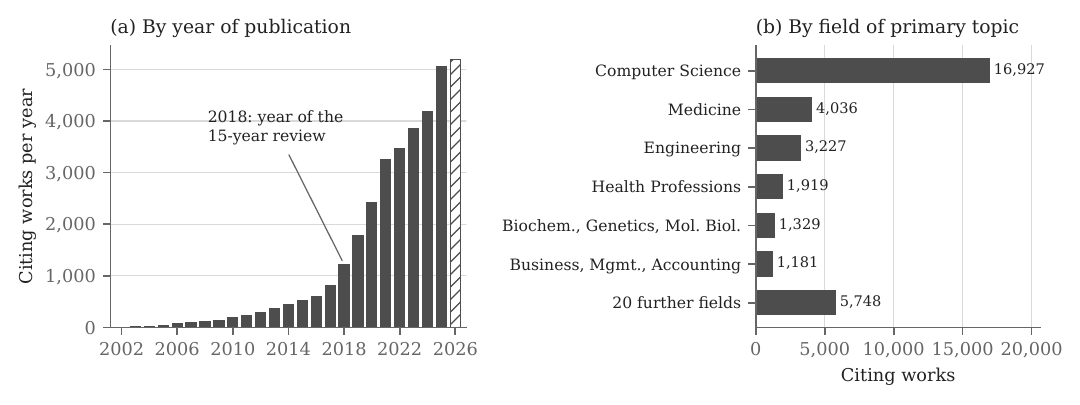}
\caption{\src{R}{}Scholarly reach of SMOTE in OpenAlex on 10 November 2026: 34,416 indexed works cite the 2002 paper. (a) Citing works by year of publication from 2002 (six works that OpenAlex dates before 2002 are left out); the hatched bar is the incomplete year 2026. (b) Citing works by the field of their primary topic (34,366 works with a topic): the six largest fields and the other 20.}
\Description{\src{A}{S}Panel (a) is a bar chart of citing works per year from 2002 to 2026, rising slowly to about 800 in 2017 and then steeply to more than 5,000 in 2025 and 2026; the 2026 bar is hatched because the year is incomplete. Panel (b) is a horizontal bar chart by field: Computer Science 16,927, Medicine 4,036, Engineering 3,227, Health Professions 1,919, Biochemistry, Genetics and Molecular Biology 1,329, Business, Management and Accounting 1,181, and 20 further fields 5,748.}
\label{fig:reach}
\end{figure}

\begin{table}[tp]
\centering\small
\caption{\src{A}{R}Works citing the SMOTE paper in OpenAlex (queried 10 November 2026; 34,416 works). Sustainable Development Goal tags are assigned by an OpenAlex classifier, and a work can carry several or none.}
\label{tab:reach}
\begin{tabularx}{\linewidth}{L{3.4cm}Y}
\toprule
Grouping & Citing works \\
\midrule
Year of publication & up to 2017: 3,994 (11.6\%); 2018 to 2026: 30,414 (88.4\%); more than 5,000 a year in 2025 and 2026 \\
Domain of primary topic & Physical Sciences 22,425; Health Sciences 6,021; Social Sciences 2,997; Life Sciences 2,924 \\
Field of primary topic & Computer Science 16,927 (49.3\%); Medicine 4,036; Engineering 3,227; Health Professions 1,919; Biochemistry, Genetics and Molecular Biology 1,329; Business, Management and Accounting 1,181; 20 further fields 5,748 \\
Sustainable Development Goal tag & Good health and well-being 10,764; Industry, innovation and infrastructure 2,648; Reduced inequalities 2,408; Sustainable cities and communities 961; Quality education 809; Zero hunger 706; Climate action 510; Affordable and clean energy 508 \\
Country of an author's institution & 159 countries and territories; China 7,899; United States 6,077; India 3,088; United Kingdom 1,906; Canada 1,307; South Korea 1,284 \\
Document type & Article 21,166; conference paper 9,367; preprint 1,403; dissertation 907; book chapter 823; review 422 \\
\bottomrule
\end{tabularx}
\end{table}

\src{A}{R}The reach of SMOTE is wide and still growing (Figure~\ref{fig:reach}, Table~\ref{tab:reach}). OpenAlex indexed 34,416 works citing the 2002 paper on 10 November 2026. Almost nine in ten of them (88.4\%) were published in 2018 or later, that is, after the 15-year review was written, and the yearly number has passed 5,000. Half of the citing works with a primary topic (50.7\%) are in a field other than computer science. Health is the largest area outside computer science among the citing works: 6,021 citing works have a primary topic in the health sciences, and 10,764 carry the tag of the third Sustainable Development Goal, good health and well-being. Authors from 159 countries and territories cite the paper, and the list is not confined to countries with large research budgets: Indonesia, Brazil, Turkey, Saudi Arabia, Bangladesh, Malaysia, Iran, and Pakistan are among the 25 countries with the most citing works. The 907 dissertations among the citing works suggest a further role, in research training, which citation counts cannot measure.

\subsection{An Enabling Infrastructure}
\label{sec:infrastructure}

\src{A}{R}Part of this reach coincides with the availability of software. SMOTE is simple enough to implement in a few lines, has one main parameter, needs no model of the data, and works with any classifier. It was available early in Weka and KEEL (Section~\ref{sec:learners}) and is now implemented in imbalanced-learn, the scikit-learn-compatible toolbox of \textcite{lemaitre2017imbalanced}, whose repository showed 7.1 thousand stars on GitHub and which was downloaded about 5.7 million times from the Python Package Index in the 30 days to 9 November 2026 (pypistats.org, accessed 10 November 2026; download counts include automated installs). The open implementation of 85 variants by \textcite{kovacs2019smotevariants} did the same for the family. Our evaluation audit points to the other side of this infrastructure: in the corpus, the change of learners coincides with the change of software, and practices without a tool behind them did not spread (Section~\ref{sec:whychanged}).

\subsection{Use}
\label{sec:use}

\src{A}{S}Several papers of the corpus evaluate on application data, which shows some of the domains in which the methods were tried. The 2002 paper itself used mammography and oil-spill data; later corpus papers use network intrusion records (Cluster-SMOTE), time series (INOS), bearing-fault data from machinery (SVDD-WSMOTE), software defect data (GDHS), customer churn and employee attrition data (AB-SMOTE), and fraud and record-linkage data (SPE). Private or application data appear in 11 of the 71 empirical studies, mostly in the early periods (36\% before 2010, 8\% since 2021); the method papers of the corpus have moved toward public benchmarks and away from applications.

\src{A}{R}Outside the corpus lies the applied literature, which accounts for an unmeasured share of the 34,416 citing works (Section~\ref{sec:reach}). \textcite{ganev2026smote} summarize it as follows: practitioners rely on SMOTE in a wide range of medical applications, including cancer diagnosis, heart disease, diabetes prediction, and genetic risk prediction; in finance, particularly for credit-card fraud detection and the prediction of customer churn; and, increasingly, as a generator of synthetic tabular data for data sharing and as a baseline against which generative models are judged.

%<<search:e_use
\src{A}{S}To see how SMOTE is used outside the method literature, we audited the applications that the screen of Section~\ref{sec:search} excluded under E1: 14,738 records that apply existing resampling to a problem of their own. We drew 300 of them at random, 100 from each of three periods (up to 2020, 2021 to 2023, and since 2024, which hold 1,198, 2,543, and 10,997 records). \src{A}{S}AI sessions coded each from its title and abstract under a written codebook (online appendix, Section~S5), and the first 50 of each period in the order of the draw also from the full text, of which 116 could be obtained from open copies or through the authors' university library. \src{A}{S}The 5 records that this coding took to propose a resampling method of their own were coded again under the rulings of Section~\ref{sec:search}, by which a standard generator, or a chain of existing steps that the paper does not compare with other resampling, is existing resampling. Of the 300 records, 289 are indeed applications. \src{A}{S}A second coding of all 300 records by separate AI sessions, blind to the first and from the same texts, agreed with it on 93.2\% of the values coded for every record (Cohen's kappa 0.54 to 0.97 by field, lowest for the kind of validation) and on 96.4\% of the safeguards coded from the 116 full texts. An AI session settled the 185 differences from the records, seeing the two codes in random order; 67 of them changed a first code, and the counts of this section use the settled codes. \src{S}{}The authors checked 86 of the settled codes against the records: every change that can alter a count of this section, and random samples of the other changes, of the codes the adjudication kept, and of the codes both coders gave, and found all of them correct.

\src{A}{S}Medicine is the largest domain: 106 of the 289 applications (37\%; 41.0\% with the periods weighted by their size, 95\% interval 33.5 to 48.5) build a model for diagnosis, prognosis, or another clinical or public-health decision. Finance follows (27), then social and text data (24), security (22), and software engineering (17). In 137 applications (47\%), the model is meant to inform a decision about a person, such as a patient, a borrower, a student, or a customer. Plain SMOTE is the resampler in 190 (66\%), 188 work on tabular data, and 76 train a deep network.

\src{A}{S}The full texts show how the safeguards of Section~\ref{sec:evaluation} fare in practice. \src{S}{}Of the 116 full texts obtained, 110 are applications, and the counts of this paragraph are of these 110. Among them, 41 state that the resampling was fitted on the training data only, 46 resampled the whole data set before splitting it, which is the leak of Section~\ref{sec:leakage}, and 23 do not say; with the periods weighted by their size, an estimated 37.0\% of the applications resample before the split (95\% interval 25.6 to 48.4). \src{A}{S}The leak shows no trend across the periods: it appears in 11 of 29 full texts up to 2020, 22 of 42 from 2021 to 2023, and 13 of 39 since 2024, although more papers since 2024 state that they fitted the resampling on the training data (21 of 39, against 10 of 29 and 10 of 42 before). In 46, the reported results were computed on test data that had themselves been resampled, so the scores do not describe the natural class distribution; for this code and the previous one, arithmetic on the numbers a paper reports counts as evidence (ruling D of the authors). The same model trained without resampling is a comparator in 55, class weights or a tuned threshold in 9, and calibration is assessed in 2. Accuracy is reported in 92, ROC-AUC in 48, and PR-AUC in 2. Data from another source validate the model in 6, a prospective evaluation appears in none, and results from deployment in none. \src{S}{}The other 34 papers drawn for full-text coding could not be obtained and may differ.
%>>search:e_use

\subsection{Outcome}
\label{sec:outcome}

\src{A}{R}We found no study, in the corpus or in our searches, that measures an outcome-level benefit of SMOTE: a prospective evaluation or a deployed system in which decisions made with a model trained on SMOTE-resampled data were compared with decisions made without it. \src{S}{}The studies that exist are retrospective. Many report higher recall, F1, or G-mean after resampling; Sections~\ref{sec:evaluation} and~\ref{sec:evidence} explain why such gains are expected from a moved operating point and do not by themselves show a better model. A small number of independent studies in clinical prediction test the premise with measures closer to utility, and they find no consistent gain in discrimination and a loss in calibration \parencite{vandengoorbergh2022harm,carriero2025harms,andersen2026tipping}; one preprint adds a decision-curve analysis in which the model trained on the natural class distribution has the highest net benefit \parencite{roesler2026class}.

\src{S}{}The benefit of SMOTE to society is plausible and widely assumed, and no study we found demonstrates it at the level of outcomes. The evidence supports three benefits. First, SMOTE gave practitioners in many fields a free, simple, and model-agnostic way to make a classifier pay attention to a rare class. Second, it is associated with a change in how such classifiers are evaluated: the 2002 paper argued that accuracy cannot be used under imbalance, and 39 of the 71 studies of our corpus still give that argument for their choice of measures (Section~\ref{sec:measures}). Third, it founded a research program, including the theory, the benchmarks, and the critiques reviewed in Section~\ref{sec:evidence}, whose findings now improve practice whether or not they favor SMOTE.

\subsection{Harm}
\label{sec:harm}

%<<search:g_harm
\src{A}{S}Four harms are documented. To find studies that measure a harm of oversampling directly, we searched PubMed on 10 November 2026 with one query for each harm that the literature names (calibration, leakage, privacy, and fairness between groups; online appendix, Section~S6) and screened the 429 distinct records on title and abstract under a written rule: a study counts when it compares a model trained on oversampled data with one trained without resampling, or, for leakage, resampling before the split with resampling of the training data only, on a measure of the harm. The screen was done by AI sessions with a single screener per record (Section~\ref{sec:ai}). The rule admits 27 studies, and 49 records need the full text. Of the 14 studies that measure calibration, 9 find it worse after oversampling, 4 find no harm, and 1 finds mixed effects; all 8 that measure leakage find estimates inflated by resampling before the split; of the 5 that compare a disparity between groups, 4 find no harm, and 1 finds mixed effects; and none of the 34 records of the privacy query compares an attack on oversampled data with one on the original data. Another 176 records report the calibration of a model trained on oversampled data without any such comparison.
%>>search:g_harm

\src{A}{R}Models trained on resampled data overestimate the probability of the rare event (Section~\ref{sec:calibration}). Where a predicted probability informs a decision about an individual, as in clinical risk prediction, this can cause direct harm, and the distortion is expected to grow with the imbalance that resampling is meant to correct (Section~\ref{sec:calibration}). Most of the comparative studies that the harm search found, nearly all of them in health care, find this loss of calibration in their own data. The reporting guideline for clinical prediction models that use machine learning now addresses it: item 13 of TRIPOD+AI asks authors who used class imbalance methods to state why and how, and to report any subsequent recalibration \parencite{collins2024tripodai}.

\src{A}{R}Oversampling before the data are split leaks information from the test set, and at least one application literature, on preterm birth prediction, has reported near-perfect results for this reason (Section~\ref{sec:leakage}). Every comparative study of leakage that the harm search found, in radiomics, clinical prediction, and other fields, finds the same inflation, and Section~\ref{sec:use} finds resampling before the split among the applications it read in full. \src{S}{}Those who rely on these published numbers work from inflated estimates.

\src{A}{R}Because each synthetic point lies on a segment between two real minority records, SMOTE-augmented and SMOTE-generated datasets expose those records. \src{S}{}\textcite{ganev2026smote} give two attacks that need only the released dataset and the knowledge that SMOTE produced it. That knowledge tells the attacker how the data were built: each synthetic record lies on the segment between a real minority record and one of its $k$ nearest minority neighbors, so the real records are the endpoints of these segments ($k$ and the imbalance ratio can be estimated from the released data). The attacks assume that no three real records lie on one line. Whenever three records of the released data lie on one line, the middle one is therefore synthetic. The first attack uses this to separate the real from the synthetic records of an augmented dataset, with perfect precision and recall. \src{A}{R}The second groups the records of a synthetic dataset into lines of three or more collinear points and keeps the points where at least three such lines meet; a synthetic record lies on one such line only, so every point kept is a real record. It reconstructs real minority records with perfect precision and a recall of 0.85 on average, which reaches one at imbalance ratios of 20 or more. The records most exposed are those of the minority, who in health and finance are often the people the model is about. The harm search found no comparative study of privacy among the PubMed records it could decide from the abstract, so the evidence for this harm comes from the attacks rather than from applications.

\src{A}{R}\textcite{hassanat2022stop}, a critical review in our corpus whose journal version is \textcite{tarawneh2022stop}, and a follow-up on medical datasets \parencite{hassanat2023jeopardy} report that synthetic points frequently lie closer to held-back majority examples than to minority examples. The direct checks of Section~\ref{sec:directchecks} exist to detect this, and only 6 of 71 studies use one. \src{A}{S}The queries of the harm search cover the other three harms and fairness between groups; this one concerns the data rather than a decision, and its evidence is this review and its follow-up.

\src{A}{R}Two further questions are open. Whether oversampling helps or harms fairness across demographic groups is not settled. Resampling by protected attribute has been proposed as a way to reduce bias in learned models \parencite{chakraborty2021bias}, and the comparative studies of fairness that the harm search found, all in health care, report smaller disparities after oversampling except one, which finds mixed effects; miscalibration that differs by group would work in the other direction, \src{A}{S}and the harm search covered fairness in PubMed only. And 2,408 citing works carry the tag of the goal ``reduced inequalities'', which an OpenAlex classifier assigns; what they do with SMOTE is the second question.

\subsection{A Balance}
\label{sec:balance}

\begin{table}[tp]
\centering\small
\caption{\src{A}{R}Benefits and harms of SMOTE to society, with the level and strength of the evidence found in this survey.}
\label{tab:balance}
\begin{tabularx}{\linewidth}{L{4.6cm}YL{2.3cm}}
\toprule
Benefit or harm & Evidence & Strength \\
\midrule
An accessible remedy for rare classes, adopted in many fields & 34,416 citing works, half outside computer science, from 159 countries; implemented in common toolboxes & Strong (reach) \\
A change in how classifiers for rare classes are evaluated & Corpus: accuracy is rarely the deciding measure; 39 of 71 studies cite its failure under imbalance & Moderate (within research) \\
A research program (variants, theory, benchmarks, critiques) & This survey & Strong \\
Better decisions in applications & Retrospective, self-reported gains on threshold measures; no outcome-level study found; independent clinical studies find no consistent gain & Weak, contested \\
Harm: overestimated risk & Simulation and real clinical data, several independent groups; worse calibration after oversampling in 9 of the 14 comparative studies of the harm search & Strong \\
Harm: inflated published results through leakage & Documented in specific application literatures; inflated estimates in all 8 comparative studies of the harm search & Strong \\
Harm: exposure of minority records & One peer-reviewed study with constructive attacks; no comparative study in PubMed & Moderate, recent \\
Harm: synthetic points that are not minority examples & One critical review in the corpus and a follow-up on medical data, by one group; direct checks in 6 of 71 studies & Limited (one group) \\
\bottomrule
\end{tabularx}
\end{table}

\src{A}{R}Table~\ref{tab:balance} summarizes the section. Twenty-five years on, SMOTE's contribution to society is real at the level of reach, method, and research culture, and unproven at the level of outcomes; its documented harms are specific and avoidable. Most of them are associated with using SMOTE as a default preprocessing step where a tuned threshold, a class weight, or no correction might have served as well, and with evaluating the result with measures that reward the shift it produces. The guidance of Section~\ref{sec:guidance} is written with these harms in mind.

\section{Guidance for Users and Authors}
\label{sec:guidance}

\subsection{Before Oversampling at All}
\label{sec:before}

\src{A}{S}The evidence of Sections~\ref{sec:evaluation} to~\ref{sec:society} suggests four questions to ask before any oversampler is chosen.

\begin{enumerate}
\item \src{A}{S}\emph{Is a probability needed, or only a label?} If the model's predicted probability informs a decision, do not resample; if resampling is used nevertheless, recalibrate on data with the original class proportions and report calibration (Section~\ref{sec:calibration}).
\item \src{A}{S}\emph{Has the cheap remedy been tried?} Train the strongest available learner without resampling, with class weights, or with a decision threshold tuned on validation data. Oversampling is worth its cost only if it beats this baseline on a measure fixed in advance (Sections~\ref{sec:baselines} and~\ref{sec:benchmarks}).
\item \src{A}{S}\emph{Is the resampler inside the training folds?} Resampling, scaling, and feature selection must be fitted on the training part of each split only (Sections~\ref{sec:protocol} and~\ref{sec:leakage}).
\item \src{A}{S}\emph{Will the resampled data leave the organization?} SMOTE-augmented or SMOTE-generated data expose the real minority records and should not be treated as anonymous (Section~\ref{sec:harm}).
\end{enumerate}

\subsection{Which SMOTE-Family Method to Use}
\label{sec:whichmethod}

\src{M}{10}A reader who needs to oversample a training set, and not to design a new method, can use the survey to choose among the 37 methods. \src{S}{}We select with three criteria: C1, performance on par with SMOTE; C2, a simple method; and C3, no loss of data. C2 and C3 can be read from the coding of the methods, so we apply them first; C1 needs an experiment and is judged last, on the methods that pass the other two. \src{M}{10}\emph{C2, a simple method}: the method changes at most two of the six decision points, does not run inside a learner, and does not learn its generator; 13 of the 37 methods qualify. \emph{C3, no loss of data}: the method keeps every original training example and stays within the SMOTE-family definition; 23 of the 37 qualify. Ten methods meet C2 and C3 (Table~\ref{tab:recommend}). \src{S}{}\emph{C1, performance on par with SMOTE}, is judged on one protocol for all ten (below), because the papers' own numbers leave it open: in their own per-dataset comparisons six of the ten, including SMOTE, do at least as well as plain SMOTE (median $\Delta \ge 0$ in Table~\ref{tab:methods}), but of the 23 papers that report a comparable difference from SMOTE, 17 lie within $\pm 0.03$ of it, and the differences come from each paper's own protocol and metric.

\begin{table}[tp]
\centering\scriptsize\setlength{\tabcolsep}{3.5pt}
\caption{\src{A}{S}SMOTE-family methods that are simple and keep all data (C2 and C3). Paper: the difference from SMOTE that each method's own paper prints ($\Delta$ as in Table~\ref{tab:methods}; plots, avg.: no per-dataset comparison with SMOTE). Code: the library used in the rerun (imblearn: imbalanced-learn; smote-var.: smote-variants; imblearn-extra: imbalanced-learn-extra), or ours, with the result of its validation against a number its paper prints (repr.: reproduced; online appendix, Section~S7). Rerun: median difference from SMOTE over families of tasks of the reference collection, with the random forest (RF) and gradient boosting (GB); C1 is met when no entry is below $-0.01$. The baselines of Section~\ref{sec:before} are shown for comparison. Class: B binary only, M multiclass only, B+M both.}
\label{tab:recommend}
% Generated by scripts/gen_recommend_table.py from benchmark/results_summary/rerun42/; do not edit by hand.
\begin{tabular}{@{}lll>{\raggedright\arraybackslash}p{1.75cm}lll>{\raggedright\arraybackslash}p{1.9cm}@{}}
\toprule
 & & Paper: & & \multicolumn{2}{l}{Rerun: $\Delta$ vs SMOTE, RF/GB} & & \\
Method & Class & $\Delta$ vs SMOTE & Code & ROC-AUC & PR-AUC & C1 & Decisions changed \\
\midrule
SMOTE (2002) & B & ref. & imblearn & ref. & ref. & ref. & none \\
Cluster-SMOTE (2006) & B & plots & smote-var. & .000/.000 & .000/$-$.002 & yes & partner \\
ADASYN (2008) & B & +.015 (5/5) & imblearn & .000/.000 & .000/$-$.001 & yes & allocation \\
Safe-Level-SMOTE (2009) & B & plots & smote-var. & .000/.000 & .000/.000 & yes & seed, placement \\
CGMOS (2016) & B & +.009 (24/30) & ours, not repr. & .000/.000 & .000/.000 & yes & seed \\
SMOM (2017) & M & +.004 (19/27) & ours, repr. & .000/.000 & .000/.000 & yes & partner \\
G-SMOTE (2019) & B & avg. & imblearn-extra & .000/.000 & .000/.000 & yes & partner, placement \\
AB-SMOTE (2020) & B & $-$.029 (2/8) & ours, repr. & .000/.000 & .000/.000 & yes & seed, partner \\
RSMOTE (2021) & B+M & +.021 (21/27) & ours, not repr. & .000/.000 & .000/.000 & yes & seed, allocation \\
FW-SMOTE (2022) & B & +.018 (30/42) & ours, close & .000/.000 & .000/.000 & yes & partner \\
\midrule
No resampling & & & & .000/.000 & .000/.000 & & \\
Class weights & & & & .000/.000 & .000/.000 & & \\
Tuned threshold & & & & .000/.000 & .000/.000 & & \\
Random oversampling & & & & $-$.001/$-$.003 & .000/.000 & & \\
\bottomrule
\end{tabular}

\end{table}

%<<search:h_rerun
\src{A}{S}The rerun on the reference collection of Appendix~\ref{app:reference} applies the protocol of the independent benchmark, with the baselines of Section~\ref{sec:before}, to all ten methods (Appendix~\ref{app:benchmark}; online appendix, Section~S7). A method meets C1 when its median difference from SMOTE over families of tasks is no worse than $-$0.01, the region of practical equivalence of our tests, in ROC-AUC and in PR-AUC with each of the two strong learners, the random forest and gradient boosting; the rule was fixed before any result was read. Five of the ten methods come from the libraries of the benchmark. The other five had no usable public implementation, and AI sessions implemented each from its paper and validated it against a number the paper prints (Section~\ref{sec:ai}): SMOM and AB-SMOTE reproduce theirs; FW-SMOTE comes close, but its paper's gain over SMOTE appears only when its configuration is chosen on the test folds; RSMOTE and CGMOS do not reproduce their papers' results, although for RSMOTE our runs reproduce the paper's own values for SMOTE and for no resampling, and CGMOS as its paper defines it chooses its seeds almost as SMOTE does (a total-variation distance of at most 0.0025 from SMOTE's uniform choice on five datasets). We run all five as their papers describe them, and Table~\ref{tab:recommend} says which reproduced. The rerun covers all 42 tasks of the collection, in 31 families.

\src{A}{S}No variant differs from SMOTE beyond that region. With both strong learners, the median difference from SMOTE over the 30 families in which SMOTE ran is at most 0.002 in ROC-AUC and in PR-AUC for every variant, and Holm-corrected Wilcoxon tests find none of these 36 comparisons significant, so every variant meets C1. The methods do differ on single tasks (within a family the ten methods span a median of 0.017 in ROC-AUC with the random forest, and more than 0.01 in 20 of the 30 families), but no variant does better or worse than SMOTE consistently enough to move the median. \src{S}{}The variants differ from SMOTE in operating point and calibration: AB-SMOTE and FW-SMOTE, which by their papers' defaults add fewer points than SMOTE, give a lower G-mean at the default threshold and a lower Brier score, and Safe-Level-SMOTE gives a higher Brier score with the random forest (online appendix, Section~S7). \src{S}{}The baselines repeat the finding of the core tier. No resampling has the same median ROC-AUC and PR-AUC as SMOTE. A median difference of zero does not mean that the two agree in every family: in PR-AUC with gradient boosting, no resampling is lower than SMOTE by more than 0.01 in 10 of the 30 families and higher by more than 0.01 in 5, and the Bayesian signed-rank test, which takes the size of every family's difference into account, gives a probability of 0.43 that SMOTE is better by more than 0.01. No resampling also gives the lower Brier score (by 0.005 with the random forest and 0.003 with gradient boosting), and a threshold tuned on the training folds gives a higher G-mean than SMOTE (by 0.041 and 0.024).

\src{A}{S}On this evidence, the ten simple methods are interchangeable in the measures that do not depend on a threshold, and none is a reason to prefer it to plain SMOTE, which remains the reference that every variant is measured against. A variant can still be chosen for its mechanism (fewer synthetic points, a different allocation or partner), but then it should be checked on the data at hand with a strong learner, a tuned threshold, and calibration reported (Section~\ref{sec:checklist}). The gains that RSMOTE and CGMOS print in their papers are not reproduced, and FW-SMOTE's only with settings chosen on test folds, so those gains should not be cited as evidence for the methods. SMOM is the only Multiclass-only method that meets C2 and C3, and the rerun, which is two-class, tests it on two classes only; the simplest alternative is not a method paper of its own: decomposing the problem one-vs-one and applying plain SMOTE to each pair of classes \parencite{fernandez2013analysing}.
%>>search:h_rerun

% v0.64 (the team lead, 23 Nov 2026): 24 -> 23 and 18 -> 17 after KNNOR was marked avg. like G-SMOTE, CISO and QC-SMOTE
% v0.2 (WP1 step 6): 26 -> 24, 19 -> 18 and seven -> six after RBBoost and MC-MBRC were marked not comparable (Data rule); the Method PDF still prints 26, 19 and seven
\src{M}{10}The six methods with the largest reported gains (median $\Delta$ above 0.03) all fail C2 or C3: they run inside a learner, learn their generator, change four or more decisions, or remove or move original examples. Four of them were also compared with SMOTE inside a wrapper and not with plain SMOTE. \src{S}{}These six methods are also the most complex, so their gains are the hardest to attribute and to reproduce.

\subsection{A Reporting Checklist}
\label{sec:checklist}

\src{A}{S}Each of the three audits ends with its own recommendations: eight for reporting data, five for describing a new variant, and an eleven-step protocol for evaluation. Table~\ref{tab:checklist} merges them into one checklist of fifteen items, each traced to the finding that motivates it.

\begin{table}[tp]
\centering\footnotesize
\caption{\src{A}{S}The SMOTE25 checklist for a study that proposes or compares a SMOTE-family method. Audit: the audits whose recommendations the item merges; Finding: the section that reports the finding behind it.}
\label{tab:checklist}
\begin{tabularx}{\linewidth}{rYL{2.2cm}L{2.5cm}}
\toprule
 & Item & Audit & Finding \\
\midrule
\multicolumn{4}{l}{\emph{The method}} \\
1 & State which of the six decision points the method changes, and keep the others as in SMOTE unless there is a reason not to & Method & Section~\ref{sec:bydecision} \\
2 & When more than one decision changes, report a comparison that changes one decision at a time & Method & Section~\ref{sec:methodview} \\
3 & State the class setting in which the generator is run, whether any multiclass dataset was reduced to two classes and how, and how nominal features are handled & Method; Data & Section~\ref{sec:classsetting} \\
\multicolumn{4}{l}{\emph{The data}} \\
4 & Name the repository and version of every dataset; state which classes form the minority and why; declare every modification and print the size and class counts of the data actually used & Data & Sections~\ref{sec:modified}, \ref{sec:samename} \\
5 & Give the imbalance-ratio formula, not only its values & Data & Section~\ref{sec:samename} \\
6 & Report the number of source families next to the number of datasets, and keep splits of one source together when testing & Data & Section~\ref{sec:landscape} \\
7 & Report at least one measure of where the minority examples lie (for example the shares of safe, borderline, rare, and outlier examples) & Data; Evaluation & Sections~\ref{sec:difficulty}, \ref{sec:wins} \\
\multicolumn{4}{l}{\emph{The experiment}} \\
8 & Use at least four learners of different kinds, including a strong ensemble, configured identically for every resampler & Evaluation & Section~\ref{sec:learners} \\
9 & Include no resampling, class weighting or threshold moving, random oversampling, and plain SMOTE as baselines, all re-run on the same folds; do not copy numbers from other papers & Evaluation; Data & Sections~\ref{sec:baselines}, \ref{sec:checkable} \\
10 & Report a threshold composite and a ranking measure, each with its formula, positive class, threshold, and averaging rule; declare the primary measure in advance; add a calibration measure when probabilities are used & Evaluation; this survey & Sections~\ref{sec:measures}, \ref{sec:calibration} \\
11 & Repeat stratified cross-validation, fit every learned step on the training folds only, and either use recommended settings for all methods or tune all of them with the same budget by nested cross-validation & Evaluation & Section~\ref{sec:protocol} \\
12 & Test with the dataset as the unit, with a correction for multiple comparisons, and report effect sizes & Evaluation & Section~\ref{sec:testing} \\
\multicolumn{4}{l}{\emph{The report}} \\
13 & Add at least one direct check of the synthetic points on held-out data & Evaluation & Section~\ref{sec:directchecks} \\
14 & Report per-dataset results for the proposal and plain SMOTE on every metric, with standard deviations, and the running time relative to SMOTE; release code, folds, and modified data & Data; Method; Evaluation & Sections~\ref{sec:checkable}, \ref{sec:testing} \\
15 & State where the method loses, the limits of the study, and that the conclusions hold for classification & Evaluation & Section~\ref{sec:conclude} \\
\bottomrule
\end{tabularx}
\end{table}

\subsection{A Position}
\label{sec:position}

\src{A}{S}Our position, limited to tabular classification, is that the field should stop proposing new SMOTE variants as a contribution in themselves. \src{S}{}The variants re-decide the same six choices (Section~\ref{sec:family}); in our corpus, recent methods change more of them at once on average, those that report running time cost more to run, and they are judged mostly by their own authors against plain SMOTE. Authors should no longer propose an interpolation variant whose only evidence is a win over plain SMOTE in its own paper, oversample before splitting the data, report accuracy or ROC-AUC alone, or call a method multiclass when it was run only on data reduced to two classes. Plain SMOTE remains useful as a baseline, though not as a default preprocessing step, and oversampling remains useful for weak learners, where the evidence still shows gains. New studies should compare against class weighting and threshold tuning with a strong learner, report PR-AUC and a calibration measure next to ROC-AUC, change one decision point at a time, and report running time relative to SMOTE.

\src{A}{S}The position rests on what the cited studies tested: tabular data, mostly two classes, and the learners they used. \src{S}{}Two kinds of evidence support it. The first is the independent evidence of Section~\ref{sec:evidence}: a preprint \parencite{elor2022smote} and a critical review \parencite{tarawneh2022stop} that questioned oversampling in 2022, and a peer-reviewed study of 2026 in which applying no rebalancing is competitive on most datasets \parencite{sakho2026rebalancing}. The second is what the audits of Sections~\ref{sec:family} to~\ref{sec:evaluation} show about how the variants are evaluated. The position allows that oversampling helps in some settings and leaves each method to be judged on its own evidence; images, text, and learned representations call for evidence of another kind (Section~\ref{sec:beyondtables}).

\section{Open Problems and a Research Agenda}
\label{sec:agenda}

\src{A}{S}The three audits and the independent evidence point to nine problems that a next generation of work could settle.

\begin{enumerate}
\item \src{M}{9}\emph{The neighborhood size.} An adaptive choice of $k$ was named as future work in 2002 and is not the subject of any method in the corpus. \src{A}{R}Theory now says why it matters: with a fixed $k$ SMOTE tends to copy the minority sample as the sample grows \parencite{sakho2026rebalancing}.
\item \src{A}{S}\emph{Attribution.} Recent variants change 3.6 decisions on average and are compared with SMOTE as a whole. Ablations that change one decision point at a time, on a shared benchmark, would show which of the six decisions carry the gains.
\item \src{A}{S}\emph{A benchmark built from measured difficulty.} The common benchmarks are small, low-dimensional, and dependent on each other through their source families, and the printed wins over SMOTE rise with the noise index. A benchmark that reports source families, varies the share of borderline and outlying minority examples, includes large and high-dimensional tasks, and keeps multiclass data multiclass would test the claims that the present benchmarks cannot.
\item \src{A}{S}\emph{Baselines that do not create data.} Class weighting appears in one of 71 studies. Whether any SMOTE-family method beats a strong learner with a tuned threshold is the central empirical question of the field, and it is almost untested by the papers that propose the methods.
\item \src{A}{R}\emph{Calibration-aware oversampling.} Resampling distorts predicted probabilities, and recalibration does not always repair the damage. Whether a generator can improve the ranking of minority examples without changing the learned prior, or with a correction that is known in closed form, is open.
\item \src{E}{7.7}\emph{Measuring the synthetic data.} The 15-year review named the relation between synthetic instances and the intrinsic characteristics of the data as the main open question. Six studies check the resampled data directly, with six different checks. A common, classifier-free measure of where synthetic points fall is still missing.
\item \src{A}{R}\emph{Privacy.} Interpolation exposes minority records. Oversampling with a stated privacy guarantee, and privacy evaluations that do not rely on distance to the closest record, are needed before SMOTE-generated data are shared.
\item \src{E}{7.3.1}\emph{Beyond classification and beyond tables.} All 71 studies evaluate through classification, and three of the 37 methods generate non-tabular data. Regression, streams, multi-label learning, and learned representations each need their own evidence.
\item \src{A}{R}\emph{Outcome-level evidence.} No study we found measures whether decisions improved because of SMOTE. Prospective or external evaluations in the fields that cite it most outside computer science, starting with health, would answer the question that Section~\ref{sec:society} has to leave open.
\end{enumerate}

\section{Limitations}
\label{sec:limitations}

\src{E}{7.14}The corpus is curated. \src{A}{S}It was chosen for its relevance to SMOTE, so its shares describe this corpus. The systematic search that tests its coverage was run in Scopus, OpenAlex, and arXiv and screened by AI sessions (Section~\ref{sec:search}). It shows that the corpus covers the literature up to 2009 closely and the literature since 2016 thinly, and that generative oversamplers, which the corpus places beyond the family, have become a large part of it (Section~\ref{sec:beyond}). \src{S}{}A comparison with the reference lists of five recent reviews, with the cited works checked by the authors, finds 136 of the 176 method papers that they cite; the 40 others are mostly generative, mixup, and graph methods whose titles carry no pair of terms of the search string, so the search undercounts that branch (Appendix~\ref{app:search}). \src{A}{S}Whether the findings of the corpus hold in that literature is tested on a random sample coded from the full texts that could be obtained.

\src{S}{}The periods are small. They hold 13 to 24 studies each, so a single study changes a period's share by 4 to 8 percentage points, and we call a difference a trend only under the rule of Section~\ref{sec:extractions}. \src{A}{S}The causes we give for changes in practice rest on direct or circumstantial evidence or are conjectures, as stated in each case, and all of that evidence is observational. \src{E}{7.14}Citation counts depend on age, venue, and topic as much as on the evaluation.

\src{M}{3.4}The extraction was assisted. The per-method accounts and class categories come from AI readers of the PDFs under fixed rules; \src{S}{}the evaluation coding was done by AI sessions, with a second reading that recorded a page and a quote for every code; \src{S}{}a blind second pass by separate AI sessions measured the stability of the codes and an AI session adjudicated its disagreements, so most codes were checked by AI passes and by the authors only in samples (Section~\ref{sec:ai}); \src{D}{1.5}and part of the difficulty measurement rests on copies of the repositories and on the assumption that a public file was used as distributed.

\src{M}{10}Differences from SMOTE come from the papers themselves. Each is computed under its own paper's metric, learner, and protocol; they cannot be compared across papers and are not a benchmark.

\src{A}{S}The benchmark has limits of its own. It uses the default settings of every sampler and learner, except for the one setting per learner tuned in its third tier; its 110 datasets are the two-class sets bundled with the imbalanced-databases package, many of them small, and SMOTE could not run on four of them; and it holds the methods with a public implementation that runs, so five methods discussed in this survey (CGMOS, SMOM, RSMOTE, FW-SMOTE, and AB-SMOTE) enter only the rerun of Section~\ref{sec:whichmethod}, implemented from their papers.

\src{A}{R}The account of reach, benefit, and harm is mostly bibliometric. It rests on one database queried on one day, on labels assigned by classifiers, and on independent studies concentrated in clinical prediction. The harm search covers PubMed only, was screened on title and abstract by AI sessions with a single screener per record, \src{A}{S}and its four queries cover calibration, leakage, privacy, and fairness between groups. The account establishes reach, describes use from a random sample of applications, and reports the outcome and harm evidence we found.

\src{S}{}\emph{The three extractions count some things differently}; the text uses one definition for each quantity. For example, the Data extraction counts 73 papers with an evaluation and the Evaluation extraction 71, because the first keeps two papers that the second treats as non-empirical: the paper on Tomek links, whose only data are a synthetic illustration, and a critical review. The online appendix (Section~S11) lists every such quantity and how each difference was settled.

\section{Conclusion}
\label{sec:conclusion}

\src{S}{}Table~\ref{tab:answers} gives the answers of this survey to the five questions of Section~\ref{sec:intro}.

\begin{table}[tp]
\centering\small
\caption{\src{S}{}The five research questions and what the survey finds.}
\label{tab:answers}
\begin{tabularx}{\linewidth}{L{2.4cm}YL{2.3cm}}
\toprule
Question & What the survey finds & Evidence \\
\midrule
RQ1, methods & Each of the 36 variants re-decides some of the six decisions of the SMOTE rule, 2.8 of them on average, and 24 of the 37 methods generate data for two classes only. & Section~\ref{sec:family} \\
RQ2, data & The evidence rests on 1,756 dataset entries, 347 of which the papers' authors modified; 990 entries can be matched with a public file. & Section~\ref{sec:data} \\
RQ3, evaluation & The median study meets 3 of seven minimum points; 35 of the 71 studies choose their settings on test scores and one compares with class weighting. The practices that spread were those that a tool made cheap and that reviewers came to expect. & Section~\ref{sec:evaluation} \\
RQ4, effect & In their own papers the proposed methods win 79.0\% of the decided comparisons with SMOTE. Independent studies find that the gains shrink with strong learners and tuned thresholds and that resampling damages calibration; in our rerun, no variant differs from SMOTE by more than 0.002 in median ROC-AUC or PR-AUC with a random forest or gradient boosting. & Sections~\ref{sec:wins}, \ref{sec:evidence} and~\ref{sec:whichmethod} \\
RQ5, reach & 34,416 indexed works cite SMOTE, half of them outside computer science. A benefit at the level of outcomes is undemonstrated, four harms are documented, and an estimated 37\% of applications resample the whole data set before splitting it. & Section~\ref{sec:society} \\
\bottomrule
\end{tabularx}
\end{table}

\src{S}{}SMOTE fixed six decisions in the simplest way, and most of the methods we analyzed change several of them. \src{A}{S}The 37 methods we analyzed change the seeds, the allocation, the partner, the placement, the filter, and the pipeline in ever larger combinations, mostly for two classes, and they are tested on small tables that are hard to identify and are related to each other. Their authors almost always find them better than SMOTE, under protocols that half of the time let the evaluation data choose the settings, and with measures that reward the shift of the decision threshold that any rebalancing produces. Independent studies, most of them with strong learners or in clinical prediction, indicate that this shift accounts for much of what synthetic oversampling does: it is valuable for weak learners that must output a label, and of little use, or harmful, where a strong learner, a tuned threshold, or a calibrated probability is available.

\src{A}{S}The technique reached far beyond the field that produced it: by November 2026 more than 34,000 indexed works cited it, half of them outside computer science and many in health. \src{S}{}A benefit at the level of outcomes has not been shown, and four harms are documented: miscalibrated risk, results inflated by leakage, the exposure of minority records, and synthetic points that are not minority examples.

\src{S}{}The 2002 paper gave researchers and practitioners in many data-intensive fields a way to act on rare classes, argued against accuracy as the measure of such models, and started the line of work whose latest results inform this survey. The field now needs better evidence: a description of each method by the decisions it changes, data that can be identified, baselines that do not create data, measures that include calibration, and studies of whether the decisions made with these models improved.

\src{A}{S}For tabular classification, we therefore take the position of Section~\ref{sec:position}: a new SMOTE variant should no longer count as a contribution in itself. The next useful contribution is a generator that has been compared, one decision at a time, with class weighting and a tuned threshold on a strong learner, with its calibration and its running time reported, or evidence that the decisions made with these models improved.

\src{S}{}The survey is also a worked example of a review carried out with AI agents under the authors' written rules. Table~\ref{tab:aisteps} lists each step with its check, and we release the instructions and the records of the checks with the data.

\section*{Declarations}

\src{S}{}\textbf{Use of AI tools.} The authors used AI agents (Claude Opus 5.5, Anthropic), working under the authors' written rules, to extract and code information from the papers of the corpus, to run the database searches, to screen the records and apply the authors' rulings, to code the two samples, to check extracted values and references against their sources, to implement five methods from their papers, to write and run the benchmark experiments, and to prepare figures and tables. The authors used a second model of the same family (Claude Fable 5.2) to check the screen, to compare the records of the search with the reference lists of five reviews, and to assist in editing the text. Section~\ref{sec:ai} and Table~\ref{tab:aisteps} give each step with its check and its result. After the last AI pass, the authors read, edited, and approved every sentence of the paper. \src{A}{S}The authors take full responsibility for the content.

\src{A}{S}\textbf{Data and code availability.} The data and code behind this survey are deposited on Zenodo \parencite{intharah2026smote25data} and, with the same files, in a GitHub repository \parencite{intharah2026smote25repo}: \url{https://doi.org/10.5281/zenodo.24212321} and \url{https://github.com/tohnperfect/smote25-deposit}. The data are licensed under CC BY 4.0 and the code under the MIT license. The deposit holds the coding sheets of the three extractions, the scripts that build every table and figure of the paper and of the online appendix, the search strings and request logs, the identifiers of every screened record with its decisions and codes, the OpenAlex snapshot, the codebooks, and the code and per-fold results of the benchmark and of the rerun on the reference collection. The online appendix (Section~S4.4) lists them, and the material that the authors keep outside the deposit: the abstracts of the records, the full texts of the papers, and the passages quoted from them. \src{A}{S}The authors also keep the results of the rerun's TabPFN-3.5 arm (Section~\ref{sec:beyondtables}) outside the deposit.

\src{A}{S}\textbf{Funding.} This research received no specific funding.

\src{A}{S}\textbf{Author contributions.} Supawit Chaingam led the method extraction and wrote the first draft of its part of the paper; Kanvara Thawarorit led the data extraction and wrote the first draft of its part of the paper; Pharamut Makjaroen led the evaluation extraction and wrote the first draft of its part of the paper. \src{S}{}Thanapong Intharah conceived and supervised the study, administered the project, orchestrated the coding, the experiments, and the visualizations, combined the three section drafts into one paper and edited it before submission, using the AI tools described above to assist in the editing. \src{A}{S}All authors approved the final version.

\src{A}{S}\textbf{Competing interests.} The authors declare no competing interests.

\printbibliography

%% Appendices are lettered (A, B, ...).
\appendix
\section{How Each Method Builds on SMOTE}
\label{app:accounts}

\src{A}{S}Section~S1 of the online appendix gives, for each of the 37 methods, an account of how its generator relates to Equation~\ref{eq:smote}, the decisions it changes, and the evidence for its class category, grouped by class category and ordered by year, with page, section, table, or equation references. \src{S}{}The authors used AI sessions to write the accounts from a full reading of each paper (Section~\ref{sec:ai}); the accounts are generated from the coding sheet by script, and their class category, year, and decisions changed agree with Table~\ref{tab:methods}. \src{S}{}Five of the accounts settle a point that a first reading left open: Loyalty-SMOTE is Binary only and GLOS Multiclass only, NROMM keeps the Core link, and OREM with its multiclass and boosting versions stays one entry; SMOM was first coded from its supplementary file, and its journal version confirms its class category (Multiclass only), its link (Core), and the one decision it changes (the partner), while its per-dataset scores are printed in the supplementary file only.

\src{S}{}The 15-year review \parencite{fernandez2018smote} lists about 90 extensions of SMOTE (Section~\ref{sec:corpus}). \src{M}{3.1}Each extension is characterized by seven properties (pp.~869--871): initial selection of instances, integration with undersampling, type of interpolation, dimensionality change, adaptive generation, relabelling, and filtering of noisy generated instances. Four of these correspond to our decision points (initial selection to seed selection, adaptive generation to sample allocation, type of interpolation to the placement rule and partly to neighbor choice, filtering to post-filtering); the other three act on the data around the generator, and generation inside a pipeline appears in that review as a separate table of ensembles.

\src{A}{R}The corpus overlaps only in part with the two largest lists of the family. Of the 17 SMOTE-family methods of the corpus published from 2003 to 2017, ten appear in the tables of the 15-year review (eight among the 90 extensions of its Table~1 and two among the 17 ensembles of its Table~2); seven do not (Cluster-SMOTE, LS, INOS, RBBoost, SCUT, CGMOS, and SMOM). \src{A}{S}That review's Table~1 also lists MDO, which our definition excludes, and a G-SMOTE \parencite{sandhan2014gsmote} that is a different method from the G-SMOTE of our corpus. \src{A}{R}Eleven of our 37 methods are among the oversamplers implemented and compared by \textcite{kovacs2019empirical}: SMOTE, SMOTE-Tomek and SMOTE-ENN, Borderline-SMOTE, Cluster-SMOTE, ADASYN, Safe-Level-SMOTE, MWMOTE, SMOTE-IPF, A-SUWO, k-means SMOTE, and DSRBF. Of the other 26, 18 were published in 2018 or later.

\section{Coding of the 71 Empirical Studies}
\label{app:coding}

\src{S}{}Section~S2 of the online appendix lists the main codes of every empirical study: setting, number of real datasets and source, learners, measures, protocol, how settings were chosen, test, code release, and the number of points met of the seven-point standard of Figure~\ref{fig:standard}. \src{A}{S}The full coding sheet gives a page reference for every code; the deposit holds it with these references, and the authors keep the passages quoted from the papers, which are excerpts of their text.

\section{A Reference Collection of Verifiable Two-Class Tasks}
\label{app:reference}

\src{A}{S}For readers who need a small benchmark that covers the measured range, we select 42 of the 238 verifiable tasks by a fixed rule on the difficulty flags and imbalance-ratio bands of Section~\ref{sec:difficulty} (online appendix, Section~S3). \src{D}{1.5}Each combination of flag profile and imbalance-ratio band that occurs (19 combinations) contributes the task that the corpus uses most, for comparability with earlier results, and the tasks with the lowest and the highest noise index, for spread. The collection spans noise indices from 0 to 1, imbalance ratios from 1.1 to 970.6, and sizes from 79 to 62,630 instances, and 13 of its files match the official KEEL table. Its tasks are not independent of one another: it holds four yeast splits, three pima versions, and two each of ecoli, glass, haberman, cleveland, heart, and iris, which should be grouped by source when results are tested.

\section{Independent Benchmark}
\label{app:benchmark}

\src{A}{S}The benchmark follows the protocol of Section~\ref{sec:protocol}. Its core tier uses the 110 two-class datasets bundled with the imbalanced-databases package, most of them from KEEL \parencite{alcala2011keel}, which fall into 36 source families. It compares 14 ways of handling the imbalance: no resampling; class weights (\texttt{class\_weight="balanced"}, for every learner that accepts them, which excludes $k$-NN); a decision threshold chosen on the training part, as the one that maximizes G-mean over out-of-fold probabilities of an inner 3-fold cross-validation; random oversampling; SMOTE; and nine SMOTE-family methods with public implementations. SMOTE, ADASYN, Borderline-SMOTE, and random oversampling come from imbalanced-learn 0.14.2 \parencite{lemaitre2017imbalanced}, Geometric SMOTE from imbalanced-learn-extra \parencite{douzas2019gsmote}, and Safe-Level-SMOTE, Cluster-SMOTE, $k$-means SMOTE, MWMOTE, SMOTE-IPF, and A-SUWO from smote-variants 1.0.1 \parencite{kovacs2019smotevariants}, each with its implementation's default settings. CGMOS, SMOM, FW-SMOTE, and AB-SMOTE have no public implementation that we could find, and the code its authors published for RSMOTE imports two modules that are not in its repository; these five enter only the rerun of Section~\ref{sec:whichmethod}, where AI sessions implemented them from their papers. Six learners are trained with fixed settings: CART, $k$-NN ($k=5$), logistic regression, an SVM with an RBF kernel and Platt scaling, a random forest of 200 trees, and histogram gradient boosting. Performance is estimated by five repetitions of stratified 5-fold cross-validation with the same folds for every method and learner; scaling, resampling, and threshold choice are fitted on the training folds only. Every fold is scored by ROC-AUC, PR-AUC, G-mean, F1, MCC, balanced accuracy, the Brier score, and the expected calibration error. One seed (20261110) fixes the folds and every sampler and learner that draws random numbers. The runs took place on an Apple M5 Pro (15 cores, 24 GB) under macOS 26.5.2 with Python 3.13 and scikit-learn 1.9.1, with 14 parallel workers.

\src{A}{S}Fold results are averaged per dataset, and datasets of one source family are averaged, so that each family counts once. For each learner and measure, a Friedman test with the Iman-Davenport correction tests for any difference among the methods, and every method is compared with SMOTE by a Wilcoxon signed-rank test with Holm correction and by a Bayesian signed-rank test with a region of practical equivalence of 0.01 \parencite{benavoli2017time}. SMOTE, Borderline-SMOTE, and Geometric SMOTE could not run on four datasets whose training folds hold five or fewer minority examples, nor on 10 of the 25 folds of a fifth, and ADASYN failed on these and on all or some folds of eight more datasets. Two of the four datasets are the only ones of their source family, so the comparisons with SMOTE use the 34 families where it ran. Of the 231,000 cells of the core tier (dataset, fold, method, and learner), 225,112 gave a result; the rest are 3,138 runs in which a sampler stopped with an error (SMOTE, Borderline-SMOTE, and Geometric SMOTE 660 each, ADASYN 1,152, A-SUWO 6) and the 2,750 cells of $k$-NN with class weights, which it does not take.

\src{S}{}For the core tier, Table~\ref{tab:benchcore} gives the median differences from SMOTE for the random forest and gradient boosting. \src{A}{S}With both learners, SMOTE does not differ from no resampling in ROC-AUC, no resampling gives the lower Brier score, and the tuned threshold beats SMOTE in G-mean; with the random forest, no resampling also has the higher PR-AUC. No other oversampler beats SMOTE in ROC-AUC or PR-AUC, and class weights do not differ from SMOTE in any of the four measures.

\begin{table}[tp]
\centering\footnotesize\setlength{\tabcolsep}{4pt}
\caption{\src{A}{S}Core tier of the benchmark: median difference from SMOTE over 34 source families, for a random forest (RF) and gradient boosting (GB). Positive values mean the method did better than SMOTE (for the Brier score, a lower score). ROC, PR, G: ROC-AUC, PR-AUC, G-mean. $^{*}$ Holm-adjusted Wilcoxon $p < 0.05$; a starred $-$0.000 is a significant difference whose median rounds to zero.}
\label{tab:benchcore}
\begin{tabular}{lrrrrrrrr}
\toprule
& \multicolumn{4}{c}{RF} & \multicolumn{4}{c}{GB} \\
\cmidrule(lr){2-5}\cmidrule(lr){6-9}
Method & ROC & PR & G & Brier & ROC & PR & G & Brier \\
\midrule
% generated by scripts/gen_benchmark_tables.py from benchmark/results_summary/core/tests_vs_smote.csv
No resampling & 0.000 & +0.006$^{*}$ & $-$0.046$^{*}$ & +0.003$^{*}$ & 0.000 & 0.000 & $-$0.042$^{*}$ & +0.002$^{*}$ \\
Class weights & 0.000 & 0.000 & +0.002 & 0.000 & $-$0.002 & $-$0.002 & $-$0.001 & 0.000 \\
Tuned threshold & 0.000 & +0.006$^{*}$ & +0.039$^{*}$ & +0.003$^{*}$ & 0.000 & 0.000 & +0.040$^{*}$ & +0.002$^{*}$ \\
Random oversampling & 0.000 & 0.000 & $-$0.010$^{*}$ & 0.000 & $-$0.004$^{*}$ & $-$0.001 & $-$0.007 & $-$0.001 \\
ADASYN & $-$0.000$^{*}$ & $-$0.003$^{*}$ & $-$0.001 & $-$0.000$^{*}$ & 0.000 & 0.000 & 0.000 & 0.000 \\
Borderline-SMOTE & 0.000 & $-$0.001 & 0.000 & 0.000 & 0.000 & $-$0.001 & 0.000 & 0.000 \\
Geometric SMOTE & 0.000 & +0.002 & 0.000 & 0.000 & 0.000 & +0.002 & 0.000 & +0.001 \\
Safe-Level-SMOTE & $-$0.000$^{*}$ & $-$0.001 & +0.006$^{*}$ & $-$0.002$^{*}$ & $-$0.001 & 0.000 & +0.006 & $-$0.001 \\
Cluster-SMOTE & 0.000 & 0.000 & $-$0.006$^{*}$ & +0.001 & 0.000 & 0.000 & $-$0.006$^{*}$ & 0.000 \\
$k$-means SMOTE & 0.000 & 0.000 & $-$0.020$^{*}$ & +0.001 & 0.000 & +0.001 & $-$0.014$^{*}$ & +0.001$^{*}$ \\
MWMOTE & $-$0.001$^{*}$ & $-$0.005$^{*}$ & $-$0.006$^{*}$ & 0.000 & 0.000 & 0.000 & $-$0.005$^{*}$ & +0.001 \\
SMOTE-IPF & 0.000 & 0.000 & 0.000 & 0.000 & 0.000 & +0.001 & $-$0.001 & 0.000 \\
A-SUWO & $-$0.002$^{*}$ & $-$0.003 & $-$0.030$^{*}$ & 0.000 & $-$0.005$^{*}$ & $-$0.006$^{*}$ & $-$0.021$^{*}$ & $-$0.001 \\
\bottomrule

\end{tabular}
\end{table}

\src{S}{}In the high-dimensional tier, the same protocol was run on five OpenML datasets \parencite{vanschoren2014openml} with at least 100 features, chosen by a rule fixed before the run (binary target, minority share at most 0.20, at most 20,000 rows, not already in the core tier): scene, Speech, musk, Internet-Advertisements, and one of the gene-expression targets of the OVA collection, drawn at random among those that loaded. \src{A}{S}Of its 10,500 cells, 10,351 gave a result; the others are A-SUWO runs that stopped with an error and $k$-NN with class weights. SMOTE beat no resampling in 24 of the 30 dataset-learner pairs by G-mean, in 18 by ROC-AUC, and in 9 by PR-AUC, and it raised the Brier score in 20 of them.

\src{S}{}In the nested tier, to test whether the default settings decide the result, 30 of the 59 core datasets with at least 30 minority examples (drawn with the seed 20261111; 21 source families) were run again with one setting of each of four learners tuned by an inner 3-fold cross-validation on PR-AUC, within three repetitions of stratified 5-fold cross-validation: the inverse regularization strength of logistic regression, the number of neighbors of $k$-NN, the minimum leaf size of the random forest, and the learning rate of gradient boosting, each over three values. \src{A}{S}Of its 25,200 cells, 24,578 gave a result; the others are ADASYN runs that stopped with an error and $k$-NN with class weights. With the tuned random forest and gradient boosting, SMOTE again does not differ from no resampling in ROC-AUC, no resampling gives the lower Brier score, and a tuned threshold beats SMOTE in G-mean. With the tuned random forest, no resampling also has the higher PR-AUC (median difference 0.012), and Geometric SMOTE is the only oversampler that beats SMOTE in PR-AUC (0.009). The online appendix (Section~S10) gives the medians.

%<<search:i_rerun_app
\src{S}{}In the reference-collection tier, the ten methods of Table~\ref{tab:recommend} and the four baselines were run under the same protocol on all 42 tasks of the reference collection (Appendix~\ref{app:reference}), which fall into 31 families of the Data section; five of the methods were implemented from their papers by AI sessions (Section~\ref{sec:whichmethod}; online appendix, Section~S7). \src{A}{S}A cloud workspace (Linux, two cores) ran 38 tasks with the same versions of scikit-learn and the samplers' libraries, and the Apple M5 Pro of the core tier ran promotion, shoppers, musk, and smoke detection. Mammography and ijcnn1 ran on both machines, whose processors and numerical libraries differ: 54\% and 35\% of their cells differ in at least one of ROC-AUC, PR-AUC, G-mean, and the Brier score, but no mean over the folds differs by more than 0.005, and no difference from SMOTE with the random forest or gradient boosting by more than 0.003; the tests use the cloud's copies. Rows with a missing value were dropped and nominal features one-hot encoded. Of its 88,200 cells, 86,154 gave a result; 996 are sampler runs that stopped with an error, where a training fold held too few minority examples for five neighbors (SMOTE, G-SMOTE, and ADASYN) or ADASYN found no minority example with a majority neighbor or no point to generate, and 1,050 are $k$-NN with class weights. SMOTE, G-SMOTE, and ADASYN could not run on ijcnn1, whose training folds hold too few minority examples, nor on some folds of one ecoli task; as in the core tier, a method's mean is taken over the folds where it ran, so the family of ijcnn1 has no mean for SMOTE, and the tests use the 30 families in which SMOTE ran. With the random forest and gradient boosting, no variant differs from SMOTE by more than 0.002 in median ROC-AUC or PR-AUC, no resampling gives the lower Brier score, and a tuned threshold the higher G-mean (Section~\ref{sec:whichmethod}).
%>>search:i_rerun_app

\section{Details of the Systematic Search}
\label{app:search}

\src{S}{}Section~\ref{sec:search} gives the sources, the search string, and the criteria (I1 to I3, E1 to E4) of the systematic search. \src{M}{3.2}A pilot of the identification stage on OpenAlex (29 October 2026; Table~\ref{tab:pilot}) returned 14,589 records for the final string and retrieved 36 of the 37 SMOTE-family papers; the one it misses, LS, has none of the generation terms in its OpenAlex title and abstract. Of the 75 corpus papers with an OpenAlex record, it retrieves 57; the 18 it misses are almost all papers that do not oversample. \src{A}{S}The pilot also exposed a limit of OpenAlex: it holds an abstract for only 17 of the 37 SMOTE-family papers, so the search was also run in Scopus.

\begin{table}[tp]
\centering\small
\caption{\src{M}{3.2, Table 2}Pilot of the search strings on OpenAlex (29 October 2026, 2002 to 2026). Recall: corpus papers with an OpenAlex record that each string retrieves.}
\label{tab:pilot}
\begin{tabular}{L{7.2cm}rrr}
\toprule
String & Records & SMOTE-family (37) & Corpus (75) \\
\midrule
S1: SMOTE named & 16,109 & 21 & 24 \\
S2: oversampling and imbalance & 10,853 & 32 & 52 \\
S3: S2 in the title only & 4,519 & 27 & 39 \\
S4: S2 and method words (propose, novel, algorithm) & 10,407 & 31 & 48 \\
S5: final string & 14,589 & 36 & 57 \\
\bottomrule
\end{tabular}
\end{table}

%<<search:a_screen
\src{A}{S}The records were screened on title and abstract under these criteria by an AI assistant (Claude Opus 5.5), in separate sessions of up to 200 records that followed a written codebook (online appendix, Section~S5), in three looks. A first pass decided every record and recorded a code, a confidence, and a one-sentence reason. A second look, blind to the first, screened again the 4,358 records that the first pass had marked unsure or decided with low or medium confidence; on these hard cases, the two looks agree on 81.9\% (Cohen's $\kappa = 0.72$). A third look decided the 2,088 records on which the two disagreed or either was unsure, weighing both earlier reasons, and marked 1,222 of them unsure. Two checks bear on the first pass. \src{S}{}On 400 records drawn at random before the screen and screened in a pilot by other sessions, the two runs agree on 95.8\% ($\kappa = 0.859$), and all 17 differences involve unsure; and of 35 SMOTE-family papers of the corpus that were matched to the records at that stage, it included 32 and marked 3 unsure.

\src{S}{}Another AI model (Claude Fable 5.2) then checked the screen, in separate sessions, under instructions that its own sessions wrote from the authors' criteria (archived with the data); Figure~\ref{fig:workflow} shows where this check, and every other step, sits and who did it. \src{A}{S}Blind to the screen's decisions, the check decided a random sample of 300 included and 300 excluded records; of the 596 that it included or excluded, it agrees with the screen on 579 (97.1\%, Cohen's $\kappa = 0.94$). It then decided, from title and abstract and under the rulings set out below, the 1,222 records that the screen had left unsure (609 include, 562 exclude, 51 unsure), corrected 11 decisions in its sample of included records, and checked the codes that AI sessions had given to 81 records of the two samples described below, correcting 6. Its decisions and corrections replace those of the screen.

\src{A}{S}Some records were deposits that their repository has since removed: of the 367 records with a Zenodo DOI, Zenodo had removed 167, nearly all as spam. A record whose only version has been removed cannot be read in full, so the authors excluded all 167 under a code of their own (E-WITHDRAWN, ruling A).

\src{A}{S}The check also raised questions about where criterion I1 ends, which the authors settled with three further rulings (an addendum to the codebook records them; online appendix, Section~S5). A chain of existing resampling steps counts as the paper's own method only if the paper presents it as its contribution and compares it with at least one other way of resampling (ruling B). A standard generator used without a stated change (a plain GAN, conditional GAN, WGAN-GP, CTGAN, VAE, diffusion or language model, KDE or GMM sampling, a copula, or standard data augmentation) never counts as the paper's own, even when it is chained with other steps and compared (B2). The task does not decide scope: segmentation, regression recast as classes, survival analysis, object detection, continual and zero-shot learning, and ranking count like classification (C). The check applied B and B2 to the records it decided. For the other included records, AI sessions (Claude Opus 5.5) recorded from each title and abstract the facts that the rulings turn on, with a confidence, and a script applied the rulings to these facts. Where this pass and the check read the same 299 included records, the pass was the stricter of the two: it fails 44 of them under B or B2, among them all 10 that the check excluded. \src{A}{S}Of the other 2,634 included records, 440 fail B or B2: the 108 whose facts were recorded with high confidence are excluded under E1, and the remaining 332 are marked unsure, together with the 51 that the check left unsure. \src{A}{S}The authors applied ruling C to the check's exclusions whose notes cite the task: they restored the four that oversample a rare class within it and kept out, under I1, the four that make data for unseen or for all classes.
%>>search:a_screen

%<<search:b_coverage_detail
\src{S}{}Of the 120 papers drawn for the coverage sample (Section~\ref{sec:search}), 88 could be obtained in full text. \src{A}{S}After the rulings above, a further session recorded their facts from the full text of the 10 papers coded eligible until then, and a paper whose method fails ruling B or B2 counts as excluded at full text (none of the 10). \src{A}{S}A second coding of these 88 papers by separate AI sessions, blind to the first and from the same full texts, agreed with it on 94.8\% of the coded values; an AI session settled the 101 differences from the papers, seeing the two codes in random order, and 47 of them changed a first code. \src{S}{}The authors checked 55 of the settled codes against the papers, among them every code that the adjudication changed, and found all of them correct.
%>>search:b_coverage_detail

\src{S}{}The authors decided one unsure record of the screen themselves, on 23 November 2026: SMOTEBoost \parencite{chawla2003smoteboost}, a corpus paper that they had read in full, which runs SMOTE inside each round of boosting as its own method and compares it with boosting without it (Section~\ref{sec:search}). The coverage sample had been coded by then. In the random order of its draw, SMOTEBoost stands among the first 30 records of the period to 2009 and would have taken the place of the thirtieth; the sample was kept as coded.

\src{S}{}The authors used an AI session (Claude Fable 5.2) to compare the records of the search with the reference lists of the five reviews of Table~\ref{tab:related} published in 2025 and 2026 \parencite{udu2025emerging,idwan2025comprehensive,hayaeianshirvan2025deep,gao2026comprehensive,yousefimehr2026balancing}, as Crossref held them on 21 November 2026. The five lists hold 695 references, 693 of which Crossref gives with a DOI or a title, and among them 594 distinct works dated 2002 or later. A reference was matched with a record by its DOI or, failing that, by its title. Of the 594 works, 220 are among the 22,426 records: 136 pass the screen and 84 are excluded, 32 of them as applications of existing resampling (E1) and 8 as left unsure (E-UNSURE). The session coded the other 374 from the title, and from the abstract where Crossref holds one: 45 as proposing a way of generating examples for under-represented classes, 16 as titles that leave this open, and 313 as other work, such as reviews, general generative models, undersampling, cost-sensitive and loss-based methods, and applications. The authors then coded the 45 and the 16: 40 propose such a method (36 of the 45 and 4 of the 16), 2 do not, and 19 were left undecided (8 of the 45 and 11 of the 16). Counting the 136 included records and these 40 as the method papers that the five reviews cite, the search finds 136 of 176 (77\%); with the 19 undecided works counted as well, it finds 136 of 195 (70\%). None of the 40 has both a generation term and an imbalance term of the search string in its title: most are generative adversarial, variational, and diffusion models used as oversamplers, mixup and feature augmentation for long-tailed recognition, and the synthesis of nodes for graphs, the branch that Section~\ref{sec:beyondtables} describes. For the review centered on variants of SMOTE \parencite{idwan2025comprehensive}, 41 of the 46 references are among the records and 34 of them pass the screen. Of the 136 cited papers that pass the screen, 18 are in the corpus. The matching and the coding of the 313 other works are reported as an AI first pass (Table~\ref{tab:aisteps}); the deposit holds each reference with its match, the session's code, and the authors' code.

\section{Details of the Checks of the Extractions}
\label{app:aichecks}

\src{A}{S}A second, blind pass followed in November 2026. Separate AI sessions, given the coding rules and the papers and blind to the first codes, recoded a seeded random sample of each extraction: 283 dataset entries and 97 printed comparisons of the Data extraction (1,706 values), the data-handling codes of all 76 papers, 11 of the 37 method papers, and 15 of the 71 empirical studies. An AI session (Claude Opus 5.5) then adjudicated every disagreement from the paper, with both passes side by side, under four conventions set by the authors: a two-class set whose class count is not printed has two classes; a value that follows exactly from printed numbers counts as printed; when a paper proposes several variants, the proposed method is the one its abstract names as the contribution, or the first it names; and modified data count as released only when the modified files are public.

\src{A}{S}On the Data values, the two passes agreed on 84.2\%, and the adjudication corrected 13 of the 1,706 values and filled 112 that the first pass had left blank. The same session checked a seeded fifth of the agreements against the papers and found all 351 correct. The online appendix (Section~S11) gives the agreement and the corrections for every coded field.

\src{S}{}The corrections are recorded with the data, and the counts of this survey rest on the verified sheets except where a correction or a convention changes them: the papers that modify or release data and the papers that give a reason for their data (online appendix, Section~S11), and the variant taken as the proposed method in four papers (Section~\ref{sec:wins}).

%<<search:k_wp3_spot
\src{S}{}The authors then checked 80 of the adjudications on a review page that showed, for each, both passes, the AI's value and reason, and the passage of the paper, and agreed with 79; the adjudication of the other stood after a further check that the authors asked for. \src{A}{S}The online appendix (Section~S11) lists the adjudications the page held.
%>>search:k_wp3_spot

%<<search:k_wp3_count
\src{S}{}The authors also checked, on a second review page, the remaining adjudications that move a count listed in Section~S11 of the online appendix, and agreed with all of them.
%>>search:k_wp3_count

\src{S}{}Working with agents affected the checks. Separate sessions can be kept blind: the second coders of both samples worked from the codebooks and the texts alone, in folders of their own, and their adjudicators saw the two codes in random order. A rule applied by a script to recorded facts can change without any record being read again: when the authors decided that only failures of rulings B and B2 recorded with high confidence are excluded, a rerun of the script applied the decision. A person's check can be prepared item by item: for each check of an adjudication, an AI session built a review page that showed every item with its codes, the reason given, and the passage of the paper, and recorded the authors' decision, which a script then merged.

\section{Denominators and Consistency Across the Three Extractions}
\label{sec:denominators}

\src{A}{S}Counts in this survey use nested sets of papers, and every share names the set it is taken from (Table~\ref{tab:corpusmap}). The core corpus holds 76 papers; 73 of them evaluate something on data and 72 name their datasets (the Data extraction); 71 are empirical studies in the sense of the Evaluation extraction; and 37 propose a SMOTE-family generator. Comparisons with SMOTE are counted at two levels, kept apart: 1,511 printed comparisons with plain SMOTE in 29 papers, and 558 comparisons with SMOTE inside a wrapper in twelve papers.

\begin{table}[tp]
\centering\small
\caption{\src{S}{}The sets of papers behind the counts of this survey, and what each count covers.}
\label{tab:corpusmap}
\begin{tabularx}{\linewidth}{L{5.6cm}rY}
\toprule
Set & Papers & What is counted on it \\
\midrule
Core corpus & 76 & Roles of the papers (Section~\ref{sec:corpus}) \\
Papers with an evaluation on data & 73 & 1,756 dataset entries, how they were obtained and modified (Section~\ref{sec:data}) \\
\quad of which name their datasets & 72 & 1,755 named entries, names and families (Section~\ref{sec:landscape}) \\
\quad of which use SMOTE as a baseline & 57 & 44 with plain SMOTE, 13 with SMOTE inside a wrapper (Section~\ref{sec:wins}) \\
\quad of which print per-dataset scores against plain SMOTE & 29 & 1,511 comparisons on 524 datasets (Figure~\ref{fig:winsmetric}) \\
Empirical studies & 71 & Evaluation design (Section~\ref{sec:evaluation}) \\
SMOTE-family methods & 37 & Decision points and class setting (Section~\ref{sec:family}) \\
\quad of which with a comparable difference from SMOTE & 23 & $\Delta$ vs SMOTE and criterion C1 (Tables~\ref{tab:methods} and~\ref{tab:recommend}) \\
\bottomrule
\end{tabularx}
\end{table}

\src{A}{S}The three extractions were coded for different questions, and some quantities appear in more than one of them, or in more than one place, under different definitions. The text uses one definition for each, in most cases that of the Data extraction, and the adjudication of the blind second pass (Section~\ref{sec:ai}) settled the two that had remained open. The online appendix (Section~S11) lists these quantities with their counts and how each difference was settled.

\clearpage
\section{Reproducibility Checklist for JAIR}
\label{app:repro}

% JAIR Reproducibility Checklist, word for word from the JAIR template (scripts/JAIR_Example_Template_2026-11-26.tex);
% written by tools/pass25_v67.py (v0.67). Each answer is one of the options the template offers; the explanations
% are in its last part. Do not reword the items: JAIR screens submissions for this checklist automatically.

Select the answers that apply to your research -- one per item.

\subsection*{All articles:}
\begin{enumerate}
\item All claims investigated in this work are clearly stated. [yes]
\item Clear explanations are given how the work reported substantiates the claims. [yes]
\item Limitations or technical assumptions are stated clearly and explicitly. [yes]
\item Conceptual outlines and/or pseudo-code descriptions of the AI methods introduced in this work are provided, and important implementation details are discussed. [yes]
\item Motivation is provided for all design choices, including algorithms, implementation choices, parameters, data sets and experimental protocols beyond metrics. [partially]
\end{enumerate}

\subsection*{Articles containing theoretical contributions:}
Does this paper make theoretical contributions? [no]

If yes, please complete the list below.

\begin{enumerate}
\item All assumptions and restrictions are stated clearly and formally. [NA]
\item All novel claims are stated formally (e.g., in theorem statements). [NA]
\item Proofs of all non-trivial claims are provided in sufficient detail to permit verification by readers with a reasonable degree of expertise (e.g., that expected from a PhD candidate in the same area of AI). [NA]
\item Complex formalism, such as definitions or proofs, is motivated and explained clearly. [NA]
\item The use of mathematical notation and formalism serves the purpose of enhancing clarity and precision; gratuitous use of mathematical formalism (i.e., use that does not enhance clarity or precision) is avoided. [NA]
\item Appropriate citations are given for all non-trivial theoretical tools and techniques. [NA]
\end{enumerate}

\subsection*{Articles reporting on computational experiments:}
Does this paper include computational experiments? [yes]

If yes, please complete the list below.

\begin{enumerate}
\item All source code required for conducting experiments is included in an online appendix or will be made publicly available upon publication of the paper. The online appendix follows best practices for source code readability and documentation as well as for long-term accessibility. [yes]
\item The source code comes with a license that allows free usage for reproducibility purposes. [yes]
\item The source code comes with a license that allows free usage for research purposes in general. [yes]
\item Raw, unaggregated data from all experiments is included in an online appendix or will be made publicly available upon publication of the paper. The online appendix follows best practices for long-term accessibility. [partially]
\item The unaggregated data comes with a license that allows free usage for reproducibility purposes. [yes]
\item The unaggregated data comes with a license that allows free usage for research purposes in general. [yes]
\item If an algorithm depends on randomness, then the method used for generating random numbers and for setting seeds is described in a way sufficient to allow replication of results. [yes]
\item The execution environment for experiments, the computing infrastructure (hardware and software) used for running them, is described, including GPU/CPU makes and models; amount of memory (cache and RAM); make and version of operating system; names and versions of relevant software libraries and frameworks. [partially]
\item The evaluation metrics used in experiments are clearly explained and their choice is explicitly motivated. [yes]
\item The number of algorithm runs used to compute each result is reported. [yes]
\item Reported results have not been ``cherry-picked'' by silently ignoring unsuccessful or unsatisfactory experiments. [yes]
\item Analysis of results goes beyond single-dimensional summaries of performance (e.g., average, median) to include measures of variation, confidence, or other distributional information. [yes]
\item All (hyper-) parameter settings for the algorithms/methods used in experiments have been reported, along with the rationale or method for determining them. [partially]
\item The number and range of (hyper-) parameter settings explored prior to conducting final experiments have been indicated, along with the effort spent on (hyper-) parameter optimisation. [yes]
\item Appropriately chosen statistical hypothesis tests are used to establish statistical significance in the presence of noise effects. [yes]
\end{enumerate}

\subsection*{Articles using data sets:}
Does this work rely on one or more data sets (possibly obtained from a benchmark generator or similar software artifact)? [yes]

If yes, please complete the list below.

\begin{enumerate}
\item All newly introduced data sets are included in an online appendix or will be made publicly available upon publication of the paper. The online appendix follows best practices for long-term accessibility with a license that allows free usage for research purposes. [yes]
\item The newly introduced data set comes with a license that allows free usage for reproducibility purposes. [yes]
\item The newly introduced data set comes with a license that allows free usage for research purposes in general. [yes]
\item All data sets drawn from the literature or other public sources (potentially including authors' own previously published work) are accompanied by appropriate citations. [yes]
\item All data sets drawn from the existing literature (potentially including authors’ own previously published work) are publicly available. [partially]
\item All new data sets and data sets that are not publicly available are described in detail, including relevant statistics, the data collection process and annotation process if relevant. [yes]
\item All methods used for preprocessing, augmenting, batching or splitting data sets (e.g., in the context of hold-out or cross-validation) are described in detail. [yes]
\end{enumerate}

\subsection*{Explanations on any of the answers above (optional):}
\begin{itemize}
\item All articles, item 1. \textbf{Yes} (research questions and contributions, Section~\ref{sec:intro}).
\item All articles, item 2. \textbf{Yes} (Sections~\ref{sec:method} to~\ref{sec:society}; levels of evidence, Section~\ref{sec:extractions}).
\item All articles, item 3. \textbf{Yes} (Section~\ref{sec:limitations}).
\item All articles, item 4. \textbf{Yes}: the one device the survey introduces, the generation template, is stated formally (Equation~\ref{eq:template}, Table~\ref{tab:decisions}).
\item All articles, item 5. \textbf{Partially}: the choice of the corpus is explained (Section~\ref{sec:corpus}), but it is curated; the systematic search that tests its coverage is reported in Sections~\ref{sec:search} and~\ref{sec:beyond}.
\item Theoretical contributions. \src{S}{}The paper makes no theoretical contribution, so the six items of this list are marked NA.
\item Experiments, the question. \textbf{Yes} (the independent benchmark of Appendix~\ref{app:benchmark}).
\item Experiments, item 1. \src{A}{S}\textbf{Yes}: deposited on Zenodo and, with the same files, on GitHub (Declarations, Data and code availability; online appendix, Section S4.4).
\item Experiments, item 2. \src{A}{S}\textbf{Yes}: the MIT license.
\item Experiments, item 3. \src{A}{S}\textbf{Yes}: the MIT license.
\item Experiments, item 4. \src{A}{S}\textbf{Partially}: the per-fold results of the benchmark and of the rerun are deposited, and the authors keep those of the rerun's TabPFN-3.5 arm under the model's license (Section~\ref{sec:beyondtables}; online appendix, Section S4.4).
\item Experiments, item 5. \src{A}{S}\textbf{Yes}: CC BY 4.0.
\item Experiments, item 6. \src{A}{S}\textbf{Yes}: CC BY 4.0.
\item Experiments, item 7. \textbf{Yes}: one seed fixes the folds and every sampler and learner, through NumPy's generator as scikit-learn uses it; the nested tier draws its datasets and inner folds with a second seed (Appendix~\ref{app:benchmark}).
\item Experiments, item 8. \src{S}{}\textbf{Partially}: for the core tier, Appendix~\ref{app:benchmark} gives the processor, its cores, the memory, the operating system, and the versions of Python, scikit-learn, and imbalanced-learn; for the cloud workspace of the rerun, it gives the operating system and the number of cores. Cache sizes are not reported.
\item Experiments, item 9. \textbf{Yes} (Section~\ref{sec:measures}).
\item Experiments, item 10. \textbf{Yes}: five repetitions of stratified 5-fold cross-validation, three in the nested tier (Appendix~\ref{app:benchmark}).
\item Experiments, item 12. \textbf{Yes}: tests over source families, median differences as effect sizes, and posterior probabilities from a Bayesian signed-rank test (Appendix~\ref{app:benchmark}).
\item Experiments, item 13. \textbf{Partially}: the samplers use their implementations' default settings and the learners the fixed settings listed in Appendix~\ref{app:benchmark}; the full values are in the released code.
\item Experiments, item 14. \textbf{Yes}: the nested tier tunes one setting per learner over three values by inner 3-fold cross-validation on PR-AUC (Appendix~\ref{app:benchmark}); the other tiers tune only the decision threshold of the tuned-threshold baseline.
\item Experiments, item 15. \textbf{Yes}: the benchmark uses Friedman, Holm-corrected Wilcoxon, and Bayesian signed-rank tests over source families (Appendix~\ref{app:benchmark}); the corpus analyses of Section~\ref{sec:wins} use clustered models (online appendix, Section~S9).
\item Data sets, the question. \textbf{Yes}: the coding sheets of the three extractions are new data sets; the benchmark uses public datasets from the imbalanced-databases package and OpenML.
\item Data sets, item 1. \src{A}{S}\textbf{Yes}: the full coding sheets are deposited with a DOI under CC BY 4.0, and the main codes of the method and evaluation extractions and the reference collection of tasks are also in the online appendix (Sections S1 to S3).
\item Data sets, item 2. \src{A}{S}\textbf{Yes}: CC BY 4.0.
\item Data sets, item 3. \src{A}{S}\textbf{Yes}: CC BY 4.0.
\item Data sets, item 4. \textbf{Yes} (the corpus papers, the KEEL repository, and OpenML).
\item Data sets, item 5. \textbf{Partially}: the corpus papers are published; some are behind paywalls, and some papers' own data are not public (Section~\ref{sec:checkable}).
\item Data sets, item 6. \textbf{Yes} (Sections~\ref{sec:extractions} and~\ref{sec:ai}; codebooks in the online appendix, Section S5).
\item Data sets, item 7. \textbf{Yes} (Section~\ref{sec:other-methods} and Appendix~\ref{app:benchmark}).
\end{itemize}

\end{document}